\documentclass[]{bytedance_seed}

\usepackage{minitoc}

\usepackage{solarized-light}
\usepackage{booktabs}
\usepackage{multirow}
\usepackage{graphicx}
\usepackage{array}
\usepackage{makecell}
\usepackage{framed}
\usepackage{hyperref}
\usepackage{amsmath}  
\usepackage{amssymb}
\usepackage[colorinlistoftodos]{todonotes}
\usepackage{longtable}
\usepackage{hhline}
\usepackage{fancyvrb}
\usepackage{fvextra}
\usepackage{CJKutf8}
\usepackage{multicol}
\usepackage{pifont}
\usepackage{cleveref}
\usepackage{tablefootnote}
\usepackage{threeparttable}
\usepackage{tabularx}
\usepackage{mdframed}
\usepackage{subcaption}
\usepackage[usestackEOL]{stackengine}
\usepackage[numbers]{natbib}
\newcommand{\commentout}[1]{}
\renewcommand{\paragraph}[1]{\noindent\textbf{#1.}\hspace*{1em}}
\usepackage{enumitem}
\setlist[itemize]{leftmargin=15pt}
\usepackage{xcolor}
\usepackage{colortbl}

\RequirePackage{xspace}
\makeatletter
\DeclareRobustCommand\onedot{\futurelet\@let@token\@onedot}
\def\@onedot{\ifx\@let@token.\else.\null\fi\xspace}

\makeatother

\newcommand{\gametars}{Game-TARS\xspace}

\usepackage[most]{tcolorbox}
\newtcolorbox[list inside=prompt,auto counter,number within=section]{prompt}[1][]{
    colbacktitle=black!60,
    coltitle=white,
    fontupper=\footnotesize,
    boxsep=5pt,
    left=0pt,
    right=0pt,
    top=0pt,
    bottom=0pt,
    boxrule=1pt,
    #1,
}

\title{\gametars: Pretrained Foundation Models\\ for Scalable Generalist Multimodal Game Agents}

\author{Bytedance Seed}

\contribution{See \hyperref[sec:contributions]{Contributions} section for a full author list.}

\abstract{
We present \gametars, a generalist game agent trained with a unified, scalable action space anchored to human-aligned native keyboard–mouse inputs. Unlike API- or GUI-based approaches, this paradigm enables large-scale continual pre-training across heterogeneous domains, including OS, web, and simulation games. \gametars is pre-trained on over 500B tokens with diverse trajectories and multimodal data. Key techniques include a decaying continual loss to reduce causal confusion and an efficient Sparse-Thinking strategy that balances reasoning depth and inference cost. Experiments show that \gametars achieves about 2 times the success rate over the previous sota model on open-world Minecraft tasks, is close to the generality of fresh humans in unseen web 3d games, and outperforms GPT-5, Gemini-2.5-Pro, and Claude-4-Sonnet in FPS benchmarks. Scaling results on training-time and test-time confirm that the unified action space sustains improvements when scaled to cross-game and multimodal data. Our results demonstrate that simple, scalable action representations combined with large-scale pre-training provide a promising path toward generalist agents with broad problem-solving abilities.
\footnotetext[1]{Project: \url{https://seed-tars.com/game-tars}}  
}
\date{October 27, 2025}
\correspondence{\email{yujia.qin@bytedance.com};
\email{shiguang.sg@bytedance.com}}
\begin{document}

\maketitle

\begin{figure*}[h]
    \centering
    \includegraphics[width=0.99\linewidth]{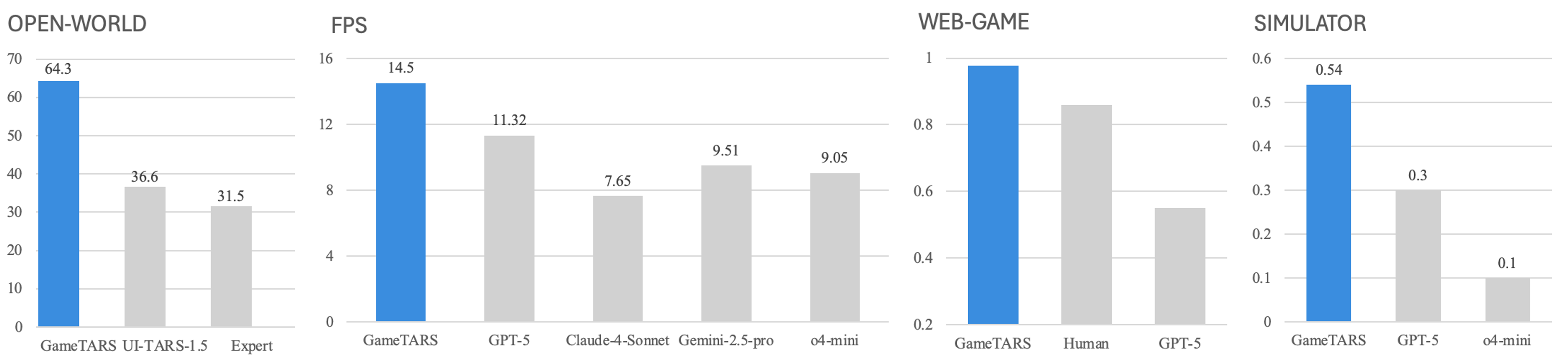}
    \caption{
    \gametars achieves a higher level of performance compared to humans, domain experts, and general VLMs in unseen 3D virtual environments, including open-world~\citep{minedojo}, FPS games~\citep{vizdoom}, web games, and simulators~\citep{miniworld}.
    }
    \label{fig:teaser_results}
\end{figure*}

\newpage
\tableofcontents
\newpage

\begin{figure}[htp]
    \centering
    \includegraphics[width=0.99\linewidth]{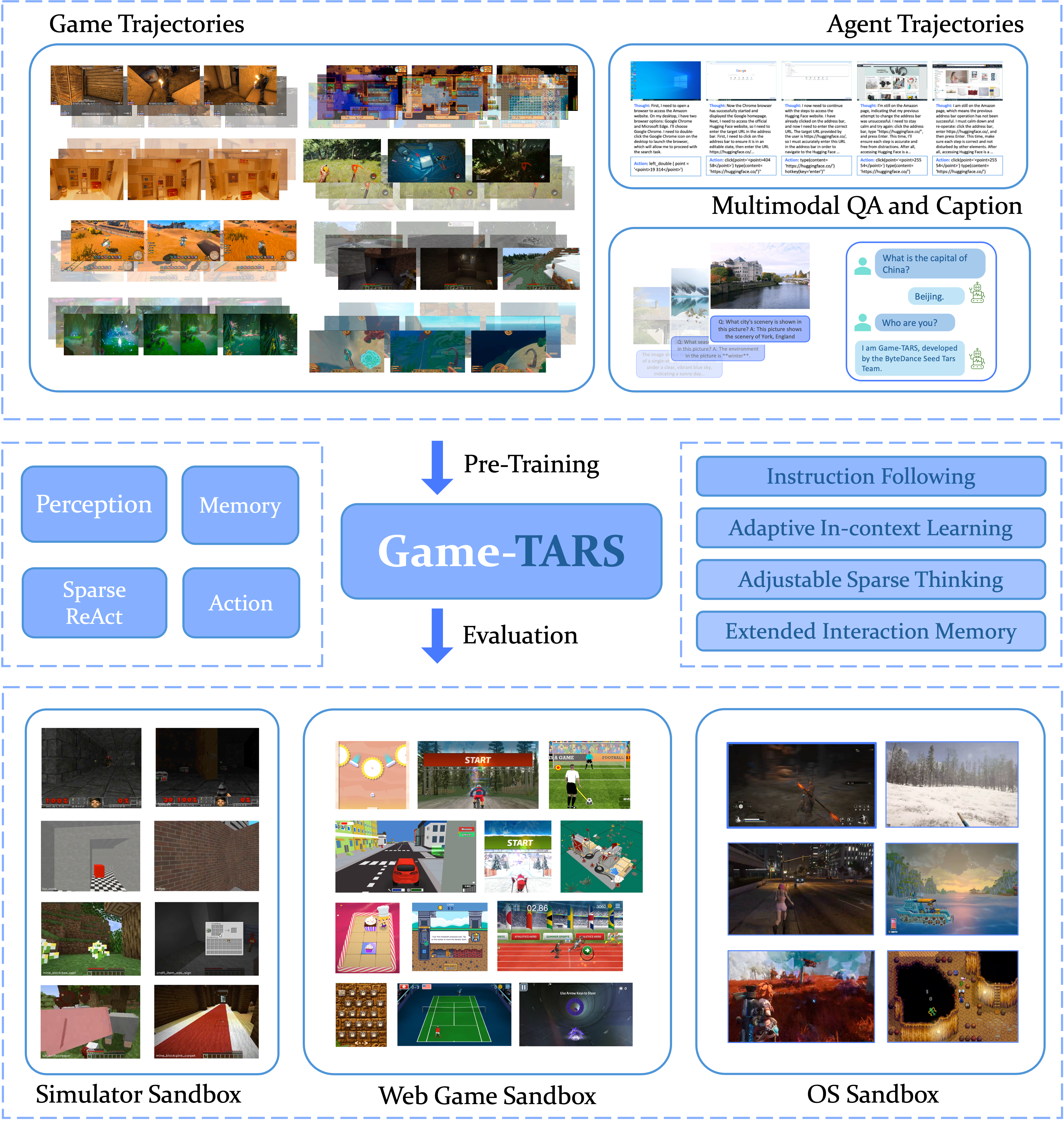}
    \caption{
    \textbf{Generalist Game Agent \gametars}.
    \gametars can interpret and respond to various human instructions across diverse environments using a single neural network with a consistent set of weights. It was pre-trained on a wide range of multimodal datasets, including vision-language question-answering, captioning, over 20k hours of game trajectories, GUI agent trajectories, and more.
    }
    \label{fig:teaser}
\end{figure}

\section{Introduction}
Building Generalist Artificial Agents capable of seamlessly interacting with complex and dynamic digital environments has emerged as a key research path toward achieving Artificial General Intelligence (AGI)~\citep{morris2023levels,gpt5intro,cheng2024exploring}. Video games, with their diverse task objectives, intricate interaction logic, and rich visual information, provide an ideal platform for both training and evaluating such agents~\citep{alphago,alphastar,bruce2024genie}. Despite substantial progress, existing approaches still face significant challenges in creating truly scalable agents with broad generalization capabilities~\citep{sima,wang2025ui,cradle}. One of the core bottlenecks is the definition of the action space: traditional agents often rely on highly customized, pre-defined action sets tailored for specific environments~\citep{openai-deepresearch,openai_2025_cua_blog,Anthropic2024MCP}. This tight coupling between the agent and the environment severely limits the agent's generalization, making it difficult to adapt to new tasks, particularly in open-world settings~\citep{minedojo,vpt}.

To overcome these limitations, we propose a novel framework for pre-training general game agents. Our central contribution introduces a fundamental shift in paradigm: we move away from application-specific high-level action instructions~\citep{wang2023voyager,wang2023describe} and instead define a scalable, unified action space directly tied to low-level computer input devices, such as the keyboard and mouse. This \textbf{native human interaction} paradigm ensures that the agent can operate within any graphical user interface (GUI) environment, laying the foundation for large-scale, cross-domain pre-training.

Utilizing this unified action space, we have developed a comprehensive agent training process that includes continual pre-training and post-training stages on a vision-language model~\citep{guo2025seed1,bai2025qwen2,openai2023gpt4v,comanici2025gemini}. In the continual pre-training phase, we employ an efficient, native ReAct paradigm~\citep{yao2022react} that incorporates \textbf{Sparse Thinking}—interleaving reasoning and action only at critical decision points to mirror human cognitive patterns. This is achieved by combining offline chain-of-thought synthesis~\citep{wei2022chainofthought} with an online \textsc{think-aloud} protocol~\citep{deitke2024molmo} to collect and refine high-quality, causally consistent multimodal trajectory data. In the post-training phase, we implement a series of targeted fine-tuning strategies to systematically enhance the agent's core capabilities. These strategies include: enforcing strict adherence to instructions through automated action space augmentation; fostering dynamic context learning via multimodal prompts and long-range trajectory learning; refining the agent's decision-making by enforcing efficient Sparse Thinking through rejection fine-tuning; and developing long-term memory abilities via a dual-layer memory architecture.

To extend the agent's capabilities beyond gaming, we also incorporate cross-domain agent trajectories from fields such as code generation, GUI automation, and research tasks into the post-training data. This diverse mix of data aims to transform the model from a specialized game-player into a versatile general computer-user with broader problem-solving capabilities. The work presented in this paper offers a comprehensive, end-to-end solution for developing the next generation of generalist agents.

We first validated the effectiveness of the unified action space and large-scale continual pretraining by testing the agent's capabilities in the open-world game Minecraft~\citep{minedojo}. Unlike grounding-based~\citep{seed2025uitars15} or API-based~\citep{wang2023voyager} actions, we found that the unified action space, when trained with less than 10B data, did not initially outperform GUI-based action agents in terms of success rate. However, the key advantage of the unified action space lies in its scalability—large-scale data can be efficiently collected, and the agent can be pretrained using a consistent format across all games. Our scaling experiments demonstrate that this unified action space provides an excellent foundation for the realization of generalist agents. When trained on a large corpus of general computer-use data, \gametars, based on the unified action space, achieved approximately a 2x performance improvement in Minecraft compared to previous state-of-the-art expert models~\citep{openha}.

Furthermore, we conducted extensive evaluations on a wide range of untapped benchmarks, including FPS-game Vizdoom~\citep{vizdoom}, 3D Simulator Miniworld~\citep{miniworld}, and online web games~\citep{qin2025ui}. The results highlight the exceptional generalization ability of our generalist agent. Our mini version of \gametars outperformed renowned pre-trained models such as Gemini-2.5-Pro~\citep{comanici2025gemini}, GPT-5~\citep{gpt5intro}, and Claude-4-Sonnet~\citep{claude4intro} on various tasks, showcasing superior performance across diverse domains. These results provide compelling evidence of the broad applicability and robustness of the unified action space and continual pretraining approach, establishing \gametars as a formidable agent in a wide variety of environments.

\section{Continual Pre-Training for Generalist Game Agent}

Our objective is to model an inherently multi-task generalist agent. We formulate this agent as a autoregressive, parameterized policy, $\pi_{\theta}$. The policy's role is to map the agent's history and current perception to a decision. 
At any given timestep $t$, the agent receives an \textbf{Observation ($o_t$)} from the environment (specifically, the current visual image frame). It also maintains an \textbf{internal memory state ($h_t$)}, which serves as a compressed summary of all past interactions. 
Following the ReAct paradigm~\citep{yao2022react}, the agent's decision-making process is modeled as a two-stage sequence:
\begin{itemize}
    \item \textbf{Reasoning ($r_t$):} An internal cognitive process that encompasses context analysis, memory retrieval, planning, and self-reflection.
    \item \textbf{Action ($a_t$):} An external interaction with the environment.
\end{itemize}
The policy $\pi_{\theta}$ is responsible for generating this entire reasoning and action chain. The core policy function at timestep $t$ is thus defined as:
\begin{equation}
    (r_t, a_t) \sim \pi_{\theta}(\cdot \mid o_t, h_t)
\end{equation}
where the memory state $h_t$ is a function of all preceding steps, i.e., $h_t = f( (r_i, a_i, o_i)_{i=0}^{t-1} )$.
We define a complete ReAct cycle—comprising the agent's reasoning, its action, and the environment's subsequent observation—as a single "step": $(r_t, a_t, o_t)$.
Therefore, a complete trajectory ($\tau$) of length $T$, representing the full log of the agent's interaction, is formalized as a sequence of these steps:
\begin{equation}
    \tau = {(r_0, a_0, o_0), (r_1, a_1, o_1), \dots, (r_T, a_T, o_T)}
\end{equation}
Our training objective is to optimize the parameters $\theta$ of our policy $\pi_{\theta}$ using a large dataset of these trajectories.




\subsection{A Scalable and Generalist Action Space for Computer-Use Agents}

\begin{table}[]
\centering
\caption{Taxonomy of computer-use agent action spaces, contrasting high-level, environment-specific APIs with the low-level, general-purpose device controls.}
\label{tab:action_space}
\resizebox{\textwidth}{!}{%
\renewcommand\arraystretch{1.1}
\begin{tabular}{@{}lclll@{}}
\toprule
Action Hierarchy & Agent & Action Level & Action & Definition \\ \midrule
\multicolumn{1}{l|}{\multirow{12}{*}{\includegraphics[width=0.35\linewidth]{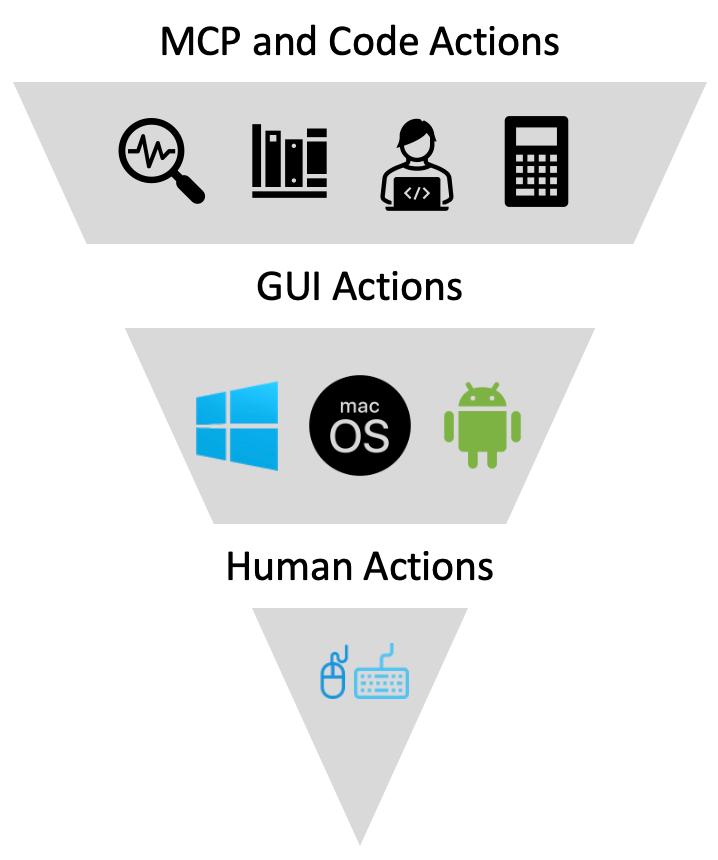}}} & MCP & Application & \texttt{Search(query)} & Retrieve the relevant URLs for the query. \\
\multicolumn{1}{l|}{} & MCP & Application & \texttt{ReadUrl(url)} & Get the content in URL. \\ 
\multicolumn{1}{l|}{} & MCP & Application & \texttt{CreateFile(filePath)} & Create a new file at the specified file path. \\ 
\multicolumn{1}{l|}{} & Code & Application & \texttt{Run(script)} & Execute a script file or program. \\
\multicolumn{1}{l|}{} & Code & Application & \texttt{Compile(code)} & Compile source code into an executable. \\ \cmidrule(l){2-5} 
\multicolumn{1}{l|}{} & GUI & System-Level(Desktop) & \texttt{Drag(x1, y1, x2, y2)} & Drag from (x1, y1) to (x2, y2) \\
\multicolumn{1}{l|}{} & GUI & System-Level(Desktop) & \texttt{scroll(200)} & Scrolls up by 200 units. \\
\multicolumn{1}{l|}{} & GUI & System-Level(Desktop) & \texttt{Hotkey(key)} & Presses the specified hotkey. \\
\multicolumn{1}{l|}{} & GUI & System-Level(Desktop) & \texttt{LeftDouble(x, y)} & Double-clicks at (x, y). \\
\multicolumn{1}{l|}{} & GUI & System-Level(Mobile) & \texttt{LongPress(x, y)} & Long presses at (x, y). \\
\multicolumn{1}{l|}{} & GUI & System-Level(Mobile) & \texttt{PressHome()} & Presses the ``home'' button. \\ \cmidrule(l){2-5} 
\multicolumn{1}{l|}{} & - & Device-Level(Mouse) & \texttt{mouseMove(dx, dy)} & Move the mouse in the dx and dy directions. \\
\multicolumn{1}{l|}{} & - & Device-Level(Mouse) & \texttt{mouseClick(buttons)} & Click the mouse buttons. \\
\multicolumn{1}{l|}{} & - & Device-Level(Keyboard) & \texttt{keyPress(keys)} & Press buttons in keyboard. \\ \bottomrule
\end{tabular}%
}
\end{table}

A critical challenge in constructing a generalist agent is the definition of an action space that is both broadly applicable and highly scalable. Traditional agents, such as those trained in environments like Minecraft~\citep{dreamerv3,vpt} or StarCraft~\citep{alphastar}, typically operate on a predefined, task-specific action space. These action spaces, whether encapsulated via game-internal protocols (e.g., Minecraft Protocol) or code-level APIs~\citep{wang2023voyager}, necessitate the design and implementation of a unique instruction set for each new environment or task. While effective for specialized domains, this approach suffers from a fundamental limitation: the tight coupling of the action space to the environment and task, which severely constrains the model's generality and scalability. Adapting the agent to a novel game or application often requires substantial manual effort in redefining action spaces, engineering API wrappers, and annotating data, thereby impeding generalization to open-world scenarios.

To partially mitigate this issue, prior work has introduced agents that operate on Graphical User Interfaces (GUIs), such as UI-TARS~\citep{qin2025ui}, by identifying and manipulating GUI elements. Although such methods improve generality, we observe that their action spaces remain constrained by the underlying operating system (OS) or specific UI frameworks. Significant discrepancies in GUI element structures and interaction modalities across Windows, macOS, and Linux, or even between different software applications, prevent the realization of a truly universal action space.

To transcend these limitations, we propose a more fundamental solution: \textbf{abstracting the agent's action space to a lower level and grounding it directly in the most universal input devices for human-computer interaction—the keyboard and mouse}. By doing so, the agent's action instruction set is decoupled from any high-level application or OS-specific API, achieving direct alignment with a human user's physical operations. We term this the \textit{Human-Native Interaction} paradigm. This paradigm ensures an unprecedented level of universality for the action space across all computer graphical environments.

Specifically, we define the agent's action space as a composition of three primitive operations:

\begin{itemize}
\item \textbf{\texttt{mouseMove(dx, dy)}:} This action simulates the two-dimensional movement of the mouse cursor on the screen. The parameters dx and dy represent the relative displacement along the horizontal and vertical axes, respectively.
\item \textbf{\texttt{mouseClick(button)}:} This action simulates a mouse button click event. The button parameter specifies the target button, such as left, right, or middle.
\item \textbf{\texttt{keyPress(keys)}:} This action simulates the press and release of one or more keyboard keys. The keys parameter can be a single key (e.g., \texttt{keyPress(w)}) or a combination of keys (e.g., \texttt{keyPress(ctrl, c)}).
\end{itemize}

We augment this core set with supplementary actions such as \texttt{Think}, \texttt{LongThinking}, and \texttt{No-Op} to facilitate more complex reasoning and behavior. The complete action space and its descriptions are provided in Table~\ref{tab:action_space}.

The principal advantage of grounding the action space in these input primitives is its exceptional scalability, enabling our generalist agent to operate in any standard computing environment without modification, much like a human user. More importantly, in contrast to traditional discrete action spaces centered on high-level maneuvers, our Human-Native Interaction paradigm is inherently \textit{time-aligned}. Each action primitive corresponds directly to a physical input at a specific point in time. This design obviates the need for special modeling of action durations or complex temporal dependencies, as the agent can autonomously learn the timing and rhythm of its operations through experience. Consequently, whether interacting with 2D pixel-based games, complex 3D open-world environments, or standard desktop applications, the agent employs the same unified and stable action space. This consistency establishes a robust foundation for large-scale, cross-domain Continual Pre-Training.

\subsection{Native Sparse ReAct Pretraining via Thinking Aloud}

\begin{figure*}[t]
    \centering
    \includegraphics[width=0.95\linewidth]{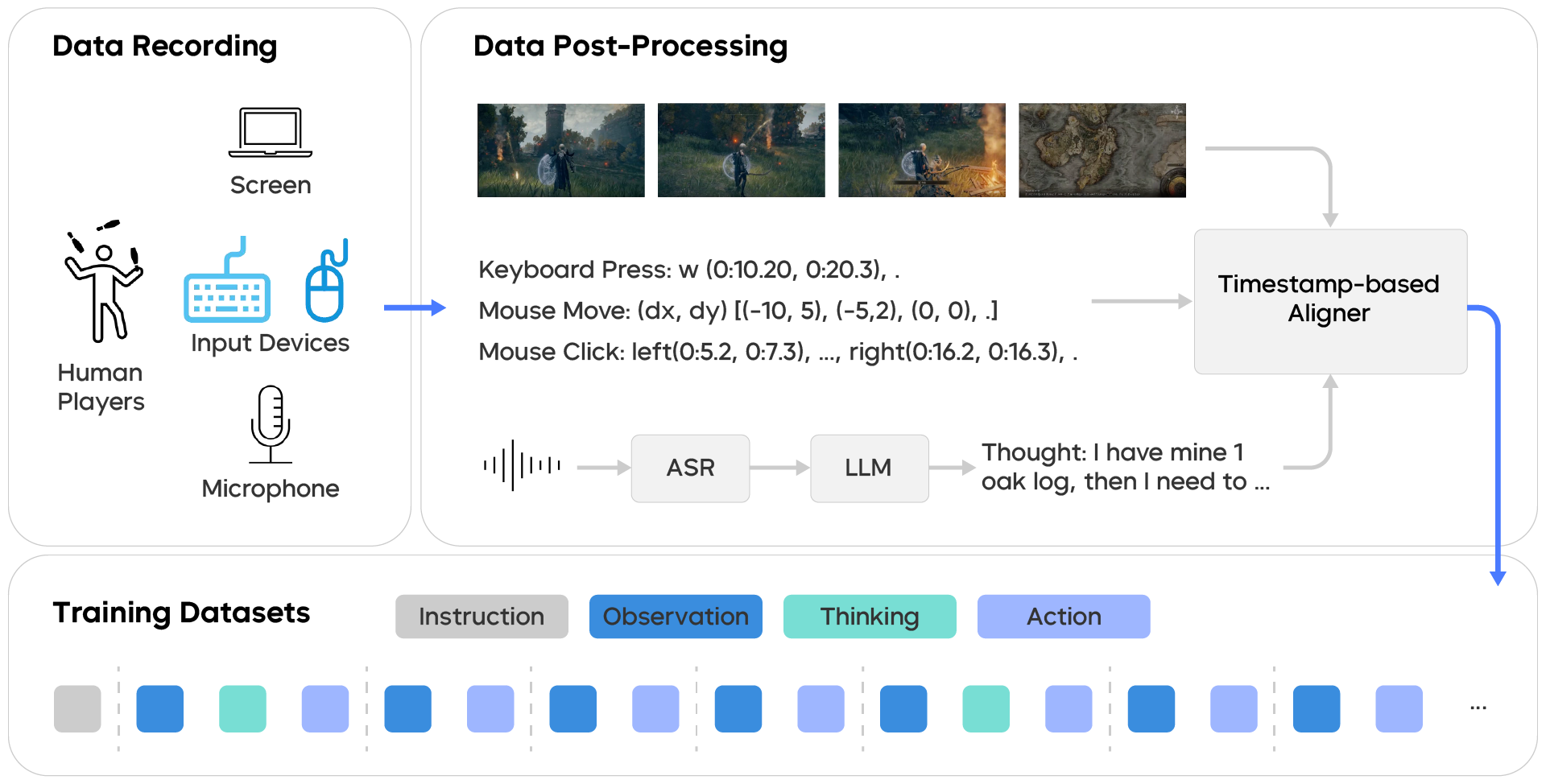}
    \caption{
    \textbf{The pipeline of \textsc{Think-Aloud} data collection and post-processing.}
    This process captures and synchronizes three types of original inputs (screen, keyboard, and mouse, audio), refines sparse-thinking through the ASR-LLM pipeline, and uses a timestamp aligner to synthesize the final (Instruction, Observation, Thinking, Action) datasets.
    }
    \label{fig:data_collection_flow}
\end{figure*}

Prior research has demonstrated that co-modeling reasoning and action as a unified output for an agent significantly enhances model generalization and performance~\citep{qin2025ui,song2025r1,yao2022react}. 
Recent methods usually employ a pre-trained vision language model~\citep{guo2025seed1} to inject reasoning steps into the raw trajectory data $\tau=\{(o_t, a_t)\}_{t=0}^{T}$ from human gameplaying, translating the observation-action pairs into the ReAct format.
However, previous studies have shown that retrospectively adding reasoning to recorded action sequences is exceptionally difficult and inefficient~\citep{wang2025ui,qin2025ui}, as accurately reconstructing the annotator's real-time thought process is nearly impossible. 
On the other hand, our time-aligned, high-frequency action modeling (with action intervals of only 50-100ms) results in action sequences with numerous repetitive and continuous operations, during which a human player's cognitive state does not significantly change. Consequently, forcing the generation of a reasoning step at every action step is both impractical and unnecessary.
Building upon this, we have designed an online \textsc{think-aloud} data collection and processing pipeline to generate high-quality, native Sparse ReAct trajectories for pre-training. 

\paragraph{Unified Trajectory Data Paradigm}
We define the pre-training trajectories in a sparse-thinking format, formalized as:
\begin{equation}
    \tau = {(r_0, a_0, o_0), \dots, (a_n, o_n), \dots, (r_m, a_m, o_m), \dots, (r_T, a_T, o_T)}.
\end{equation}
Here, some timesteps $m$ contain the complete tuple of reasoning, action, and observation $(r_m,a_m,o_m)$, while other timesteps $n$ only contain action-observation pairs $(a_n,o_n)$. To ensure the structural integrity of each trajectory, we mandate that reasoning steps must be included at the beginning ($t=0$) and end ($t=T$) of the trajectory, corresponding to planning and summarization, respectively.

\paragraph{Online Think-Aloud Data Collection}
Inspired by \citet{deitke2024molmo}, we instead adopt a think-aloud protocol, requiring annotators to verbalize their thoughts in real-time via audio while performing tasks, as demonstrated in \autoref{fig:data_collection_flow}. This method allows us to collect ReAct sequences containing native, in-situ reasoning. To ensure the density and quality of the reasoning data, we instruct annotators to provide thoughts for at least three key moments per minute.

\paragraph{Data Recording and Preliminary Processing}
Our data acquisition system utilizes a parallel recording program to synchronously capture a player's screen frames (as observations $o_t$), low-level mouse and keyboard input signals (as actions $a_t$), and the audio stream from the "\textsc{think-aloud}" process (as raw thoughts $r_t$). After collection, we use an ASR model~\citep{radford2023robust} to transcribe the audio into text, which is then refined using a Large Language Model (LLM) to remove noise and enhance logical consistency.

\paragraph{Causal Alignment via Visual Anchors}
The precise alignment of multi-modal data, particularly ensuring the causality of observation-action pairs, is a critical step. In practice, we found that due to minor latencies in the recording system, relying solely on timestamps causes action signals to lag behind their corresponding screen frames. This temporal misalignment severely disrupts the data's causal relationship. If used directly for training, it can lead the agent model to degenerate into an Inverse Dynamics Model (IDM)~\citep{vpt}—where the model learns to infer actions from state differences rather than deciding based on the current state—which would significantly impair its performance in practical applications. We developed a realignment method based on visual anchors to overcome this issue. We use the on-screen mouse cursor as a visual anchor and employ a visual grounding model~\citep{wang2025ui} to precisely track its pixel position in each frame. By comparing the tracked changes in cursor position with the recorded mouse movements \texttt{(dx, dy)}, we can accurately realign each action signal to the exact observation frame in which it was executed, thereby completely correcting the timestamp bias and restoring the causal chain.

\paragraph{Trajectory Construction and Refinement}
After achieving causal alignment, we construct training sequences centered on observation frames. For the time window between any two consecutive frames, all recorded keyboard and mouse operations are merged into a single action event; a lack of operations is marked as a no-op. Finally, to reinforce the semantic association between reasoning and action, we implement a \textbf{thought timestamp relocation} strategy. This involves intelligently adjusting the timestamp of a thought to align with the subsequent action to which it is most semantically relevant, ultimately producing highly synchronized and causally consistent \texttt{(observation $o_t$, reasoning $r_t$, action $a_t$)} training sequences.

\subsection{Continual Pre-training with Decaying Loss Function}

In our auto-regressive training framework, the agent predicts an action $a$ at each timestep $t$ based on the comprehensive history of all preceding observations and actions, denoted by $H_t=(o_1,a_1,o_2,a_2,\ldots,a_{t-1},o_t)$. 
A critical challenge we identified in our preliminary experiments is the severe imbalance in the action distribution within non-Markovian agent trajectories~\citep{zhao2023learning}, and this trend intensifies as the trajectory length increases. 
Specifically, temporally persistent actions, where $a_t=a_{t-1}$, constitute the vast majority (over 90\%) of the action sequence\footnote{The detailed value is different on different games.}. This high frequency of repetitive, low-entropy actions causes the standard cross entropy loss to be dominated by these easy samples and leads to the causal confusion in imitation learning~\citep{de2019causal}, thereby masking the learning signal from the more challenging and crucial action transitions (where $a_t\neq a_{t-1}$). 
Previous methods, such as VPT~\citep{vpt}, remove all no-ops from the trajectory; however, this disrupts the complete action distribution, as waiting is necessary in some scenarios. 
To mitigate this issue, we introduce a history-aware loss re-weighting scheme that dynamically attenuates the loss contribution from consecutive identical actions. Formally, we redefine the training loss for a trajectory $\tau=(o_1,a_1,\ldots,o_T,a_T)$ as:
\begin{equation}
    \mathcal{L} = -\sum_{t=1}^{T} \omega_t \log P(a_t\mid H_t)
\end{equation}
The weight $\omega_t$ for the action at timestep $t$ is determined by its consistency with the immediate past. Let $k_t$ be the number of consecutive timesteps including $t$ where the action has remained unchanged (i.e., $a_t=a_{t-1}=\ldots=a_{t-k_t} $ and $a_{t-k_t}\neq a_{t-k_t+1}$. The weight $\omega_t$ is then defined as an exponential decay:
\begin{equation}
    \omega_t=\gamma^{k_t-1}
\end{equation}
Here, $\gamma\in(0,1)$ is a decay factor hyperparameter (in our implementation, $\gamma=1/2$). This formulation assigns a full weight of $1$ (since $\gamma^0=1$) to any action that is new or different from the previous one, while exponentially reducing the weight for each subsequent repetition. By systematically down-weighting the loss from monotonous action sequences, this method compels the model to focus more of its capacity on learning the high-entropy decision boundaries associated with action changes, leading to a more robust and capable policy.

\begin{figure*}[t]
    \centering
    \includegraphics[width=0.95\linewidth]{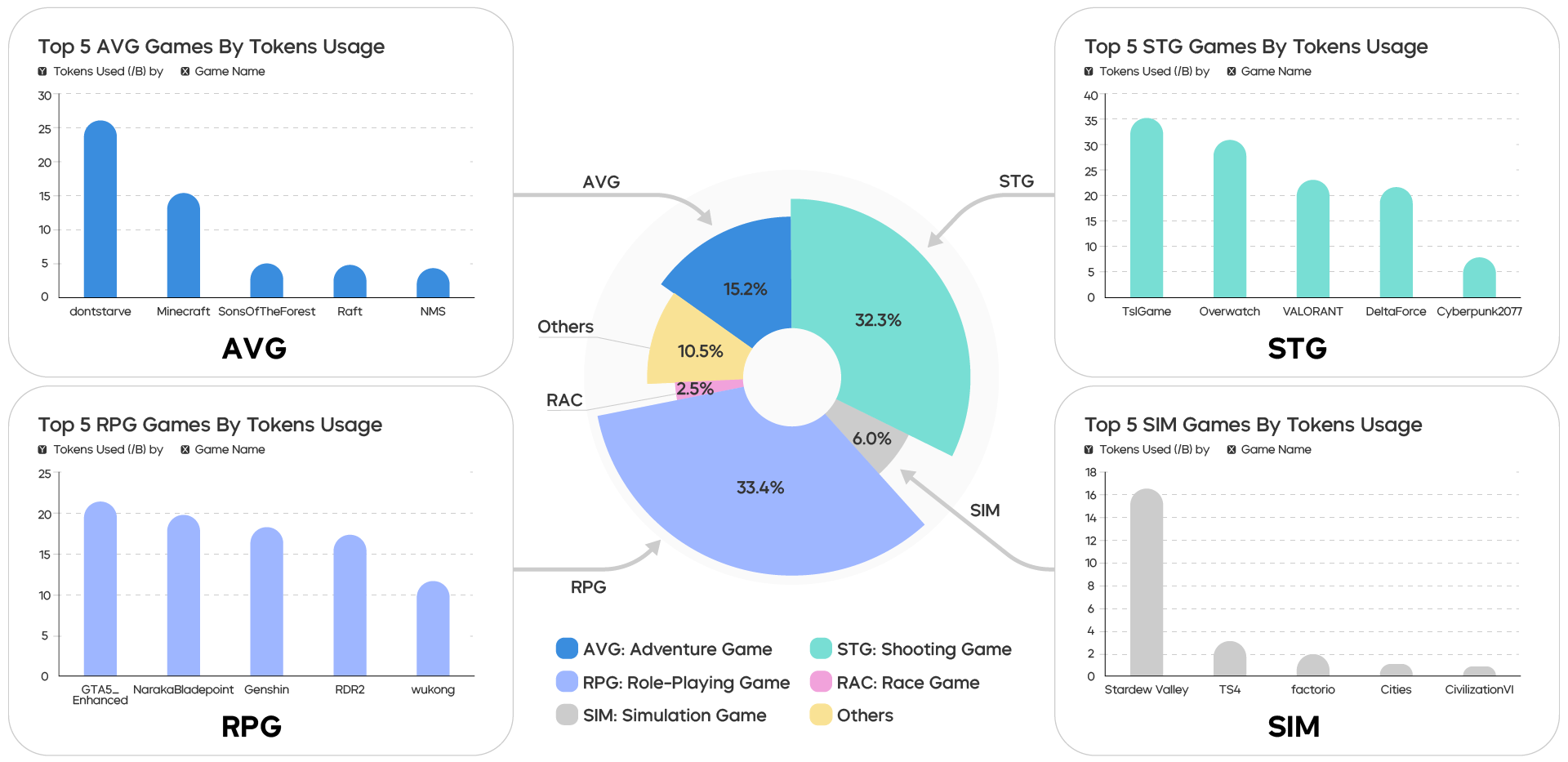}
    \caption{
    Distribution of different game types in the \gametars training dataset.
    }
    \label{fig:dataset_distribution}
\end{figure*}

\begin{figure*}[t]
    \centering
    \includegraphics[width=0.95\linewidth]{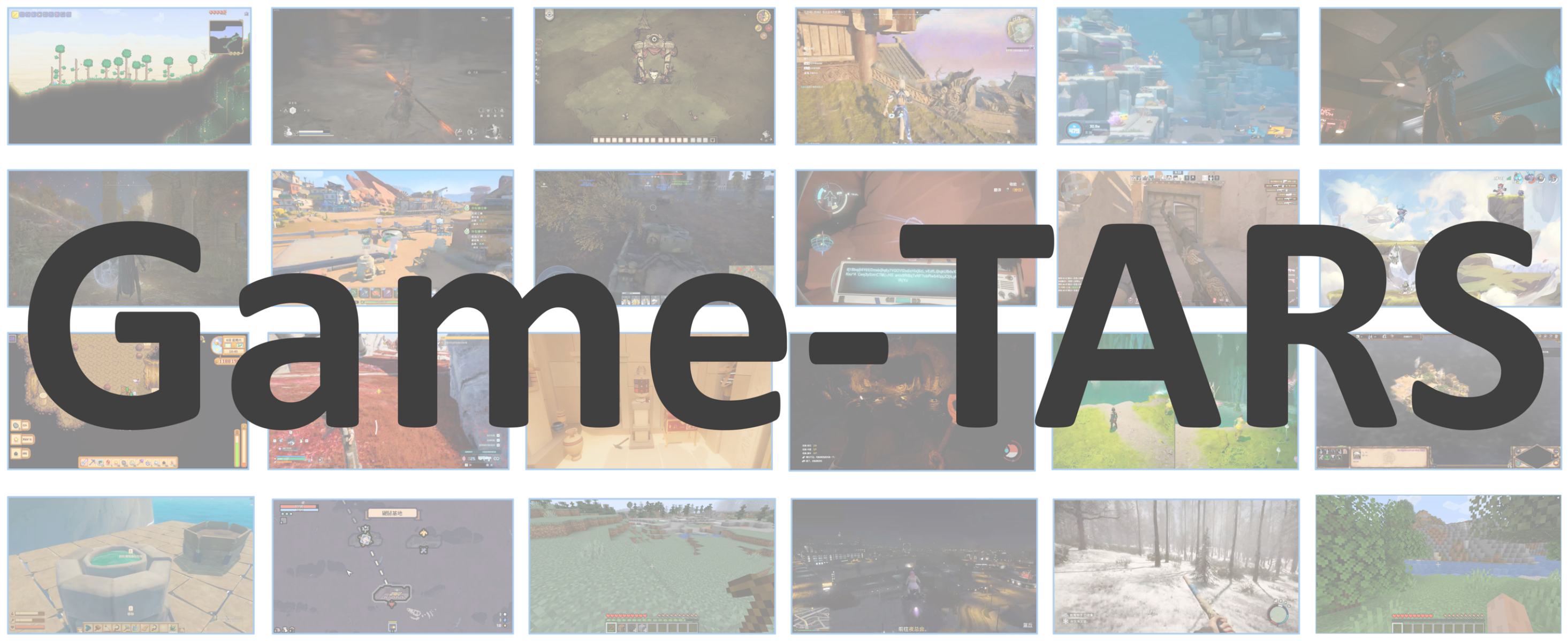}
    \caption{
    \gametars is trained on a wide range of games, including adventure, shooting, role-playing, and racing.
    }
    \label{fig:training_dataset_demo}
\end{figure*}

\section{Post-Training for Generalist Game Agent}

After a large-scale continual pre-training on over 20k game hours (about 500B tokens), we introduce a post-training stage to further enhance the agent's execution capabilities and interactive intelligence in specific tasks. This stage focuses on strengthening three core abilities: \textbf{instruction following}, \textbf{sparse thinking}, and \textbf{long-term memory}.

\subsection{Instruction Following}
To build a generalist agent that can interact seamlessly across different games and application environments, the primary task is to ensure it can strictly follow the instructions for a given task, particularly in understanding and using the action space of a specific environment. Although continual pre-training endows the model with powerful general interaction abilities, it also leads to the learning of a prior distribution over actions based on a massive dataset. This distributional bias presents two major challenges during the post-training phase:

\begin{itemize}
\item \textbf{Action Space Violation:} The model tends to output the actions defined in the pre-training data, but they are invalid in the current task. For example, even if the System Prompt specifies the \texttt{UpArrow} as the jump key, the model might still output \texttt{Space} to perform a jump out of habit.
\item \textbf{Behavioral Inertia:} The model may ignore current observations and instructions, instead relying on the frequency distribution of actions to minimize the loss function. This manifests as the model frequently outputting common actions from the pre-training data (e.g., the \texttt{W} key) rather than making decisions based on the current context.
\end{itemize}

\paragraph{Automatic Action Space Augmentation}
To address these issues, we propose a fine-tuning strategy called Automatic Action Space Augmentation. The core idea of this strategy is to decouple the semantic function of an action from its specific key binding during training. We preserve the semantics of actions in the original trajectory data but randomly replace their corresponding key bindings. For instance, in a trajectory, we might randomly replace the original \texttt{W} key used for "move forward" with the \texttt{X} key, while simultaneously defining the function of \texttt{X} as "move forward" in the System Prompt. Through this random substitution, we compel the model to rely on the System Prompt to understand the currently valid action space and its semantics, rather than merely memorizing and reproducing high-frequency action patterns from the pre-training data. Furthermore, this method effectively mitigates the problem of highly imbalanced action distributions in the training data, preventing the model from gaming the loss function by exploiting this prior distribution.

\paragraph{Inverse-Dynamics Prediction}
To further reinforce the model's understanding of action semantics, we introduce Inverse-Dynamics Prediction as an auxiliary training task. This task requires the model to infer the action $a_t$ that caused a state transition, given the current observation $o_t$ and the next observation $o_{t+1}$. This is formally represented as:
\begin{equation}
    a_t \sim P(\cdot \mid A, [o_i, a_i]_{i=1}^{t-1}, o_t, o_{t+1}),
\end{equation}
where $A$ is the predefined action space described in natural language. This task compels the model to learn the causal relationship between actions and changes in the environmental state, thereby fostering a deeper understanding of the actual effects of actions. Finally, throughout the entire post-training phase, we provide explicit task instructions (Instruction, $I$) in the System Prompt for each video trajectory. This shifts the model's prediction target from simple imitation learning to conditional instruction following, with the final training objective expressed as:
$$ P(a_t \mid I, A, o_t). $$

\subsection{Multimodal Prompts Elicit In-Context Learning}
Another core insight of the post-training stage is that the agent should be able to learn from historical interactions and dynamically adjust its behavioral strategies. This is crucial for enhancing its generalization capabilities in unseen games. We define this ability as the agent's In-Context Learning (ICL)~\citep{jiang2022vima,dong2022survey}: the agent dynamically adjusts the distribution of its future actions, including the selection of discrete actions and the scaling of continuous actions, by observing the effects of previous actions on the environment.

\paragraph{Contextual Understanding of Discrete Actions via Multimodal Prompts}
We found that relying solely on language-based \texttt{System Prompts} is often insufficient to accurately and comprehensively express the semantics of all actions. In complex action games, the function of certain keys (e.g., special skills or combos) is highly context-dependent and can even vary between characters. Such complex semantics are difficult to describe exhaustively with text. To address this, we introduce a multimodal prompting mechanism. For each game, we select video clips demonstrating key "action-effect" pairs to provide as in-context examples for the agent. By observing these examples, the agent can intuitively understand the actual effect of specific actions in that environment, rather than relying on textual system prompts. During training, we even randomly mask the textual action descriptions, $A$, in the \texttt{System Prompt}, forcing the agent to infer the action space from the multimodal context.

\paragraph{Adaptive Calibration of Continuous Actions}
For continuous actions, such as mouse movements, the practical significance of their numerical values varies dramatically across different environments. In-game mouse sensitivity settings, differences in user hardware, and personalized key bindings can all cause the same \texttt{(dx,dy)} values to produce vastly different in-game responses. To enable the agent to adapt to this dynamism, we adopt the following strategies:

\begin{itemize}
\item \textbf{Long-Horizon, Non-Markovian Trajectory Training:} We use non-Markovian trajectory data for autoregressive training and extend the length of a single trajectory to at least 80 consecutive frames. This ensures that the agent has a sufficiently long history of interactions to reference for calibration when making decisions.
\item \textbf{Initial Action Loss Masking:} To prevent the gradient direction from being distorted by uncalibrated actions at the beginning of a trajectory (due to a lack of historical context), we mask the action loss for the first $T$ mask steps, excluding them from gradient propagation.
\item \textbf{Continuous Action Data Augmentation:} We apply random scaling to the mouse movements in the trajectories. For each training segment, we multiply all \texttt{(dx,dy)} values by a random scaling factor $k$, while ensuring their absolute values do not exceed a cap of $180$. This forces the agent to learn to infer the correct scaling factor from the feedback of historical actions, rather than relying on a fixed numerical range.
\end{itemize}

Finally, to further enhance the model's ability to learn long-term dependencies, our training data also includes a subset of ultra-long sequences with trajectories exceeding 240 steps.

\subsection{Sparse Thinking via Reinforcement Fine-Tuning}
During the post-training phase, we observed that both the reasoning data synthesized offline by LLMs and that recorded online by humans exhibit a degree of abstraction, preventing their content from precisely guiding subsequent action generation. Furthermore, the "think-aloud" data collection mechanism led some players to pause game actions while verbalizing their thoughts, resulting in a large volume of reasoning text being associated with "\texttt{no-op}" actions. This decoupling of reasoning and action weakens the model's ability to learn the causal relationship between them.

\paragraph{Locating the sparse thinking Steps}
Not all steps within a trajectory require explicit reasoning for the agent to act correctly. To optimize the fine-tuning process, we first identify the timestamps where reasoning is most critical. We employ a system prompt to let \gametars generate the actions only, $\pi_{action}$, which is required to predict the next action based solely on the history of observations and actions. We then evaluate \gametars on the offline trajectories. A timestep $t$ is identified as requiring reasoning augmentation if the action-only model fails to predict the ground-truth action $a_t^{gt}$. This suggests that a simple reactive policy is insufficient at this stage, and a more deliberative reasoning process is required. This procedure yields a set of timestamps $S_r$ targeted for reasoning augmentation:
\begin{equation}
S_r = { t \in {1, \dots, T} \mid I(\pi_{\text{action}}(a \mid o_t, (o_i, a_i)_{i=0}^{t-1}) = a_t^{\text{gt}}) = 0 }
\end{equation}
Where $I(\cdot)$ is the indicator function used to identify unmatched generated actions. The subsequent Rejection Fine-Tuning is then applied exclusively to the timestamps within this set $S_r$.

\paragraph{Refining Thought-Action Chains via Rejection Sampling}
To strengthen the causal link between reasoning and action, we adopted a Rejection Fine-Tuning (RFT) approach to refine and enhance the offline trajectory data~\citep{jaech2024openai}. We use the actions from the original trajectories as a critic to filter and generate high-quality reasoning-action pairs through Rejection Sampling.

Specifically, for any given timestep $t$ in a trajectory $\tau$, we have the agent model $\pi$ generate a candidate thought $r_t^\text{cand}$ based on the observation $o_t$, and then predict an action $a_t^\text{pred}$ based on this thought. We compare the predicted action with the ground-truth action from the human demonstration, $a_t^{gt}$. The candidate thought $r_t^\text{cand}$ is considered valid and retained for subsequent fine-tuning only if the predicted action matches the ground-truth action. This process can be formally described as:
\begin{equation}
    \text{Accept } r_t^{\text{cand}} \text{ iff } I(\pi(a_t \mid o_t, r_t^{\text{cand}}) = a_t^{\text{gt}}) = 1
\end{equation}
where $I(\cdot)$ is the indicator function. Through this method, we filter out a significant amount of abstract reasoning that is irrelevant to the actual behavior, retaining only those thought chains that directly lead to correct actions.

After obtaining high-quality reasoning data through rejection sampling, we perform two post-processing steps:
\begin{itemize}
\item \textbf{Thought Consolidation:} We found that in segments where the same action is performed consecutively, the effective reasoning content also shows a high degree of repetitiveness. Therefore, we merge these consecutive, identical thoughts, retaining only the instance at the first timestep to reduce data redundancy.
\item \textbf{LLM Rewriting and Density Control:} We further utilize an LLM to rewrite the filtered reasoning texts, making their expression more concise and their logic clearer. Concurrently, we actively control the density of reasoning in the trajectories (i.e., the proportion of steps with reasoning to the total number of steps) to better align with a natural human cognitive rhythm.
\end{itemize}

\begin{figure*}[t]
    \centering
    \includegraphics[width=0.95\linewidth]{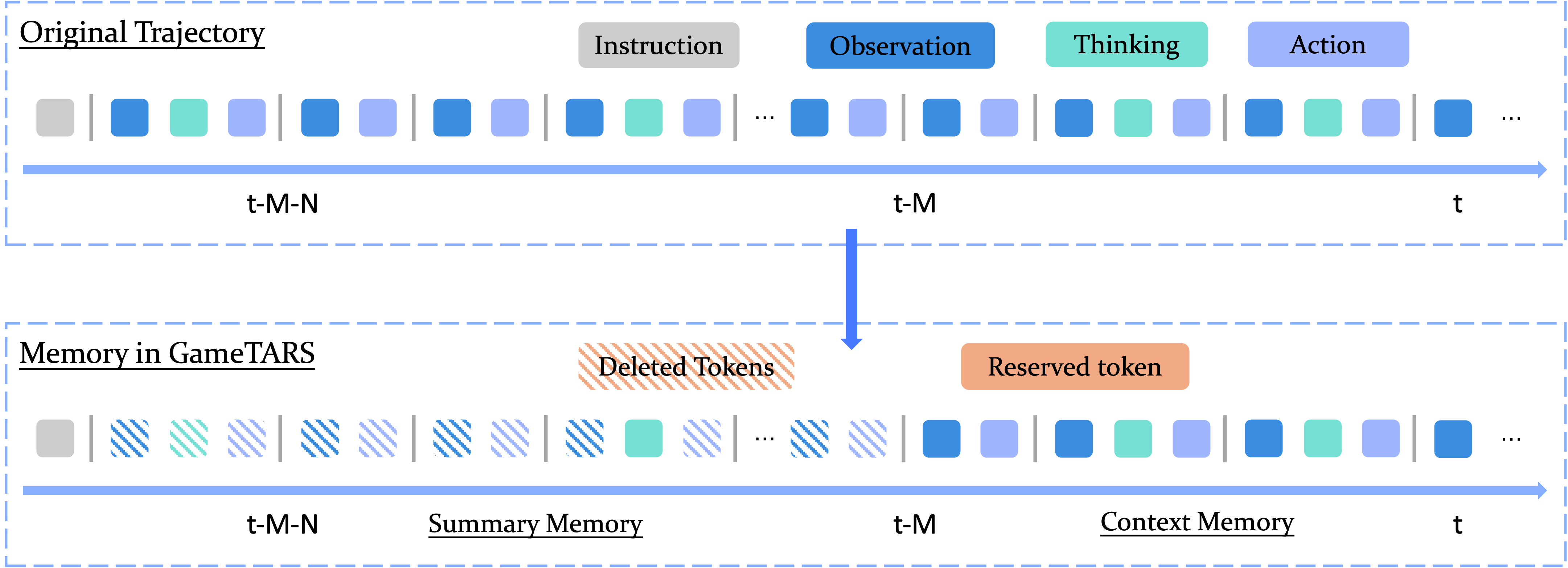}
    \caption{
    \textbf{Memory Mechanism in \gametars Agent} is divided into two areas: context and summary, corresponding to long-term and short-term memory, respectively. The context memory retains a complete record of observations, thoughts, and actions for $M$ steps. In contrast, the summary memory only retains the thinking content for $N$ steps. Older information (beyond $t-M-N$) is popped using a sliding window approach.
    }
    \label{fig:memory}
\end{figure*}

\subsection{Long Episodic Memory Compression with Sparse Thought}
To succeed in complex tasks requiring long-range planning and memory, an agent must be able to integrate historical information that far exceeds the standard context window length. To this end, we have designed a two-tiered memory mechanism that remains consistent during both training and inference, effectively balancing high-fidelity recent memory with highly compressed long-term memory.

\paragraph{Tier 1: Short-Term Contextual Memory}
This tier serves as the agent's "working memory" and is stored directly within the model's context window. It maintains an interleaved sequence in the form of a queue, containing the most recent observations and their corresponding ReAct pairs, refined through the "sparse thinking" process. This sequence is updated using a sliding window, with a standard context length of 32k tokens. However, because encoded visual images consume a substantial number of tokens, the temporal span of interaction history that this window can accommodate is limited.

\paragraph{Tier 2: Long-Term Summary Memory}
When historical information is moved out of the short-term context window, it is compressed and stored in the second memory tier. To overcome the storage bottleneck imposed by visual tokens, we leverage the output of the "sparse" mechanism—sparse and highly refined thought texts—as summaries of historical events. Each step that is evicted from the context no longer retains its large visual observation data; instead, only the corresponding thought text is preserved. Thanks to the sparsity of thoughts and the high compression ratio of textual representation, this method allows the agent to retain a long-term task context of over 2000 interaction steps at an extremely low token cost.

\paragraph{Ultra-Long Context Training}
To further enhance the agent's ability to process and utilize long-range dependencies, we have included ultra-long sequences in our training data, with context window lengths reaching up to 128k tokens. By training on data of this scale, the agent learns to reason and plan over much broader time horizons, thereby improving its performance on tasks that require long-term memory.

\subsection{Post-Training with Cross-Source Agentic Trajectories}
Our ultimate goal is to build a generalist agent whose capabilities extend beyond the domain of gaming. Previous research has shown that incorporating diverse agentic trajectory data from various domains into the training process can significantly improve a model's general cognitive abilities, such as reflection, planning, and decision-making~\citep{wang2025ui}. Therefore, in the post-training stage, we not only use massive amounts of game data but also introduce a range of cross-source agentic trajectories, with the aim of cultivating a more comprehensive generalist agent.

We posit that the visual understanding and rapid-response capabilities learned through continual pre-training in game environments provide the agent with a solid "sensory-motor" foundation. Building on this, we introduce several types of non-gaming agent data to strengthen its higher-level cognitive functions:

\begin{itemize}
\item \textbf{Code Agent Trajectories:} This data comes from agents performing programming tasks and includes the entire process from problem comprehension and logical decomposition to code implementation. By learning from these trajectories, the agent can enhance its structured thinking and long-range logical reasoning abilities~\citep{qwen3coder,SWEverifiedintro,yang2024sweagent}.
\item \textbf{GUI Agent Trajectories:} This data covers the process of completing specific tasks in desktop or web applications (e.g., booking a flight, managing files). It strengthens the agent's ability to understand and execute multi-step instructions in non-gaming environments~\citep{seed2025uitars15,openai_2025_cua_blog}.
\item \textbf{MCP Agent Trajectories:} This data records an agent's process of information retrieval, document reading, content synthesis, and summarization to complete complex research tasks. Learning from this data can significantly enhance the agent's ability to plan, reflect, and integrate knowledge within the ReAct loop~\citep{openai-deepresearch,feng2025retool,thinking1.6,seed2025seedthinking,wang2025actingreasoningmoreteaching}.
\end{itemize}

By co-training on these diverse data sources alongside game data, we aim to achieve a synergistic learning effect: game data solidifies the foundational interaction skills, while cross-domain agentic data builds more abstract and general problem-solving capabilities on top of this foundation. This process ultimately facilitates the model's transformation from a generalist game agent into a general computer-use agent.

\section{Experiment}

\subsection{Training Details}

\begin{table}[]
\centering
\caption{Training recipes in different stages of GameTARS agent.}
\label{tab:training_recipes}
\resizebox{0.7\textwidth}{!}{%
\begin{tabular}{@{}lclc@{}}
\toprule
Stage & Pre-Training &  & Post-Training \\ \midrule
Image Sizes & (640,360)$\sim$(1920,1080) &  & (640,360)$\sim$(1280,720) \\
Image Tokens & 208B &  & 10B \\
Text Tokens & 326B &  & 29B\\
Total Tokens & 526B &  & 40B \\
Game Number & $\sim$100 &  & $\sim$500 \\
Game Hours & $\sim20k$ & & $\sim2k$ \\
Freeze ViT & N & & Y \\
Maximum Learning Rate & 2e-5 &  & 1.4e-5 \\
Minimum Learning Rate & 2e-6 &  & 2e-6 \\
Batch Size & 2048 &  & 512 \\
Maximum Token Sequence Length & 32,000 &  & 128,000 \\
Maximum Image Sequence Length & 80 &  & 300 \\
Training Steps & 16,000 &  & 10,000 \\
Warmup Steps & 1000 &  & 750 \\ \bottomrule
\end{tabular}%
}
\end{table}

The training process for our agent comprises two main phases: continual pre-training and post-training (see \autoref{tab:training_recipes} for detailed training parameters). 
During the early stages of this project, we explored a multi-stage training strategy, consisting of the following steps: 1) training on pure action trajectories; 2) continuing training on offline-synthesized ReAct trajectories; 3) further training on online-collected human "think-aloud" ReAct trajectories; and 4) fine-tuning with general multimodal data to enable capability annealing. However, our experimental results indicated that this staged approach had a detrimental effect on the model's final performance. As a result, we adopted a unified, single-stage continual pre-training method, blending all data sources together, and trained the model on over 500 billion tokens. Subsequently, during the post-training phase, we fine-tuned the model on approximately 20 billion carefully curated high-quality tokens.

\paragraph{Model Details}
We trained three versions of \gametars, namely \gametars-MoE-mini, \gametars-MoE-Large, and \gametars-Dense.
\gametars-MoE-mini is initialized from the pre-trained checkpoint of Seed-VL-1.5~\citep{guo2025seed1}, and leverages all of its
continual-training data. The architecture includes a 632M-parameter vision encoder and a Mixture-of-Experts (MoE) LLM with 2.5B active parameters. \gametars-MoE-large is a larger version of MoE.
To ensure a fair comparison with other open-source models, we trained \gametars-Dense using the Qwen2.5-VL-7B-Instruct checkpoint~\citep{bai2025qwen2} and the VeOmni training strategies~\citep{ma2025veomni}. 
Due to training and data limitations, we only used game data for continual pretraining of the dense \gametars.

\paragraph{Dataset Details}
Our training dataset represents an unprecedented scale, both in terms of breadth and depth. The game data includes over 500 popular games covering a wide range of genres. For detailed statistics on the types, durations, and diversity of the game data, please refer to \autoref{fig:dataset_distribution}. The overall token distribution across different data sources is as follows:
\begin{enumerate}
    \item \textbf{Pure Action Game Trajectories}: 61 billion (61B) tokens across 50 open-world games.
    \item \textbf{Offline-Synthesized ReAct Trajectories}: 45 billion (45B) tokens across 50 games. These data and the pure-action section use initial game data from different sources.
    \item \textbf{Online Human "Think-Aloud" ReAct Trajectories}: 127 billion (127B) tokens across 500 games. We selected games with a focus on reasoning in this part of the data and enhanced the diversity of the games.
    \item \textbf{Cross-Domain Agent Data}: over 20 billion (20B) tokens including deep research, codex, and GUI trajectories~\citep{SWEverifiedintro,xie2024osworld,wei2025browsecomp}.
    \item \textbf{General Multimodal Data}: 200 billion (200B) tokens, including question-answering, image captioning, visual question answering (VQA), and video captioning~\citep{seed2025seed1,guo2025seed1}.
\end{enumerate}

\subsection{Main Results}

\subsubsection{Experimental Setups}

To comprehensively evaluate the generalization capabilities of our agent model, we focused on testing its performance in games and environments that were not encountered during the training phase. We carefully selected a set of representative benchmarks to assess the agent's adaptability and performance across a range of tasks and environments. The selected environments include:

\begin{enumerate}
    \item \textbf{Minecraft} Although the model was exposed to a large amount of Minecraft data during training, the open-ended nature and task diversity of this environment make it an ideal platform for testing the model's ability to perform new tasks in a familiar setting. We utilized a variety of tasks from Minedojo~\citep{minedojo} and MCU~\citep{mcu} to ensure the model was tested on a broad spectrum of challenges, including task completion efficiency, flexibility, and reasoning capabilities in this open-world environment. 
    \item \textbf{Web Games} To ensure the novelty of the evaluation environment, we selected multiple games from the online platform Poki for testing. These games span across genres such as racing, puzzle, and other action types, providing an effective means of testing the model's zero-shot generalization ability in entirely new visual and interaction paradigms. This evaluation serves to measure the model's reasoning and decision-making performance in previously unseen games and environments.
    \item \textbf{Vizdoom} Vizdoom is a semi-realistic 3D environment based on the classic game Doom, commonly used to evaluate an agent’s decision-making ability in fast-paced \textbf{FPS} scenarios that require rapid responses. By testing in this dynamic environment, we can assess the agent's ability to react quickly to rapidly changing situations and evaluate its decision-making speed and accuracy under competitive, high-stakes conditions~\citep{vizdoom}.
    \item \textbf{MiniWorld} MiniWorld is a minimalist 3D environment simulator designed to test an agent’s navigation, object interaction, and basic physical reasoning skills. In this environment, the agent is required to perform basic tasks such as finding target objects, avoiding obstacles, and handling physical simulations. It serves as an effective testbed for evaluating the model's ability to manage simple physical and interactive tasks~\citep{miniworld}.
\end{enumerate}

These benchmarks encompass a variety of game types, task complexities, and interaction modalities, ensuring a comprehensive evaluation of the agent’s generalization ability in new, unseen environments. By testing across these diverse setups, we aim to verify the model's performance in unfamiliar settings, providing insights into its potential for general-purpose intelligence.

\subsubsection{Evaluation on Open-World Minecraft}

We first conducted a systematic evaluation of the agent's instruction-following capability on unseen tasks within the Minecraft environment. For this purpose, we adopted the MCU benchmark~\citep{mcu}, which contains more than 800 diverse and easy-to-verify tasks. By evaluating on this large and heterogeneous task set, we can quantitatively assess the agent's ability to understand novel instructions and generalize them into concrete action sequences. To ensure fairness, we manually curated 800+ distinct initial environments such that each task could be completed within 600 steps. We further controlled the random seed to guarantee that the agent had not been exposed to these environments during training. Each task was paired with a unique natural language instruction that guided the agent's behavior. Performance was measured using Average Success Rate and Average Completion Steps for evaluating instruction-following ability. As baselines, we compared against agents trained exclusively on Minecraft, including both policy-based and VLM-based approaches.

To provide a more fine-grained analysis, we grouped the tasks into three categories: (1) Embodied tasks, which evaluate exploration and object interaction in 3D space; (2) GUI tasks, which assess crafting and manipulation in 2D interfaces; and (3) Combat tasks, which focus on survival and combat capabilities. Each task was evaluated at least three times to reduce randomness in results.

\begin{table}[t]
\centering
\caption{
\textbf{Evaluation results in open-world Minecraft.}
The dense model of GameTARS is trained exclusively on game datasets, without incorporating datasets from other agents trajectories or general visual question-answering.
}
\label{tab:minecraft_results}
\resizebox{0.99\textwidth}{!}{%
\renewcommand\arraystretch{1.1}
\begin{tabular}{@{}lcccccccccccccc@{}}
\toprule
\multirow{2}{*}{Agents} & \multicolumn{4}{c}{Embodied Tasks} &  & \multicolumn{4}{c}{GUI Tasks} &  & \multicolumn{4}{c}{Combat Tasks} \\ \cmidrule(lr){2-5} \cmidrule(lr){7-10} \cmidrule(l){12-15} 
 & \includegraphics[scale=0.03]{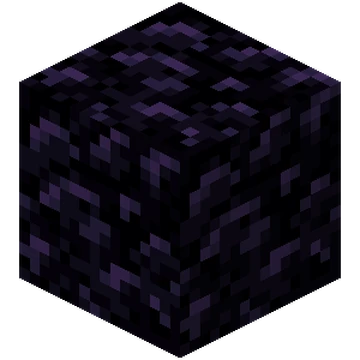} & \includegraphics[scale=0.04]{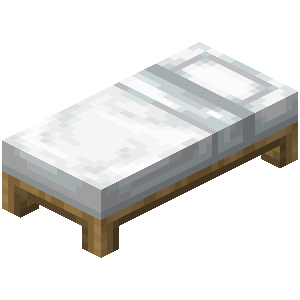} & Steps & ASR &  & \includegraphics[scale=0.08]{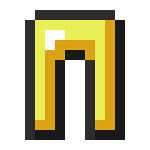} & \includegraphics[scale=0.04]{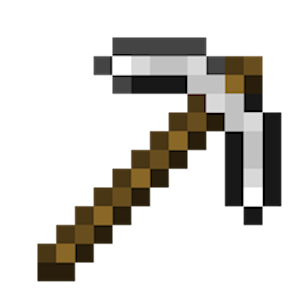} & Steps & ASR &  & \includegraphics[scale=0.015]{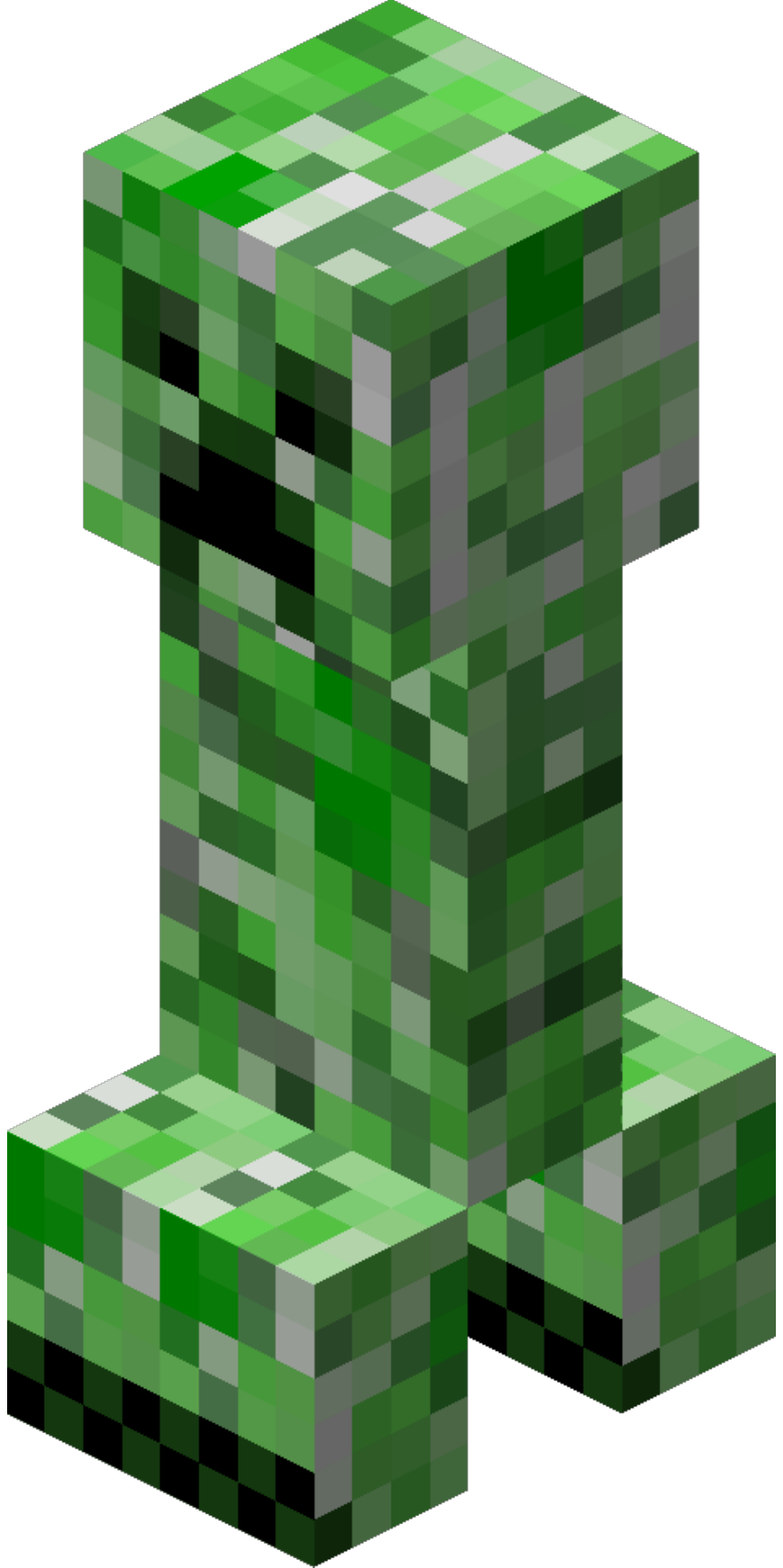} & \includegraphics[scale=0.2]{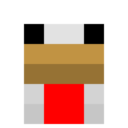} & Steps & ASR \\ \midrule
\multicolumn{15}{l}{\cellcolor[HTML]{F0F4FF}\textsc{Policy-based Agents in Minecraft}} \\
VPT~\citep{vpt} & $.00$ & $.00$ & $377$ & $6.0^{\pm11.4}$ &  & $.00$ & $.20$ & $398$ & $0.8^{\pm3.3}$ &  & $.00$ & $.15$ & $396$ & $3.6^{\pm7.7}$ \\
STEVE-1~\citep{lifshitz2023steve} & $.10$ & $.25$ & $384$ & $8.0^{\pm17.0}$ &  & $.00$ & $.30$ & $391$ & $3.2^{\pm8.4}$ &  & $.00$ & $.20$ & $395$ & $3.9^{\pm 12.0}$ \\
Dreamer V3~\citep{dreamerv3} & $.00$ & $.00$ & $382$ & $3.0^{\pm 10.2}$ &  & - & - & - & - &  & $.00$ & $.05$ & $370$ & $3.1^{\pm11.4}$ \\ \midrule
\multicolumn{15}{l}{\cellcolor[HTML]{F0FBEF}\textsc{VLM-based Agents in Minecraft}} \\
JARVIS-VLA~\citep{jarvisvla} & $.25$ & $.80$ & $305$ & $30.0^{\pm35.4}$ &  & $.35$ & $.45$ & $339$ & $25.1^{\pm23.9}$ &  & $.00$ & $.20$ & $352$ & $18.5^{\pm22.7}$ \\
OmniJARVIS~\citep{wang2024omnijarvis} & $.00$ & $.25$ & $363$ & $24.4^{\pm31.1}$ &  & $.00$ & $.00$ & $393$ & $3.0^{\pm 7.5}$ &  & $.00$ & $.20$ & $371$ & $8.5^{\pm17.9}$ \\
OpenHA~\citep{openha} & $.10$ & $.45$ & $287$ & $30.1^{\pm 13.9}$ &  & $.25$ & $.30$ & $314$ & $32.5^{\pm9.2}$ &  & $.00$ & $.30$ & $316$ & {\color[HTML]{245BDB} $31.9^{\pm13.7}$} \\
UI-TARS-1.5~\citep{seed2025uitars15} & $.30$ & $.60$ & $290$ & {\color[HTML]{245BDB} $42.1^{\pm 20.4}$} &  & $.30$ & $.50$ & $320$ & {\color[HTML]{245BDB} $36.7^{\pm 17.2}$} &  & $.00$ & $.60$ & $346$ & $31.0^{\pm 16.4}$ \\ \midrule
\multicolumn{15}{l}{\cellcolor[HTML]{FEF1F1}\textsc{Our Generalist Game Agent}} \\
GameTARS-Dense$^*$ & $.35$ & $.85$ & $373$ & $50.4^{\pm20.7}$ &  & $.40$ & $.55$ & $406$ & $39.1^{\pm 27.5}$ &  & $.10$ & $.70$ & $372$ & $38.1^{\pm24.6}$ \\
GameTARS-MoE-mini & $.70$ & $.85$ & $188$ & $72.0^{\pm 21.5}$ &  & $.65$ & $.55$ & $288$ & $55.4^{\pm 17.6}$ &  & $.40$ & $.75$ & $271$ & $66.1^{\pm 25.2}$ \\
\bottomrule
\end{tabular}}
\end{table}

As shown in Table~\ref{tab:minecraft_results}, our generalist agent \gametars substantially outperforms all expert baselines trained solely on Minecraft. 
In particular, the \gametars-MoE-mini variant achieved the highest success rates across all task categories, with significant improvements in both instruction-following accuracy and sample efficiency. These results confirm that large-scale continual pretraining with a unified action space enables strong generalization even in complex, open-world environments.

A closer inspection of the results highlights several key findings. First, \gametars exhibits particularly strong performance in GUI tasks, where the success rate reached 55.0\%, surpassing all prior baselines by a large margin. This demonstrates the effectiveness of the unified action space in handling tasks that require reasoning and execution across both 2D graphical interfaces and 3D embodied environments. Second, in embodied tasks, \gametars achieved a remarkable 72.0\% success rate while completing tasks with substantially fewer steps, underscoring its ability to generalize complex exploration and interaction behaviors. Finally, in combat tasks, the model achieved 66.0\% success, more than doubling the performance of the strongest VLM-based baseline.

Taken together, these results validate our central hypothesis: large-scale continual pretraining with a unified, device-level action space not only boosts overall task success but also enhances the agent’s adaptability to diverse instruction modalities. \gametars therefore establishes a new state-of-the-art in open-world Minecraft benchmarks and provides compelling evidence of the viability of generalist game agents.

\subsubsection{Evaluation on Unseen Benchmarks and Simulators}

\begin{figure}[htp]
    \centering
    \begin{minipage}[b]{0.95\textwidth} 
        \centering
        \includegraphics[width=0.95\textwidth]{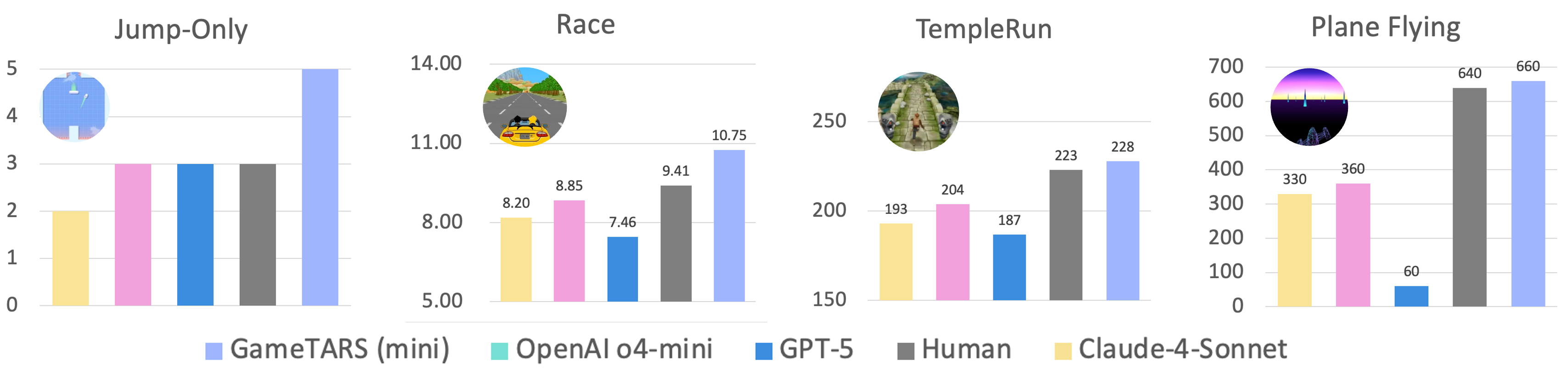} 
        \caption{
        \textbf{Evaluation results in Web 3D games on Poki.} 
    Evaluation results in Web 3D games on Poki. Metrics include checkpoint time (Race), levels completed (Jump-Only), short-horizon score (Temple Run), and survival score (Airplane Flying).
        }
        \label{fig:poki_eval}
    \end{minipage}
    \vspace{10pt}  
    
    \begin{minipage}[b]{0.95\textwidth}  
        \centering
        \includegraphics[width=0.95\textwidth]{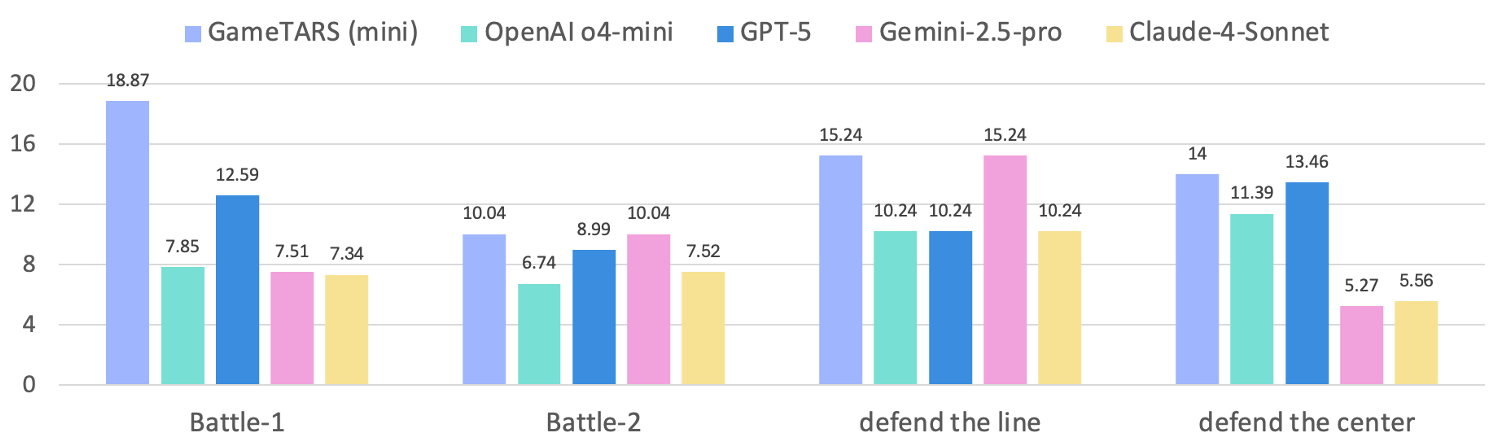}
        \caption{
        \textbf{Evaluation results in FPS Vizdoom benchmark across four maps~\citep{vizdoom}.}
     The reported metric is the average episode reward. \gametars-mini significantly outperforms all baselines.
        }
        \label{fig:doom_eval}
    \end{minipage}
    \vspace{10pt}
    
    \begin{minipage}[b]{0.95\textwidth}  
        \centering
        \includegraphics[width=0.95\textwidth]{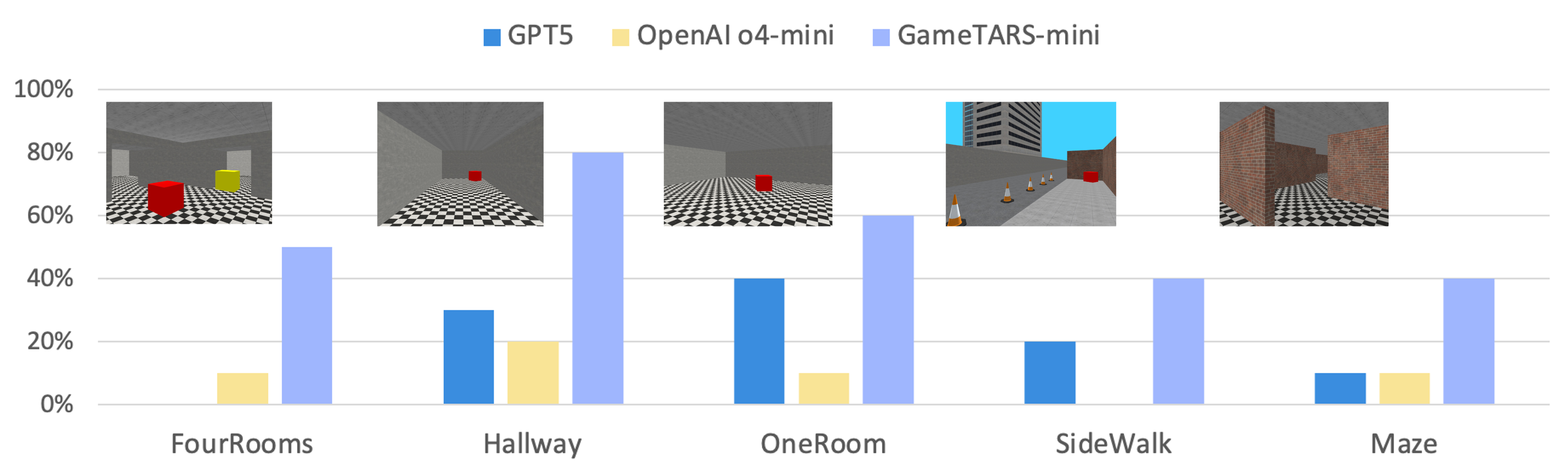}
        \caption{
        \textbf{Evaluation results in Miniworld Simulator~\citep{miniworld}.}
        }
        \label{fig:miniworld_eval}
    \end{minipage}
\end{figure}

The goal of \gametars is not to achieve a higher score on trained games, but to provide a good foundation model for a generalist game agent. Therefore, further evaluation is conducted on unseen games, including web games, FPS Doom, and 3D simulators.
We first evaluated our model in browser-based 3D web games to test its zero-shot generalization ability in entirely novel visual and interaction paradigms. We selected four representative games from the Poki platform\footnote{https://poki.com/}: Race (racing), Jump-Only (action), Temple Run (endless running), and Airplane Flying (3D navigation). Evaluation metrics were task-specific: in Race, we measured the time to reach the first checkpoint (lower is better); in Jump-Only, the number of levels completed within 20 seconds; in Temple Run, the score achieved within 20 seconds; and in Airplane Flying, the cumulative score before failure. 

All games were wrapped into a browser sandbox environment to ensure a controlled and reproducible evaluation\footnote{Due to the inference latency of the model, we packaged these web games into a sandbox and used time-pause to ensure causal interaction.}. We used GPT-5 and Human players as baselines. For the human baseline, we recruited at least four individuals with no prior experience in the selected games and reported their best scores. For GPT-5 and our agent, we conducted at least five runs per game and reported the best-of-5 (Bo5) results. Both models were provided with identical system prompts, including basic game rules and action space definitions.

As illustrated in Figure~\ref{fig:poki_eval}, \gametars-mini demonstrated competitive generalization compared to human players, achieving near-human or superior performance across most games. Notably, the model outperformed fresh human players in Jump-Only and Airplane Flying, highlighting its ability to adapt to novel tasks and mechanics without task-specific training.

We further evaluated our agent in Vizdoom~\citep{vizdoom}, a classic FPS environment commonly used for benchmarking decision-making under fast-paced and adversarial conditions. We selected four standard maps: Battle-1, Battle-2, Defend the Line, and Defend the Center. The evaluation metric was the cumulative reward achieved per episode. Baselines included leading general-purpose models such as GPT-5~\citep{gpt5intro}, Claude-4-Sonnet~\citep{claude4intro}, and Gemini-2.5-Pro~\citep{comanici2025gemini}.

Results in Figure~\ref{fig:doom_eval} show that \gametars-mini consistently outperformed all baselines across almost all maps, achieving substantial gains in both survival and combat effectiveness. Beyond raw scores, qualitative analysis revealed that the agent exhibited advanced FPS behaviors, including effective corner-peeking, evasive maneuvers to dodge projectiles, and adaptive targeting strategies. These results underscore the strong transferability of behaviors learned from diverse game data to previously unseen FPS environments.

To further assess the agent’s ability to generalize to tasks requiring navigation, memory, and exploration, we evaluated it in the MiniWorld 3D simulator. This environment presents simplified but diverse tasks involving spatial reasoning, obstacle avoidance, and object interaction. Across nearly all tested maps, \gametars demonstrated robust performance, successfully navigating new environments and completing tasks despite having no prior exposure to MiniWorld during training.

Together, these results confirm the broad generalization capability of \gametars across unseen environments, ranging from browser-based games and FPS simulators to navigation-focused 3D environments. The findings provide strong evidence that large-scale continual pretraining with a unified action space enables the emergence of transferable skills applicable far beyond the training distribution.

\begin{figure*}[t]
    \centering
    \includegraphics[width=0.95\linewidth]{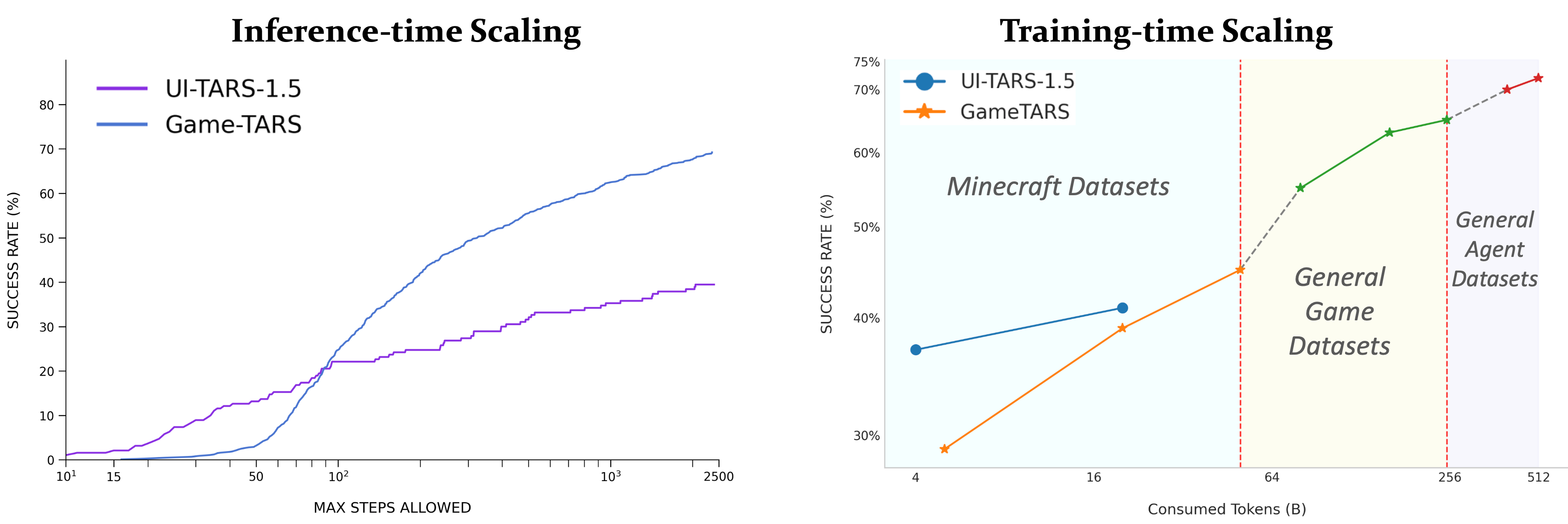}
    \caption{
    \textbf{(Left) Inference-time Scaling Experiments.} \gametars can improve the task success rates through exploration in inference-time steps.
    \textbf{(Right) Training-time Scaling Experiments.} The performance of \gametars improves as the training data increases, including in-domain, out-of-domain, and other agentic trajectory tokens.
    }
    \label{fig:training_scaling}
\end{figure*}

\subsection{Analysis and Discussions}

In this section, we present a set of analyses to provide deeper insights into the design and performance of \gametars. Our discussion covers three main aspects: sparse thinking, loss decay across training paradigms, and scaling experiments. These analyses aim to ensure a fair and systematic evaluation of the model’s capabilities and limitations.

\subsubsection{Scaling Experiments on Training Datasets and Inference Steps}

We first conducted scaling experiments on the continual pre-training datasets to examine how performance improves with increasing amounts and diversity of training data. The results are summarized in Figure~\ref{fig:training_scaling}, and several key insights emerge:

\begin{enumerate}
    \item Specified action space performs well with small datasets but is hard to scale. When using visual grounding–based actions (e.g., GUI action spaces), agents achieve relatively strong performance with limited training data. However, such action spaces require extensive preprocessing of raw trajectories and do not naturally transfer across heterogeneous games. As a result, their scalability and ceiling performance are inherently constrained ({\color{blue}blue curve}).

    \item Unified action space requires more data initially but scales better. By contrast, the proposed unified device-level action space is highly scalable but not natively supported by pre-trained VLMs. As a result, its performance lags behind GUI-based approaches when trained on small datasets. However, after pre-training on large-scale game data, its upper-bound performance approaches that of GUI-based action spaces ({\color{orange}orange curve}).

    \item Cross-game scaling brings substantial gains. The unified action space enables consistent representation across diverse games, allowing large-scale aggregation of heterogeneous data. Incorporating 200B additional tokens from other games led to a significant improvement in Minecraft performance, boosting the success rate from 42\% to 65\%. This observation echoes the \textbf{Bitter Lesson} in AI: general and simple representations, when combined with scaling in data and compute, yield stable and long-term improvements~\citep{sutton2019bitter}.

    \item Multimodal and cross-domain agentics data further enhance generalization. Extending pre-training to include multimodal data (e.g., captioning, VQA, video understanding) and agent trajectories from other domains led to an additional improvement, pushing performance to 72\%. This finding demonstrates that generalist agent models can benefit from heterogeneous sources of action–trajectory data, acquiring broadly transferable decision-making capabilities.
\end{enumerate}

Together, these results confirm the scalability advantages of the unified action space: while GUI-based representations may provide a short-term advantage in low-data regimes, only the unified paradigm enables sustained performance gains through cross-game and cross-domain scaling.

\subsubsection{Ablation on sparse thinking }

\begin{table}[]
\centering
\caption{Ablation experiments on different thinking patterns in GameTARS across different evaluation benchmarks.}
\label{tab:thinking_ablation}
\resizebox{0.55\textwidth}{!}{%
\begin{tabular}{@{}llllll@{}}
\toprule
Method & Minecraft & Doom & Race $\downarrow$ & 2048 & Avg. Tokens $\downarrow$ \\ \midrule
No-thinking & 55\% & 13.2 & 12.4s & 445 & 22 \\
Greedy Thinking & 45\% & 7.2 & 11.3s & 986 & 68 \\
Efficient Thinking & 63\% & 11.5 & 9.3s & 924 & 37 \\ \bottomrule
\end{tabular}%
}
\end{table}

To better understand the role of reasoning in decision-making, we conducted ablation experiments on different thinking patterns, including No-thinking, Greedy Thinking, and sparse thinking. The results are reported in Table~\ref{tab:thinking_ablation}.

\textbf{The necessity of explicit reasoning is strongly task-dependent}. In reasoning-intensive games such as 2048, where planning and long-horizon inference are critical, incorporating thinking at every step significantly improves performance (986 average tokens consumed per step). By contrast, in fast-paced environments such as Doom, excessive reasoning (Greedy Thinking) introduces latency and even degrades performance. One primary source of degradation is hallucinated reasoning loops, where the agent incorrectly assumes that certain goals have already been achieved and prematurely terminates actions.

\textbf{The No-thinking baseline performs relatively well in reaction-heavy games like Doom} but fails to generalize to tasks that require compositional reasoning (e.g., crafting tasks in Minecraft or the strategic planning required in 2048). On the other hand, Greedy Thinking—forcing the agent to reason at every step—incurs substantial computational overhead, increases inference cost, and amplifies the risk of cascading reasoning errors.

Our proposed sparse thinking strategy achieves the best trade-off between performance and inference efficiency. The agent learns to selectively engage in reasoning at key decision points (e.g., planning in Minecraft crafting tasks or assessing critical survival situations in Doom), while re-using or following previously established reasoning in routine steps. This reduces redundant computation, lowers average token consumption (37 tokens/step vs. 68 tokens/step in Greedy Thinking), and achieves state-of-the-art performance across most benchmarks (63\% success rate in Minecraft, 924 points in 2048).

Interestingly, qualitative trajectory analysis shows that the agent develops an implicit meta-cognition mechanism: it chooses when to reason and when to act reflexively. For instance, in Minecraft crafting sequences, the agent reasons carefully at the start of a task (to identify needed resources and steps) but executes subsequent mining or crafting actions without redundant re-planning. Similarly, in 2048, it reasons at every move due to the inherent combinatorial structure of the game, while in VizDoom it minimizes reasoning to maintain responsiveness.

These findings highlight that reasoning is not universally beneficial; rather, its effectiveness depends on the interaction between task structure and agent inference cost. Sparse thinking emerges as a general strategy that balances adaptability, efficiency, and robustness, enabling the agent to approximate human-like decision-making where deliberation is applied selectively at critical junctures.

\begin{table}[]
\centering
\caption{
\textbf{Ablation experiments on the validation of decaying loss.}
The ASR and N-Acc represents average success rates on online evaluations and non-repetitive accuracy on offline datasets, respectively.
}
\label{tab:ablation_loss}
\resizebox{\textwidth}{!}{%
\renewcommand\arraystretch{1.1}
\begin{tabular}{@{}lllllllllllllll@{}}
\toprule
\multirow{2}{*}{Model} & \multicolumn{3}{l}{Online Embodied Tasks} &  & \multicolumn{3}{l}{Online Combat Tasks} &  & \multicolumn{3}{l}{Online GUI Tasks} &  & \multicolumn{2}{l}{Offline Evaluation} \\ \cmidrule(lr){2-4} \cmidrule(lr){6-8} \cmidrule(lr){10-12} \cmidrule(l){14-15} 
 & ASR & Diversity & Steps &  & ASR & Diversity & Steps &  & ASR & Diversity & Steps &  & Accuracy & N-Acc \\ \midrule
w/o Decaying  Loss & 44.90\% & 60\% & 168 &  & 14.54\% & 50\% & 305 &  & 37.75\% & 80\% & 244 &  & 59\% & 12\% \\
w/ Decaying Loss & 46.15\% & 90\% & 172 &  & 22.22\% & 60\% & 247 &  & 42.26\% & 100\% & 164 &  & 47\% & 39\% \\ \bottomrule
\end{tabular}%
}
\end{table}

\subsubsection{Validation on Continual Pre-Training Loss}

We further investigated the impact of applying a decaying continual pre-training loss. The analysis was conducted from both training dynamics and downstream evaluation perspectives (Table~\ref{tab:ablation_loss}).

From a purely optimization perspective, introducing a decaying loss initially leads to a slightly higher absolute training loss compared to the non-decaying baseline. Nevertheless, the overall decreasing trend remains stable, and both approaches converge to comparable final loss values. This observation indicates that the decaying strategy does not compromise convergence stability, while potentially reshaping the optimization trajectory.

We then conducted ablation studies on downstream online environments, measuring task-level Average Success Rate (ASR), trajectory diversity, and average completion steps. The results reveal several advantages of the decaying loss: 1) Improved task success rates across embodied, combat, and GUI tasks, with particularly large gains in combat scenarios (from 14.5\% to 22.2\%). 2) Substantially shorter trajectories, indicating that agents are less likely to become stuck or generate redundant action loops. 3) Enhanced behavioral diversity, especially in GUI and embodied tasks, where diversity increased from 60–80\% to nearly 100\%.

These improvements suggest that decaying loss helps mitigate the tendency of agents to fall into repetitive behaviors, thereby producing more robust and efficient action sequences.

To further probe this effect, we performed offline action prediction experiments across 1,000 trajectories containing more than 100k steps. We report two metrics: 1) Average action prediction accuracy across all steps. 2) Non-repetitive accuracy, i.e., accuracy on steps where the predicted action differs from the previous one.

Interestingly, while the non-decaying baseline achieves higher overall accuracy (59\% vs. 47\%), the decaying loss yields a dramatic improvement in non-repetitive accuracy (39\% vs. 12\%, a +28\% absolute gain). This indicates that models trained without decaying loss tend to exploit dataset bias by repeatedly copying the previous action, “hacking” the accuracy metric without genuinely improving decision quality. In contrast, decaying loss discourages such degenerate strategies, forcing the agent to learn meaningful state-dependent action prediction.

The ablation results demonstrate that although decaying loss sacrifices some global prediction accuracy, it provides substantial benefits in downstream performance by reducing repetition, increasing action diversity, and improving robustness across online tasks. This aligns with the broader goal of continual pre-training: to shape learning signals in a way that promotes adaptive, generalizable behaviors rather than exploiting dataset artifacts.

\section{Related Works}

Our work builds upon and extends several key research areas in artificial intelligence: specialized game-playing agents, the emerging paradigm of generalist agents, and agents designed for broader computer interaction tasks. We situate \gametars by drawing a critical distinction based on the agent's action space, arguing that a unified, human-friendly paradigm is fundamental to achieving true scalability and generality.

\subsection{Game Agents}
The history of AI is deeply intertwined with games as benchmarks for intelligence. Foundational work in closed-world games demonstrated superhuman performance but relied on highly specialized action spaces. Deep Q-Networks (DQN) mastered numerous Atari games from raw pixels using a low-dimensional, discrete action space~\citep{dqn}. AlphaGo and its successors achieved world-champion level Go play via deep neural networks and Monte Carlo Tree Search, with actions tied to the game's formal rules~\citep{alphago}. Similarly, AlphaStar excelled in StarCraft II using a custom, game-specific API, limiting transferability ~\citep{alphastar}. While highly successful, these agents' tight coupling of action representation to specific environments remains a bottleneck for generalization.

The open-world environment of Minecraft has become a key testbed for more generalizable agents. VPT pioneered learning behavioral priors from large-scale video data using the native human interface (keyboard/mouse), aligning with our approach, though training was confined to Minecraft~\citep{vpt}. Other methods like DreamerV3 learn world models via RL, solving hard-exploration tasks without human data but lacking demonstrated cross-domain generalization~\citep{dreamerv3}.

The rise of pre-trained vision-language models (VLMs) and large language models (LLMs) has enabled new possibilities for multi-task game agents in open worlds. Voyager~\citep{wang2023voyager} leverages LLMs for long-horizon planning but executes actions via high-level APIs (e.g., JavaScript code), creating a semantic gap by learning to sequence functions rather than acquiring sensorimotor skills~\citep{wang2023describe,wang2024jarvis,li2024optimus,yuan2023skill,proagent}. Cradle demonstrated strong instruction-following across multiple games using a unified agentic workflow~\citep{cradle}. Other approaches achieve strong performance via VLM-based fine-tuning (SFT) or RL within specific games~\citep{zhai2024fine,driess2023palm,wang2024omnijarvis,seed2025uitars15,mcreward}. Furthermore, general AI models like Gemini-2.5~\citep{comanici2025gemini}, GPT-5~\citep{gpt5intro}, and Claude 4 \citep{claude4intro} are increasingly evaluated on games (e.g., Chess, Pokémon), highlighting gaming as a crucial benchmark for foundational model capabilities \citep{zhang2025videogamebench,shi2025korgym}.

\subsection{Generalist Agents}
The pursuit of a single agent capable of mastering multiple domains has led to the development of generalist architectures. Early DQN research showed that a unified RL algorithm could achieve superhuman performance across many Atari games, albeit requiring separate training runs for each~\citep{dqn}. Gato was a seminal work demonstrating a multi-modal, multi-task, multi-embodiment agent trained with a single set of weights by serializing all data into a token sequence for a Transformer~\citep{gato}. However, its heterogeneous, discretized action space still required domain-specific tokenization, falling short of a truly universal interface. More recently, the SIMA project shares a core philosophy with \gametars: training an agent to follow natural language instructions across diverse 3D environments using a generic, human-like interface (pixels in, keyboard/mouse out)~\citep{sima}. 
This marks a critical shift toward a unified action paradigm. However, initial SIMA agents were evaluated primarily on short-horizon tasks and limited in-domain games, underscoring the learning challenge. 
VPT built a foundational model for reinforcement learning in an open world by pre-training on IDM and video data from the internet. However, this foundational model can only be used for Minecraft and is difficult to generalize to other game scenarios~\citep{vpt}.
Recently, with the rise of pre-trained models such as LLM~\citep{qwen3technicalreport,jaech2024openai,meta2025llama} and VLM~\citep{claude4intro,gpt5intro,comanici2025gemini}, developing a pre-trained foundation model for agents has become mainstream, such as embodied foundation models~\citep{team2025gemini,nvidia2025gr00t,palme,pi0.5_2025,cheang2025gr} and computer-use foundation models~\citep{openai_2025_cua_blog,wang2025ui,wang2025opencua,kimiteam2025kimik2openagentic}.
\gametars builds on this shared vision, demonstrating enhanced performance and scalability through large-scale continual pre-training on vision language models.

\subsection{Broader AI Agents} 
From a broader perspective, there is significant potential for AI agents based on VLM or LLM to be implemented in various fields, including computer use and embodied agents~\citep{openx,rth,brohan2023rt,openvla,palme}.
Research into GUI and Web Agents (e.g., using Mind2Web~\citep{deng2023mind2web} or UI-TARS \citep{wang2025ui}) often focuses on grounding high-level plans to specific UI elements~\citep{li2025screenspot,hong2024cogagent,wu2024atlas,xu2024aguvis,lu2025arpo,wang2025actingreasoningmoreteaching,luo2025gui}. 
By operating directly on low-level keyboard/mouse outputs, \gametars learns an end-to-end policy that implicitly addresses grounding. 
Similarly, agents for software engineering like Claude~\citep{claude4intro} and SWE-agent~\citep{yang2024sweagent,SWEverifiedintro,jimenez2023swebench} tackle complex, logic-driven tasks. 
Protocols such as MCP~\citep{Anthropic2024MCP} have introduced standardized mechanisms for flexible tool integration, enabling agents to seamlessly interact with search engines, file parsers, or external APIs~\citep{feng2025retool,li2025torlscalingtoolintegratedrl,wang2024rat,song2025r1}.
Incorporating data from such domains in post-training enhances \gametars' high-level reasoning, complementing its foundational sensorimotor skills.

\section{Conclusions}
We present \gametars, a generalist game agent utilizing a unified, scalable action space based on native keyboard-mouse inputs. This human-native paradigm bypasses the limitations of API/GUI approaches, enabling large-scale continual pre-training (>500B tokens) across diverse domains (game trajectories, ai agents trajectories, multimodal data). \gametars demonstrates strong instruction-following capabilities and employs an efficient Sparse Thinking strategy, balancing reasoning depth with inference cost. Experiments show \gametars achieves ~2x SOTA success rate in Minecraft, nears human-level generalization in unseen web games, and outperforms leading models (GPT-5, Gemini-2.5-Pro, Claude-4-Sonnet) on FPS benchmarks. Scaling results confirm that simple, scalable action representations paired with massive pre-training offer a promising path toward generalist computer-use agents.

\section*{Contributions}

\textbf{Authors} Zihao Wang$^{1,2}$, Xujing Li$^1$, Yining Ye$^1$, Junjie Fang$^1$, Haoming Wang$^1$, Longxiang Liu$^1$, Shihao Liang$^1$, Junting Lu$^1$, Zhiyong Wu$^1$, Jiazhan Feng$^1$, Wanjun Zhong$^1$, Zili Li$^1$, Yu Wang$^1$, Yu Miao$^1$, Bo Zhou$^1$, Yuanfan Li$^1$, Hao Wang$^1$, Zhongkai Zhao$^1$, Faming Wu$^1$, Zhengxuan Jiang$^3$, Weihao Tan$^1$, Heyuan Yao$^1$, Shi Yan$^1$, Xiangyang Li$^1$, Yitao Liang$^2$, Yujia Qin$^1$, Guang Shi$^1$

\textbf{Afflications} $^1$ByteDance Seed, $^2$Peking University, $^3$M-A-P

\textbf{Acknowledgment} We greatly thank Muyao Li for discussions. And we appreciate Cheng Yuan and Hao Chen's efforts on the \gametars website.

\clearpage

\bibliographystyle{plainnat}
\bibliography{main}

\begin{thebibliography}{87}
\providecommand{\natexlab}[1]{#1}
\providecommand{\url}[1]{\texttt{#1}}
\expandafter\ifx\csname urlstyle\endcsname\relax
  \providecommand{\doi}[1]{doi: #1}\else
  \providecommand{\doi}{doi: \begingroup \urlstyle{rm}\Url}\fi

\bibitem[{Anthropic}(2024)]{Anthropic2024MCP}
{Anthropic}.
\newblock Introducing the model context protocol, 2024.
\newblock URL \url{https://www.anthropic.com/news/model-context-protocol}.

\bibitem[anthropic(2025)]{claude4intro}
anthropic.
\newblock Introducing claude 4, 2025.
\newblock URL \url{https://www.anthropic.com/news/claude-4}.

\bibitem[Bai et~al.(2025)Bai, Chen, Liu, Wang, Ge, Song, Dang, Wang, Wang, Tang, et~al.]{bai2025qwen2}
Shuai Bai, Keqin Chen, Xuejing Liu, Jialin Wang, Wenbin Ge, Sibo Song, Kai Dang, Peng Wang, Shijie Wang, Jun Tang, et~al.
\newblock Qwen2. 5-vl technical report.
\newblock \emph{arXiv preprint arXiv:2502.13923}, 2025.

\bibitem[Baker et~al.(2022)Baker, Akkaya, Zhokov, Huizinga, Tang, Ecoffet, Houghton, Sampedro, and Clune]{vpt}
Bowen Baker, Ilge Akkaya, Peter Zhokov, Joost Huizinga, Jie Tang, Adrien Ecoffet, Brandon Houghton, Raul Sampedro, and Jeff Clune.
\newblock Video pretraining (vpt): Learning to act by watching unlabeled online videos.
\newblock \emph{Advances in Neural Information Processing Systems}, 35:\penalty0 24639--24654, 2022.

\bibitem[Belkhale et~al.(2024)Belkhale, Ding, Xiao, Sermanet, Vuong, Tompson, Chebotar, Dwibedi, and Sadigh]{rth}
Suneel Belkhale, Tianli Ding, Ted Xiao, Pierre Sermanet, Quon Vuong, Jonathan Tompson, Yevgen Chebotar, Debidatta Dwibedi, and Dorsa Sadigh.
\newblock Rt-h: Action hierarchies using language.
\newblock \emph{arXiv preprint arXiv:2403.01823}, 2024.

\bibitem[Brohan et~al.(2023)Brohan, Brown, Carbajal, Chebotar, Chen, Choromanski, Ding, Driess, Dubey, Finn, et~al.]{brohan2023rt}
Anthony Brohan, Noah Brown, Justice Carbajal, Yevgen Chebotar, Xi~Chen, Krzysztof Choromanski, Tianli Ding, Danny Driess, Avinava Dubey, Chelsea Finn, et~al.
\newblock Rt-2: Vision-language-action models transfer web knowledge to robotic control.
\newblock \emph{arXiv preprint arXiv:2307.15818}, 2023.

\bibitem[Bruce et~al.(2024)Bruce, Dennis, Edwards, Parker-Holder, Shi, Hughes, Lai, Mavalankar, Steigerwald, Apps, et~al.]{bruce2024genie}
Jake Bruce, Michael~D Dennis, Ashley Edwards, Jack Parker-Holder, Yuge Shi, Edward Hughes, Matthew Lai, Aditi Mavalankar, Richie Steigerwald, Chris Apps, et~al.
\newblock Genie: Generative interactive environments.
\newblock In \emph{Forty-first International Conference on Machine Learning}, 2024.

\bibitem[ByteDance(2025)]{thinking1.6}
ByteDance.
\newblock Seed-thinking-1.6, 2025.
\newblock URL \url{https://seed.bytedance.com/zh/seed1_6}.

\bibitem[Cheang et~al.(2025)Cheang, Chen, Cui, Hu, Huang, Kong, Li, Li, Liu, Ma, et~al.]{cheang2025gr}
Chilam Cheang, Sijin Chen, Zhongren Cui, Yingdong Hu, Liqun Huang, Tao Kong, Hang Li, Yifeng Li, Yuxiao Liu, Xiao Ma, et~al.
\newblock Gr-3 technical report.
\newblock \emph{arXiv preprint arXiv:2507.15493}, 2025.

\bibitem[Cheng et~al.(2024)Cheng, Zhang, Zhang, Meng, Hong, Li, Wang, Wang, Yin, Zhao, et~al.]{cheng2024exploring}
Yuheng Cheng, Ceyao Zhang, Zhengwen Zhang, Xiangrui Meng, Sirui Hong, Wenhao Li, Zihao Wang, Zekai Wang, Feng Yin, Junhua Zhao, et~al.
\newblock Exploring large language model based intelligent agents: Definitions, methods, and prospects.
\newblock \emph{arXiv preprint arXiv:2401.03428}, 2024.

\bibitem[Chevalier-Boisvert et~al.(2023)Chevalier-Boisvert, Dai, Towers, de~Lazcano, Willems, Lahlou, Pal, Castro, and Terry]{miniworld}
Maxime Chevalier-Boisvert, Bolun Dai, Mark Towers, Rodrigo de~Lazcano, Lucas Willems, Salem Lahlou, Suman Pal, Pablo~Samuel Castro, and Jordan Terry.
\newblock Minigrid \& miniworld: Modular \& customizable reinforcement learning environments for goal-oriented tasks.
\newblock \emph{CoRR}, abs/2306.13831, 2023.

\bibitem[Comanici et~al.(2025)Comanici, Bieber, Schaekermann, Pasupat, Sachdeva, Dhillon, Blistein, Ram, Zhang, Rosen, et~al.]{comanici2025gemini}
Gheorghe Comanici, Eric Bieber, Mike Schaekermann, Ice Pasupat, Noveen Sachdeva, Inderjit Dhillon, Marcel Blistein, Ori Ram, Dan Zhang, Evan Rosen, et~al.
\newblock Gemini 2.5: Pushing the frontier with advanced reasoning, multimodality, long context, and next generation agentic capabilities.
\newblock \emph{arXiv preprint arXiv:2507.06261}, 2025.

\bibitem[De~Haan et~al.(2019)De~Haan, Jayaraman, and Levine]{de2019causal}
Pim De~Haan, Dinesh Jayaraman, and Sergey Levine.
\newblock Causal confusion in imitation learning.
\newblock \emph{Advances in neural information processing systems}, 32, 2019.

\bibitem[Deitke et~al.(2024)Deitke, Clark, Lee, Tripathi, Yang, Park, Salehi, Muennighoff, Lo, Soldaini, et~al.]{deitke2024molmo}
Matt Deitke, Christopher Clark, Sangho Lee, Rohun Tripathi, Yue Yang, Jae~Sung Park, Mohammadreza Salehi, Niklas Muennighoff, Kyle Lo, Luca Soldaini, et~al.
\newblock Molmo and pixmo: Open weights and open data for state-of-the-art multimodal models.
\newblock \emph{arXiv e-prints}, pages arXiv--2409, 2024.

\bibitem[Deng et~al.(2023)Deng, Guu, Pasupat, Aky{\"u}rek, Zhuang, Chen, Hashimoto, Guu, and Liang]{deng2023mind2web}
Xiang Deng, Kelvin Guu, Panupong Pasupat, Afra Aky{\"u}rek, Sheng Zhuang, Wenlong Chen, Tatsunori Hashimoto, Kelvin Guu, and Percy Liang.
\newblock Mind2web: Towards a generalist agent for the web.
\newblock In \emph{NeurIPS Datasets and Benchmarks}, 2023.
\newblock URL \url{https://arxiv.org/abs/2306.06070}.

\bibitem[Dong et~al.(2022)Dong, Li, Dai, Zheng, Ma, Li, Xia, Xu, Wu, Liu, et~al.]{dong2022survey}
Qingxiu Dong, Lei Li, Damai Dai, Ce~Zheng, Jingyuan Ma, Rui Li, Heming Xia, Jingjing Xu, Zhiyong Wu, Tianyu Liu, et~al.
\newblock A survey on in-context learning.
\newblock \emph{arXiv preprint arXiv:2301.00234}, 2022.

\bibitem[Driess et~al.(2023{\natexlab{a}})Driess, Xia, Sajjadi, Lynch, Chowdhery, Ichter, Wahid, Tompson, Vuong, Yu, et~al.]{palme}
Danny Driess, Fei Xia, Mehdi~SM Sajjadi, Corey Lynch, Aakanksha Chowdhery, Brian Ichter, Ayzaan Wahid, Jonathan Tompson, Quan Vuong, Tianhe Yu, et~al.
\newblock Palm-e: An embodied multimodal language model.
\newblock \emph{arXiv preprint arXiv:2303.03378}, 2023{\natexlab{a}}.

\bibitem[Driess et~al.(2023{\natexlab{b}})Driess, Xia, Sajjadi, Lynch, Chowdhery, Wahid, Tompson, Vuong, Yu, Huang, et~al.]{driess2023palm}
Danny Driess, Fei Xia, Mehdi~SM Sajjadi, Corey Lynch, Aakanksha Chowdhery, Ayzaan Wahid, Jonathan Tompson, Quan Vuong, Tianhe Yu, Wenlong Huang, et~al.
\newblock Palm-e: An embodied multimodal language model.
\newblock 2023{\natexlab{b}}.

\bibitem[Fan et~al.(2022)Fan, Wang, Jiang, Mandlekar, Yang, Zhu, Tang, Huang, Zhu, and Anandkumar]{minedojo}
Linxi Fan, Guanzhi Wang, Yunfan Jiang, Ajay Mandlekar, Yuncong Yang, Haoyi Zhu, Andrew Tang, De-An Huang, Yuke Zhu, and Anima Anandkumar.
\newblock Minedojo: Building open-ended embodied agents with internet-scale knowledge.
\newblock \emph{Advances in Neural Information Processing Systems}, 35:\penalty0 18343--18362, 2022.

\bibitem[Feng et~al.(2025)Feng, Huang, Qu, Zhang, Qin, Zhong, Jiang, Chi, and Zhong]{feng2025retool}
Jiazhan Feng, Shijue Huang, Xingwei Qu, Ge~Zhang, Yujia Qin, Baoquan Zhong, Chengquan Jiang, Jinxin Chi, and Wanjun Zhong.
\newblock Retool: Reinforcement learning for strategic tool use in llms.
\newblock \emph{arXiv preprint arXiv:2504.11536}, 2025.

\bibitem[Guo et~al.(2025)Guo, Wu, Zhu, Leng, Shi, Chen, Fan, Wang, Jiang, Wang, et~al.]{guo2025seed1}
Dong Guo, Faming Wu, Feida Zhu, Fuxing Leng, Guang Shi, Haobin Chen, Haoqi Fan, Jian Wang, Jianyu Jiang, Jiawei Wang, et~al.
\newblock Seed1. 5-vl technical report.
\newblock \emph{arXiv preprint arXiv:2505.07062}, 2025.

\bibitem[Hafner et~al.(2023)Hafner, Pasukonis, Ba, and Lillicrap]{dreamerv3}
Danijar Hafner, Jurgis Pasukonis, Jimmy Ba, and Timothy Lillicrap.
\newblock Mastering diverse domains through world models.
\newblock \emph{arXiv preprint arXiv:2301.04104}, 2023.

\bibitem[Hong et~al.(2024)Hong, Wang, Lv, Xu, Yu, Ji, Wang, Wang, Dong, Ding, et~al.]{hong2024cogagent}
Wenyi Hong, Weihan Wang, Qingsong Lv, Jiazheng Xu, Wenmeng Yu, Junhui Ji, Yan Wang, Zihan Wang, Yuxiao Dong, Ming Ding, et~al.
\newblock Cogagent: A visual language model for gui agents.
\newblock In \emph{Proceedings of the IEEE/CVF Conference on Computer Vision and Pattern Recognition}, pages 14281--14290, 2024.

\bibitem[Intelligence et~al.(2025)Intelligence, Black, Brown, Darpinian, Dhabalia, Driess, Esmail, Equi, Finn, Fusai, Galliker, Ghosh, Groom, Hausman, Ichter, Jakubczak, Jones, Ke, LeBlanc, Levine, Li-Bell, Mothukuri, Nair, Pertsch, Ren, Shi, Smith, Springenberg, Stachowicz, Tanner, Vuong, Walke, Walling, Wang, Yu, and Zhilinsky]{pi0.5_2025}
Physical Intelligence, Kevin Black, Noah Brown, James Darpinian, Karan Dhabalia, Danny Driess, Adnan Esmail, Michael Equi, Chelsea Finn, Niccolo Fusai, Manuel~Y. Galliker, Dibya Ghosh, Lachy Groom, Karol Hausman, Brian Ichter, Szymon Jakubczak, Tim Jones, Liyiming Ke, Devin LeBlanc, Sergey Levine, Adrian Li-Bell, Mohith Mothukuri, Suraj Nair, Karl Pertsch, Allen~Z. Ren, Lucy~Xiaoyang Shi, Laura Smith, Jost~Tobias Springenberg, Kyle Stachowicz, James Tanner, Quan Vuong, Homer Walke, Anna Walling, Haohuan Wang, Lili Yu, and Ury Zhilinsky.
\newblock {$\pi_{0.5}$}: A vision-language-action model with open-world generalization.
\newblock \emph{arXiv preprint arXiv:2504.16054}, 2025.
\newblock URL \url{https://arxiv.org/abs/2504.16054}.

\bibitem[Jaech et~al.(2024)Jaech, Kalai, Lerer, Richardson, El-Kishky, Low, Helyar, Madry, Beutel, Carney, et~al.]{jaech2024openai}
Aaron Jaech, Adam Kalai, Adam Lerer, Adam Richardson, Ahmed El-Kishky, Aiden Low, Alec Helyar, Aleksander Madry, Alex Beutel, Alex Carney, et~al.
\newblock Openai o1 system card.
\newblock \emph{arXiv preprint arXiv:2412.16720}, 2024.

\bibitem[Jiang et~al.(2022)Jiang, Gupta, Zhang, Wang, Dou, Chen, Fei-Fei, Anandkumar, Zhu, and Fan]{jiang2022vima}
Yunfan Jiang, Agrim Gupta, Zichen Zhang, Guanzhi Wang, Yongqiang Dou, Yanjun Chen, Li~Fei-Fei, Anima Anandkumar, Yuke Zhu, and Linxi Fan.
\newblock Vima: General robot manipulation with multimodal prompts.
\newblock \emph{arXiv preprint arXiv:2210.03094}, 2\penalty0 (3):\penalty0 6, 2022.

\bibitem[Jimenez et~al.(2023)Jimenez, Yang, Wettig, Yao, Pei, Press, and Narasimhan]{jimenez2023swebench}
Carlos~E Jimenez, John Yang, Alexander Wettig, Shunyu Yao, Kexin Pei, Ofir Press, and Karthik Narasimhan.
\newblock {SWE}-bench: Can language models resolve real-world github issues?
\newblock \emph{arXiv preprint arXiv:2310.06770}, 2023.

\bibitem[Kim et~al.(2025)Kim, Pertsch, Karamcheti, Xiao, Balakrishna, Nair, Rafailov, Foster, Lam, Sanketi, Vuong, Kollar, Burchfiel, Tedrake, Sadigh, Levine, Liang, and Finn]{openvla}
Moo~Jin Kim, Karl Pertsch, Siddharth Karamcheti, Ted Xiao, Ashwin Balakrishna, Suraj Nair, Rafael Rafailov, Ethan~P Foster, Grace Lam, Pannag~R Sanketi, Quan Vuong, Thomas Kollar, Benjamin Burchfiel, Russ Tedrake, Dorsa Sadigh, Sergey Levine, Percy Liang, and Chelsea Finn.
\newblock Openvla: An open-source vision-language-action model.
\newblock In Pulkit Agrawal, Oliver Kroemer, and Wolfram Burgard, editors, \emph{Proceedings of The 8th Conference on Robot Learning}, volume 270 of \emph{Proceedings of Machine Learning Research}, pages 2679--2713. PMLR, 06--09 Nov 2025.
\newblock URL \url{https://proceedings.mlr.press/v270/kim25c.html}.

\bibitem[Li et~al.(2024{\natexlab{a}})Li, Yang, Wang, Zhu, Zhou, Qiao, Wang, Li, Lu, and Dai]{mcreward}
Hao Li, Xue Yang, Zhaokai Wang, Xizhou Zhu, Jie Zhou, Yu~Qiao, Xiaogang Wang, Hongsheng Li, Lewei Lu, and Jifeng Dai.
\newblock Auto mc-reward: Automated dense reward design with large language models for minecraft.
\newblock In \emph{Proceedings of the IEEE/CVF Conference on Computer Vision and Pattern Recognition (CVPR)}, pages 16426--16435, June 2024{\natexlab{a}}.

\bibitem[Li et~al.(2025{\natexlab{a}})Li, Meng, Lin, Luo, Tian, Ma, Huang, and Chua]{li2025screenspot}
Kaixin Li, Ziyang Meng, Hongzhan Lin, Ziyang Luo, Yuchen Tian, Jing Ma, Zhiyong Huang, and Tat-Seng Chua.
\newblock Screenspot-pro: Gui grounding for professional high-resolution computer use.
\newblock \emph{arXiv preprint arXiv:2504.07981}, 2025{\natexlab{a}}.

\bibitem[Li et~al.(2025{\natexlab{b}})Li, Wang, He, Ma, and Liang]{jarvisvla}
Muyao Li, Zihao Wang, Kaichen He, Xiaojian Ma, and Yitao Liang.
\newblock Jarvis-vla: Post-training large-scale vision language models to play visual games with keyboards and mouse.
\newblock \emph{arXiv preprint arXiv:2503.16365}, 2025{\natexlab{b}}.

\bibitem[Li et~al.(2025{\natexlab{c}})Li, Zou, and Liu]{li2025torlscalingtoolintegratedrl}
Xuefeng Li, Haoyang Zou, and Pengfei Liu.
\newblock Torl: Scaling tool-integrated rl, 2025{\natexlab{c}}.
\newblock URL \url{https://arxiv.org/abs/2503.23383}.

\bibitem[Li et~al.(2024{\natexlab{b}})Li, Xie, Shao, Chen, Jiang, and Nie]{li2024optimus}
Zaijing Li, Yuquan Xie, Rui Shao, Gongwei Chen, Dongmei Jiang, and Liqiang Nie.
\newblock Optimus-1: Hybrid multimodal memory empowered agents excel in long-horizon tasks.
\newblock \emph{Advances in neural information processing systems}, 37:\penalty0 49881--49913, 2024{\natexlab{b}}.

\bibitem[Lifshitz et~al.(2023)Lifshitz, Paster, Chan, Ba, and McIlraith]{lifshitz2023steve}
Shalev Lifshitz, Keiran Paster, Harris Chan, Jimmy Ba, and Sheila McIlraith.
\newblock Steve-1: A generative model for text-to-behavior in minecraft.
\newblock \emph{Advances in Neural Information Processing Systems}, 36:\penalty0 69900--69929, 2023.

\bibitem[Lin et~al.(2023)Lin, Wang, Ma, and Liang]{mcu}
Haowei Lin, Zihao Wang, Jianzhu Ma, and Yitao Liang.
\newblock Mcu: A task-centric framework for open-ended agent evaluation in minecraft.
\newblock \emph{arXiv preprint arXiv:2310.08367}, 2023.

\bibitem[Lu et~al.(2025)Lu, Zhong, Liu, Fu, and Jia]{lu2025arpo}
Fanbin Lu, Zhisheng Zhong, Shu Liu, Chi-Wing Fu, and Jiaya Jia.
\newblock Arpo: End-to-end policy optimization for gui agents with experience replay.
\newblock \emph{arXiv preprint arXiv:2505.16282}, 2025.

\bibitem[Luo et~al.(2025)Luo, Wang, He, Chen, Li, and Xia]{luo2025gui}
Run Luo, Lu~Wang, Wanwei He, Longze Chen, Jiaming Li, and Xiaobo Xia.
\newblock Gui-r1: A generalist r1-style vision-language action model for gui agents.
\newblock \emph{arXiv preprint arXiv:2504.10458}, 2025.

\bibitem[Ma et~al.(2025)Ma, Zheng, Shi, Zhao, Jia, Huang, Lin, Li, Yang, Peng, et~al.]{ma2025veomni}
Qianli Ma, Yaowei Zheng, Zhelun Shi, Zhongkai Zhao, Bin Jia, Ziyue Huang, Zhiqi Lin, Youjie Li, Jiacheng Yang, Yanghua Peng, et~al.
\newblock Veomni: Scaling any modality model training with model-centric distributed recipe zoo.
\newblock \emph{arXiv preprint arXiv:2508.02317}, 2025.

\bibitem[Meta(2025)]{meta2025llama}
AI~Meta.
\newblock The llama 4 herd: The beginning of a new era of natively multimodal ai innovation.
\newblock \emph{https://ai.meta.com/blog/llama-4-multimodal-intelligence/}, 2025.

\bibitem[Mnih et~al.(2015)Mnih, Kavukcuoglu, Silver, Rusu, Veness, Bellemare, Graves, Riedmiller, Fidjeland, Ostrovski, et~al.]{dqn}
Volodymyr Mnih, Koray Kavukcuoglu, David Silver, Andrei~A Rusu, Joel Veness, Marc~G Bellemare, Alex Graves, Martin Riedmiller, Andreas~K Fidjeland, Georg Ostrovski, et~al.
\newblock Human-level control through deep reinforcement learning.
\newblock \emph{nature}, 518\penalty0 (7540):\penalty0 529--533, 2015.

\bibitem[Morris et~al.(2023)Morris, Sohl-Dickstein, Fiedel, Warkentin, Dafoe, Faust, Farabet, and Legg]{morris2023levels}
Meredith~Ringel Morris, Jascha Sohl-Dickstein, Noah Fiedel, Tris Warkentin, Allan Dafoe, Aleksandra Faust, Clement Farabet, and Shane Legg.
\newblock Levels of agi for operationalizing progress on the path to agi.
\newblock \emph{arXiv preprint arXiv:2311.02462}, 2023.

\bibitem[NVIDIA et~al.(2025)NVIDIA, :, Bjorck, Castañeda, Cherniadev, Da, Ding, Fan, Fang, Fox, Hu, Huang, Jang, Jiang, Kautz, Kundalia, Lao, Li, Lin, Lin, Liu, Llontop, Magne, Mandlekar, Narayan, Nasiriany, Reed, Tan, Wang, Wang, Wang, Wang, Xiang, Xie, Xu, Xu, Ye, Yu, Zhang, Zhang, Zhao, Zheng, and Zhu]{nvidia2025gr00t}
NVIDIA, :, Johan Bjorck, Fernando Castañeda, Nikita Cherniadev, Xingye Da, Runyu Ding, Linxi~"Jim" Fan, Yu~Fang, Dieter Fox, Fengyuan Hu, Spencer Huang, Joel Jang, Zhenyu Jiang, Jan Kautz, Kaushil Kundalia, Lawrence Lao, Zhiqi Li, Zongyu Lin, Kevin Lin, Guilin Liu, Edith Llontop, Loic Magne, Ajay Mandlekar, Avnish Narayan, Soroush Nasiriany, Scott Reed, You~Liang Tan, Guanzhi Wang, Zu~Wang, Jing Wang, Qi~Wang, Jiannan Xiang, Yuqi Xie, Yinzhen Xu, Zhenjia Xu, Seonghyeon Ye, Zhiding Yu, Ao~Zhang, Hao Zhang, Yizhou Zhao, Ruijie Zheng, and Yuke Zhu.
\newblock Gr00t n1: An open foundation model for generalist humanoid robots.
\newblock \emph{arXiv preprint arXiv: 2503.14734}, 2025.

\bibitem[O'Neill et~al.(2023)O'Neill, Rehman, Gupta, Maddukuri, Gupta, Padalkar, Lee, Pooley, Gupta, Mandlekar, et~al.]{openx}
Abby O'Neill, Abdul Rehman, Abhinav Gupta, Abhiram Maddukuri, Abhishek Gupta, Abhishek Padalkar, Abraham Lee, Acorn Pooley, Agrim Gupta, Ajay Mandlekar, et~al.
\newblock Open x-embodiment: Robotic learning datasets and rt-x models.
\newblock \emph{arXiv preprint arXiv:2310.08864}, 2023.

\bibitem[{OpenAI}(2023)]{openai2023gpt4v}
{OpenAI}.
\newblock Gpt-4v(ision) system card.
\newblock \url{https://openai.com/index/gpt-4v-system-card/}, 2023.
\newblock Accessed: 2025-04-23.

\bibitem[openai(2024)]{SWEverifiedintro}
openai.
\newblock Introducing swe-bench verified, 2024.
\newblock URL \url{https://openai.com/index/introducing-swe-bench-verified/}.

\bibitem[OpenAI(2025{\natexlab{a}})]{gpt5intro}
OpenAI.
\newblock Introducing gpt 5, 2025{\natexlab{a}}.
\newblock URL \url{https://openai.com/index/introducing-gpt-5/}.

\bibitem[OpenAI(2025{\natexlab{b}})]{openai-deepresearch}
OpenAI.
\newblock Introducing deep research - openai.
\newblock \url{https://openai.com/index/introducing-deep-research/}, 2025{\natexlab{b}}.

\bibitem[OpenAI(2025{\natexlab{c}})]{openai_2025_cua_blog}
OpenAI.
\newblock Computer-using agent (cua).
\newblock \url{https://openai.com/index/computer-using-agent/}, 2025{\natexlab{c}}.
\newblock Research preview / blog.

\bibitem[Qin et~al.(2025)Qin, Ye, Fang, Wang, Liang, Tian, Zhang, Li, Li, Huang, et~al.]{qin2025ui}
Yujia Qin, Yining Ye, Junjie Fang, Haoming Wang, Shihao Liang, Shizuo Tian, Junda Zhang, Jiahao Li, Yunxin Li, Shijue Huang, et~al.
\newblock Ui-tars: Pioneering automated gui interaction with native agents.
\newblock \emph{arXiv preprint arXiv:2501.12326}, 2025.

\bibitem[Qwen(2025)]{qwen3coder}
Qwen.
\newblock Qwen-3-coder.
\newblock \url{https://qwenlm.github.io/blog/qwen3-coder}, 2025.

\bibitem[Raad et~al.(2024)Raad, Ahuja, Barros, Besse, Bolt, Bolton, Brownfield, Buttimore, Cant, Chakera, et~al.]{sima}
Maria~Abi Raad, Arun Ahuja, Catarina Barros, Frederic Besse, Andrew Bolt, Adrian Bolton, Bethanie Brownfield, Gavin Buttimore, Max Cant, Sarah Chakera, et~al.
\newblock Scaling instructable agents across many simulated worlds.
\newblock \emph{arXiv preprint arXiv:2404.10179}, 2024.

\bibitem[Radford et~al.(2023)Radford, Kim, Xu, Brockman, McLeavey, and Sutskever]{radford2023robust}
Alec Radford, Jong~Wook Kim, Tao Xu, Greg Brockman, Christine McLeavey, and Ilya Sutskever.
\newblock Robust speech recognition via large-scale weak supervision.
\newblock In \emph{International conference on machine learning}, pages 28492--28518. PMLR, 2023.

\bibitem[Reed et~al.(2022)Reed, Zolna, Parisotto, Colmenarejo, Novikov, Barth-Maron, Gimenez, Sulsky, Kay, Springenberg, et~al.]{gato}
Scott Reed, Konrad Zolna, Emilio Parisotto, Sergio~Gomez Colmenarejo, Alexander Novikov, Gabriel Barth-Maron, Mai Gimenez, Yury Sulsky, Jackie Kay, Jost~Tobias Springenberg, et~al.
\newblock A generalist agent.
\newblock \emph{arXiv preprint arXiv:2205.06175}, 2022.

\bibitem[Seed(2025{\natexlab{a}})]{seed2025seedthinking}
Bytedance Seed.
\newblock Seed-thinking-v1. 5: Advancing superb reasoning models with reinforcement learning.
\newblock Technical report, Technical report, ByteDance, 2025. URL https://github. com/ByteDance-Seed~…, 2025{\natexlab{a}}.

\bibitem[Seed(2025{\natexlab{b}})]{seed2025uitars15}
ByteDance Seed.
\newblock Ui-tars-1.5.
\newblock \url{https://seed-tars.com/1.5}, 2025{\natexlab{b}}.

\bibitem[Seed et~al.(2025)Seed, Chen, Fan, Liu, Liu, Lin, Wang, Wang, Wei, Xu, et~al.]{seed2025seed1}
ByteDance Seed, Jiaze Chen, Tiantian Fan, Xin Liu, Lingjun Liu, Zhiqi Lin, Mingxuan Wang, Chengyi Wang, Xiangpeng Wei, Wenyuan Xu, et~al.
\newblock Seed1. 5-thinking: Advancing superb reasoning models with reinforcement learning.
\newblock \emph{arXiv preprint arXiv:2504.13914}, 2025.

\bibitem[Shi et~al.(2025)Shi, Yang, Liu, Bu, Chen, Zhou, Ma, Wen, Wang, He, et~al.]{shi2025korgym}
Jiajun Shi, Jian Yang, Jiaheng Liu, Xingyuan Bu, Jiangjie Chen, Junting Zhou, Kaijing Ma, Zhoufutu Wen, Bingli Wang, Yancheng He, et~al.
\newblock Korgym: A dynamic game platform for llm reasoning evaluation.
\newblock \emph{arXiv preprint arXiv:2505.14552}, 2025.

\bibitem[Silver et~al.(2017)Silver, Schrittwieser, Simonyan, Antonoglou, Huang, Guez, Hubert, Baker, Lai, Bolton, et~al.]{alphago}
David Silver, Julian Schrittwieser, Karen Simonyan, Ioannis Antonoglou, Aja Huang, Arthur Guez, Thomas Hubert, Lucas Baker, Matthew Lai, Adrian Bolton, et~al.
\newblock Mastering the game of go without human knowledge.
\newblock \emph{nature}, 550\penalty0 (7676):\penalty0 354--359, 2017.

\bibitem[Song et~al.(2025)Song, Jiang, Min, Chen, Chen, Zhao, Fang, and Wen]{song2025r1}
Huatong Song, Jinhao Jiang, Yingqian Min, Jie Chen, Zhipeng Chen, Wayne~Xin Zhao, Lei Fang, and Ji-Rong Wen.
\newblock R1-searcher: Incentivizing the search capability in llms via reinforcement learning.
\newblock \emph{arXiv preprint arXiv:2503.05592}, 2025.

\bibitem[Sutton(2019)]{sutton2019bitter}
Richard Sutton.
\newblock The bitter lesson.
\newblock \emph{Incomplete Ideas (blog)}, 13\penalty0 (1):\penalty0 38, 2019.

\bibitem[Tan et~al.(2024)Tan, Zhang, Xu, Xia, Ding, Li, Zhou, Yue, Jiang, Li, et~al.]{cradle}
Weihao Tan, Wentao Zhang, Xinrun Xu, Haochong Xia, Ziluo Ding, Boyu Li, Bohan Zhou, Junpeng Yue, Jiechuan Jiang, Yewen Li, et~al.
\newblock Cradle: Empowering foundation agents towards general computer control.
\newblock \emph{arXiv preprint arXiv:2403.03186}, 2024.

\bibitem[Team et~al.(2025)Team, Abeyruwan, Ainslie, Alayrac, Arenas, Armstrong, Balakrishna, Baruch, Bauza, Blokzijl, et~al.]{team2025gemini}
Gemini~Robotics Team, Saminda Abeyruwan, Joshua Ainslie, Jean-Baptiste Alayrac, Montserrat~Gonzalez Arenas, Travis Armstrong, Ashwin Balakrishna, Robert Baruch, Maria Bauza, Michiel Blokzijl, et~al.
\newblock Gemini robotics: Bringing ai into the physical world.
\newblock \emph{arXiv preprint arXiv:2503.20020}, 2025.

\bibitem[Team(2025{\natexlab{a}})]{kimiteam2025kimik2openagentic}
Kimi Team.
\newblock Kimi k2: Open agentic intelligence, 2025{\natexlab{a}}.
\newblock URL \url{https://arxiv.org/abs/2507.20534}.

\bibitem[Team(2025{\natexlab{b}})]{qwen3technicalreport}
Qwen Team.
\newblock Qwen3 technical report, 2025{\natexlab{b}}.
\newblock URL \url{https://arxiv.org/abs/2505.09388}.

\bibitem[Vinyals et~al.(2019)Vinyals, Babuschkin, Czarnecki, Mathieu, Dudzik, Chung, Choi, Powell, Ewalds, Georgiev, et~al.]{alphastar}
Oriol Vinyals, Igor Babuschkin, Wojciech~M Czarnecki, Micha{\"e}l Mathieu, Andrew Dudzik, Junyoung Chung, David~H Choi, Richard Powell, Timo Ewalds, Petko Georgiev, et~al.
\newblock Grandmaster level in starcraft ii using multi-agent reinforcement learning.
\newblock \emph{nature}, 575\penalty0 (7782):\penalty0 350--354, 2019.

\bibitem[Wang et~al.(2023{\natexlab{a}})Wang, Xie, Jiang, Mandlekar, Xiao, Zhu, Fan, and Anandkumar]{wang2023voyager}
Guanzhi Wang, Yuqi Xie, Yunfan Jiang, Ajay Mandlekar, Chaowei Xiao, Yuke Zhu, Linxi Fan, and Anima Anandkumar.
\newblock Voyager: An open-ended embodied agent with large language models.
\newblock \emph{arXiv preprint arXiv:2305.16291}, 2023{\natexlab{a}}.

\bibitem[Wang et~al.(2025{\natexlab{a}})Wang, Zou, Song, Feng, Fang, Lu, Liu, Luo, Liang, Huang, et~al.]{wang2025ui}
Haoming Wang, Haoyang Zou, Huatong Song, Jiazhan Feng, Junjie Fang, Junting Lu, Longxiang Liu, Qinyu Luo, Shihao Liang, Shijue Huang, et~al.
\newblock Ui-tars-2 technical report: Advancing gui agent with multi-turn reinforcement learning.
\newblock \emph{arXiv preprint arXiv:2509.02544}, 2025{\natexlab{a}}.

\bibitem[Wang et~al.(2025{\natexlab{b}})Wang, Qian, Zhong, Chen, Qiu, Huang, Jin, Wang, Wong, and Ji]{wang2025actingreasoningmoreteaching}
Hongru Wang, Cheng Qian, Wanjun Zhong, Xiusi Chen, Jiahao Qiu, Shijue Huang, Bowen Jin, Mengdi Wang, Kam-Fai Wong, and Heng Ji.
\newblock Acting less is reasoning more! teaching model to act efficiently, 2025{\natexlab{b}}.
\newblock URL \url{https://arxiv.org/abs/2504.14870}.

\bibitem[Wang et~al.(2025{\natexlab{c}})Wang, Wang, Lu, Yang, Xie, Wang, Deng, Guo, Xu, Wu, et~al.]{wang2025opencua}
Xinyuan Wang, Bowen Wang, Dunjie Lu, Junlin Yang, Tianbao Xie, Junli Wang, Jiaqi Deng, Xiaole Guo, Yiheng Xu, Chen~Henry Wu, et~al.
\newblock Opencua: Open foundations for computer-use agents.
\newblock \emph{arXiv preprint arXiv:2508.09123}, 2025{\natexlab{c}}.

\bibitem[Wang et~al.(2023{\natexlab{b}})Wang, Cai, Chen, Liu, Ma, and Liang]{wang2023describe}
Zihao Wang, Shaofei Cai, Guanzhou Chen, Anji Liu, Xiaojian~Shawn Ma, and Yitao Liang.
\newblock Describe, explain, plan and select: interactive planning with llms enables open-world multi-task agents.
\newblock \emph{Advances in Neural Information Processing Systems}, 36:\penalty0 34153--34189, 2023{\natexlab{b}}.

\bibitem[Wang et~al.(2024{\natexlab{a}})Wang, Cai, Liu, Jin, Hou, Zhang, Lin, He, Zheng, Yang, et~al.]{wang2024jarvis}
Zihao Wang, Shaofei Cai, Anji Liu, Yonggang Jin, Jinbing Hou, Bowei Zhang, Haowei Lin, Zhaofeng He, Zilong Zheng, Yaodong Yang, et~al.
\newblock Jarvis-1: Open-world multi-task agents with memory-augmented multimodal language models.
\newblock \emph{IEEE Transactions on Pattern Analysis and Machine Intelligence}, 2024{\natexlab{a}}.

\bibitem[Wang et~al.(2024{\natexlab{b}})Wang, Cai, Mu, Lin, Zhang, Liu, Li, Liu, Ma, and Liang]{wang2024omnijarvis}
Zihao Wang, Shaofei Cai, Zhancun Mu, Haowei Lin, Ceyao Zhang, Xuejie Liu, Qing Li, Anji Liu, Xiaojian~Shawn Ma, and Yitao Liang.
\newblock Omnijarvis: Unified vision-language-action tokenization enables open-world instruction following agents.
\newblock \emph{Advances in Neural Information Processing Systems}, 37:\penalty0 73278--73308, 2024{\natexlab{b}}.

\bibitem[Wang et~al.(2024{\natexlab{c}})Wang, Liu, Lin, Li, Ma, and Liang]{wang2024rat}
Zihao Wang, Anji Liu, Haowei Lin, Jiaqi Li, Xiaojian Ma, and Yitao Liang.
\newblock Rat: Retrieval augmented thoughts elicit context-aware reasoning in long-horizon generation.
\newblock \emph{arXiv preprint arXiv:2403.05313}, 2024{\natexlab{c}}.

\bibitem[Wang et~al.(2025{\natexlab{d}})Wang, Li, He, Wang, Mu, Liu, and Liang]{openha}
Zihao Wang, Muyao Li, Kaichen He, Xiangyu Wang, Zhancun Mu, Anji Liu, and Yitao Liang.
\newblock Openha: A series of open-source hierarchical agentic models in minecraft.
\newblock \emph{arXiv preprint arXiv:2509.13347}, 2025{\natexlab{d}}.

\bibitem[Wei et~al.(2022)Wei, Wang, Schuurmans, Bosma, Chi, Xia, Le, and Zhou]{wei2022chainofthought}
Jason Wei, Xuezhi Wang, Dale Schuurmans, Maarten Bosma, Ed~H. Chi, F.~Xia, Quoc Le, and Denny Zhou.
\newblock Chain-of-thought prompting elicits reasoning in large language models.
\newblock \emph{Neural Information Processing Systems}, 2022.

\bibitem[Wei et~al.(2025)Wei, Sun, Papay, McKinney, Han, Fulford, Chung, Passos, Fedus, and Glaese]{wei2025browsecomp}
Jason Wei, Zhiqing Sun, Spencer Papay, Scott McKinney, Jeffrey Han, Isa Fulford, Hyung~Won Chung, Alex~Tachard Passos, William Fedus, and Amelia Glaese.
\newblock Browsecomp: A simple yet challenging benchmark for browsing agents.
\newblock \emph{arXiv preprint arXiv:2504.12516}, 2025.

\bibitem[Wu et~al.(2024)Wu, Wu, Xu, Wang, Sun, Jia, Cheng, Ding, Chen, Liang, et~al.]{wu2024atlas}
Zhiyong Wu, Zhenyu Wu, Fangzhi Xu, Yian Wang, Qiushi Sun, Chengyou Jia, Kanzhi Cheng, Zichen Ding, Liheng Chen, Paul~Pu Liang, et~al.
\newblock Os-atlas: A foundation action model for generalist gui agents.
\newblock \emph{arXiv preprint arXiv:2410.23218}, 2024.

\bibitem[Wydmuch et~al.(2018)Wydmuch, Kempka, and Ja{\'s}kowski]{vizdoom}
Marek Wydmuch, Micha{\l} Kempka, and Wojciech Ja{\'s}kowski.
\newblock Vizdoom competitions: Playing doom from pixels.
\newblock \emph{IEEE Transactions on Games}, 2018.
\newblock IEEE Transactions on Games outstanding paper award 2022.

\bibitem[Xie et~al.(2024)Xie, Zhang, Chen, Li, Zhao, Cao, Hua, Cheng, Shin, Lei, et~al.]{xie2024osworld}
Tianbao Xie, Danyang Zhang, Jixuan Chen, Xiaochuan Li, Siheng Zhao, Ruisheng Cao, Toh~J Hua, Zhoujun Cheng, Dongchan Shin, Fangyu Lei, et~al.
\newblock Osworld: Benchmarking multimodal agents for open-ended tasks in real computer environments.
\newblock \emph{Advances in Neural Information Processing Systems}, 37:\penalty0 52040--52094, 2024.

\bibitem[Xu et~al.(2024)Xu, Wang, Wang, Lu, Xie, Saha, Sahoo, Yu, and Xiong]{xu2024aguvis}
Yiheng Xu, Zekun Wang, Junli Wang, Dunjie Lu, Tianbao Xie, Amrita Saha, Doyen Sahoo, Tao Yu, and Caiming Xiong.
\newblock Aguvis: Unified pure vision agents for autonomous gui interaction.
\newblock \emph{arXiv preprint arXiv:2412.04454}, 2024.

\bibitem[Yang et~al.(2024)Yang, Jimenez, Wettig, Lieret, Yao, Narasimhan, and Press]{yang2024sweagent}
John Yang, Carlos~E Jimenez, Alexander Wettig, Kilian Lieret, Shunyu Yao, Karthik Narasimhan, and Ofir Press.
\newblock {SWE}-agent: Agent-computer interfaces enable automated software engineering.
\newblock \emph{arXiv preprint arXiv:2405.15793}, 2024.

\bibitem[Yao et~al.(2022)Yao, Zhao, Yu, Du, Shafran, Narasimhan, and Cao]{yao2022react}
Shunyu Yao, Jeffrey Zhao, Dian Yu, Nan Du, Izhak Shafran, Karthik Narasimhan, and Yuan Cao.
\newblock React: Synergizing reasoning and acting in language models.
\newblock \emph{arXiv preprint arXiv:2210.03629}, 2022.

\bibitem[Yuan et~al.(2023)Yuan, Zhang, Wang, Xie, Cai, Dong, and Lu]{yuan2023skill}
Haoqi Yuan, Chi Zhang, Hongcheng Wang, Feiyang Xie, Penglin Cai, Hao Dong, and Zongqing Lu.
\newblock Skill reinforcement learning and planning for open-world long-horizon tasks.
\newblock \emph{arXiv preprint arXiv:2303.16563}, 2023.

\bibitem[Zhai et~al.(2024)Zhai, Bai, Lin, Pan, Tong, Zhou, Suhr, Xie, LeCun, Ma, et~al.]{zhai2024fine}
Simon Zhai, Hao Bai, Zipeng Lin, Jiayi Pan, Peter Tong, Yifei Zhou, Alane Suhr, Saining Xie, Yann LeCun, Yi~Ma, et~al.
\newblock Fine-tuning large vision-language models as decision-making agents via reinforcement learning.
\newblock \emph{Advances in neural information processing systems}, 37:\penalty0 110935--110971, 2024.

\bibitem[Zhang et~al.(2025)Zhang, Griffiths, Narasimhan, and Press]{zhang2025videogamebench}
Alex~L Zhang, Thomas~L Griffiths, Karthik~R Narasimhan, and Ofir Press.
\newblock Videogamebench: Can vision-language models complete popular video games?
\newblock \emph{arXiv preprint arXiv:2505.18134}, 2025.

\bibitem[Zhang et~al.(2023)Zhang, Yang, Hu, Wang, Li, Sun, Zhang, Zhang, Liu, Zhu, et~al.]{proagent}
Ceyao Zhang, Kaijie Yang, Siyi Hu, Zihao Wang, Guanghe Li, Yihang Sun, Cheng Zhang, Zhaowei Zhang, Anji Liu, Song-Chun Zhu, et~al.
\newblock Proagent: Building proactive cooperative ai with large language models.
\newblock \emph{arXiv preprint arXiv:2308.11339}, 2023.

\bibitem[Zhao et~al.(2023)Zhao, Kumar, Levine, and Finn]{zhao2023learning}
Tony~Z Zhao, Vikash Kumar, Sergey Levine, and Chelsea Finn.
\newblock Learning fine-grained bimanual manipulation with low-cost hardware.
\newblock \emph{arXiv preprint arXiv:2304.13705}, 2023.

\end{thebibliography}

\clearpage

\beginappendix
\section{Rollout Trajectories}

\subsection{Rollouts in Minecraft}

\begin{figure}[h]
    \centering
    \begin{minipage}[b]{0.95\textwidth} 
        \centering
        \includegraphics[width=0.95\textwidth]{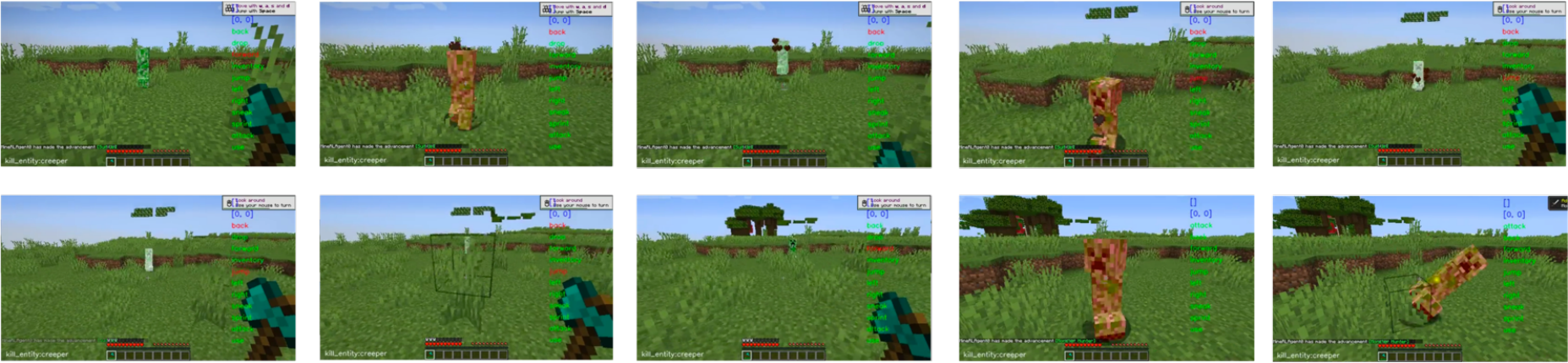}  
        \caption{
        \gametars has mastered advanced combat skills. In battles with creepers, he avoids damage from creeper explosions by constantly retreating and also uses jumping strikes to increase his damage.
        }
        \label{fig:minecraft_creeper_demo}
    \end{minipage}
    \vspace{1pt}  
    
    \begin{minipage}[b]{0.95\textwidth} 
        \centering
        \includegraphics[width=0.95\textwidth]{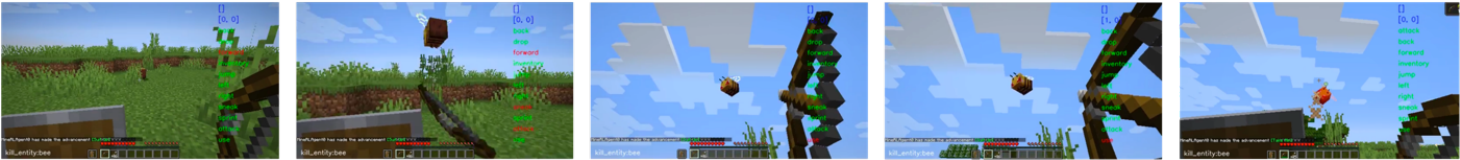}
    \caption{
    \gametars can use tools and interact with objects in the environment.
    }
    \label{fig:minecraft_bowuse_demo}
    \end{minipage}
    \vspace{1pt}
    
    \begin{minipage}[b]{0.95\textwidth} 
        \centering
        \includegraphics[width=0.95\textwidth]{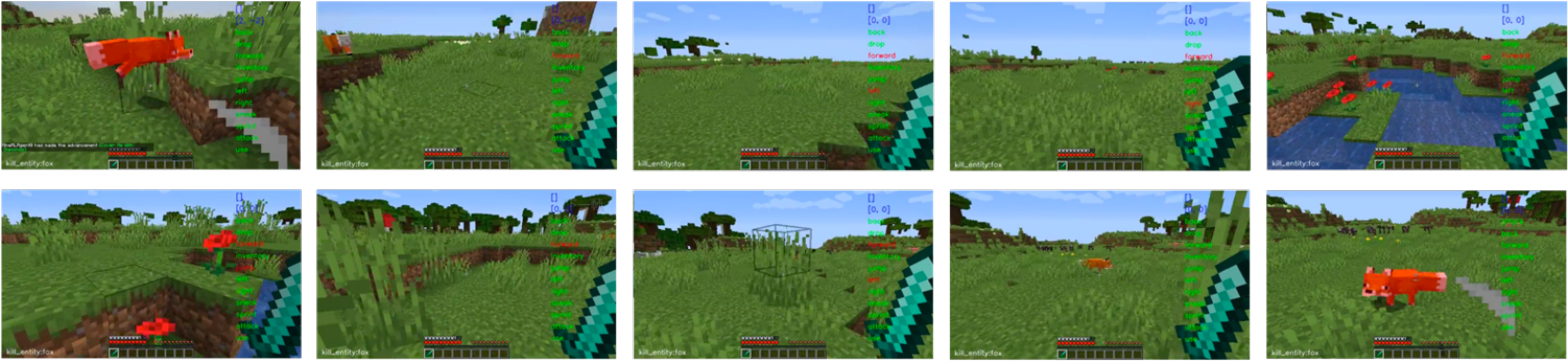}
    \caption{
    \gametars have good long-term memory, allowing them to remember the movement trajectory of foxes during the process of killing them, and to track the foxes over a long period through navigation and exploration skills.
    }
    \label{fig:minecraft_tracking_demo}
    \end{minipage}
    \vspace{1pt}

    \begin{minipage}[b]{0.95\textwidth}  
        \centering
        \includegraphics[width=0.95\textwidth]{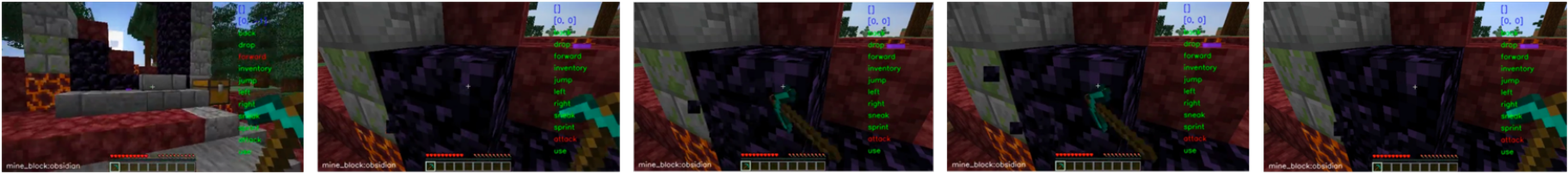}
    \caption{
    \gametars has great action consistency and can break obsidian blocks that require a 10-second long press to obtain.
    }
    \label{fig:minecraft_consistency_demo}
    \end{minipage}
    \vspace{1pt}

    \begin{minipage}[b]{0.95\textwidth}  
        \centering
        \includegraphics[width=0.95\textwidth]{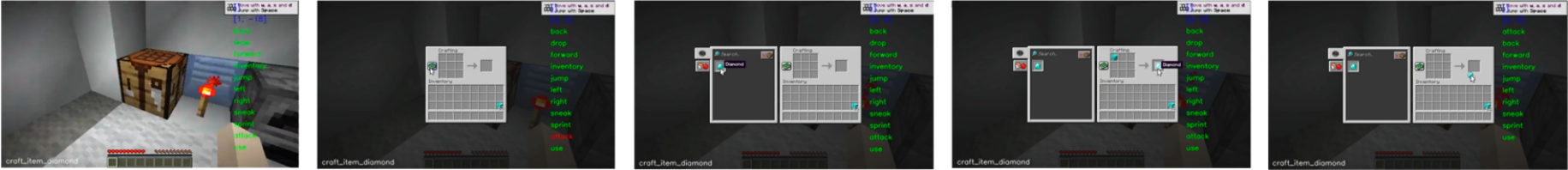}
    \caption{
    \gametars can use unified actions to freely switch and interact between the embodied 3D interface and the 2D GUI interface.
    }
    \label{fig:minecraft_gui_demo}
    \end{minipage}
\end{figure}

\subsection{Rollouts in Unseen Games}

\begin{table}[h]
\centering
\caption{Rollout trajectory of \gametars on Poki Plane game. We skip some timestamps. }
\label{tab:demo_poki_plane}
\resizebox{\textwidth}{!}{%
\begin{tabular}{lll}
\toprule
$t$ & Observation & Model Response \\ \midrule
0 & \begin{minipage}[c]{0.25\textwidth}  \centering \includegraphics[scale=0.08]{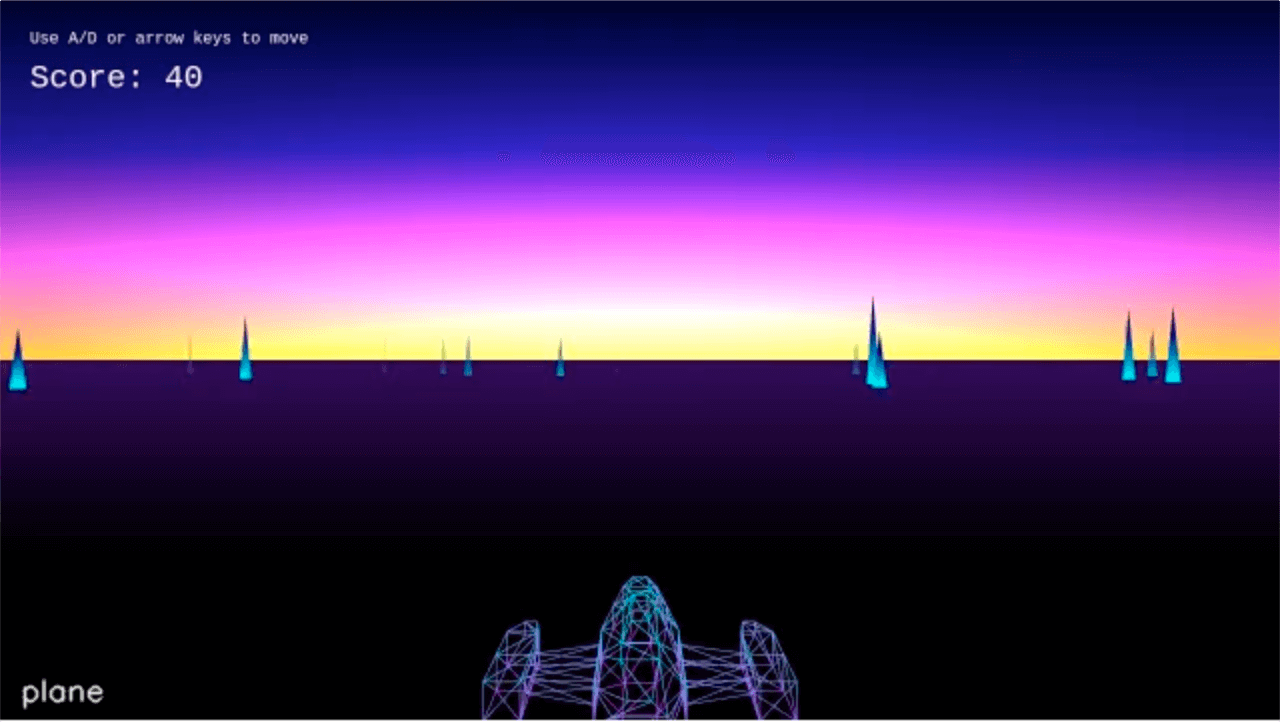}\end{minipage} & \begin{tabular}[c]{@{}l@{}}Thought: Just entered a new scene, and there are obstacles ahead. I need to quickly adjust the direction to avoid them; I \\can't let the plane crash, or else the game will be over.  Action: no\_op\end{tabular} \\

\multicolumn{3}{l}{...} \\ 

6 & \begin{minipage}[c]{0.25\textwidth}  \centering \includegraphics[scale=0.08]{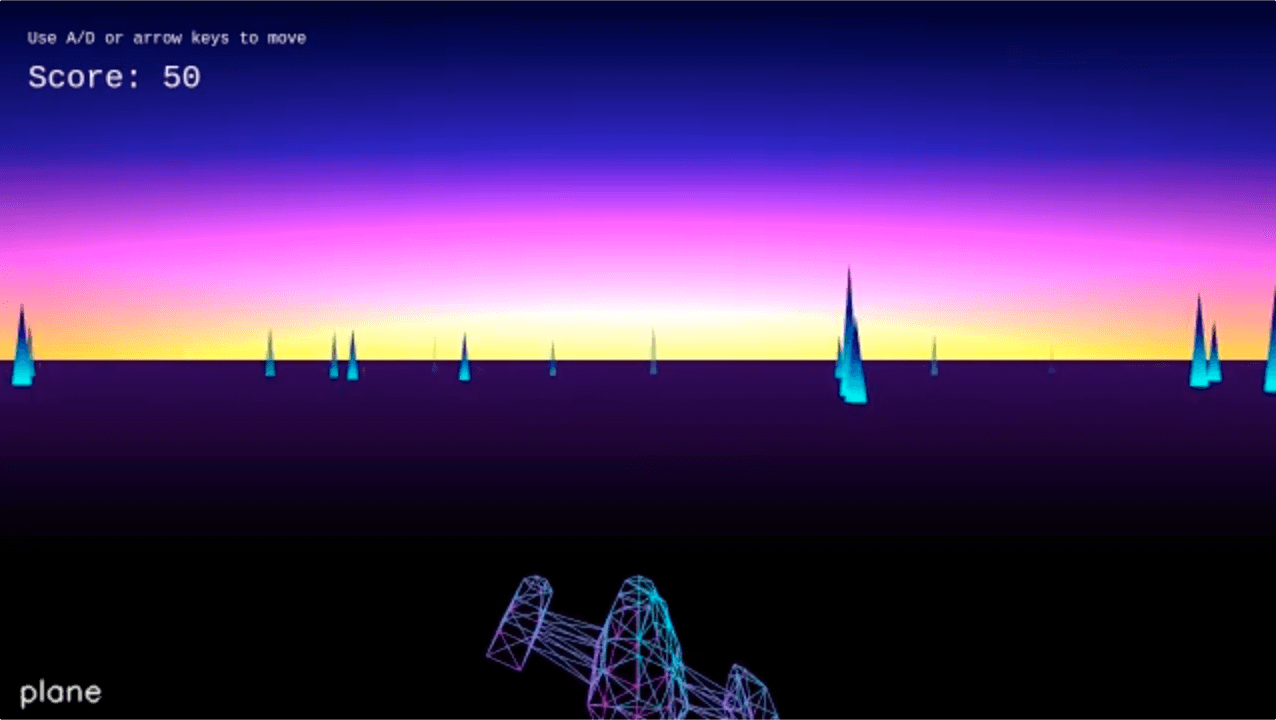}\end{minipage} & \begin{tabular}[c]{@{}l@{}}Action: keyPress(d)\end{tabular} \\

\multicolumn{3}{l}{...} \\ 

14 & \begin{minipage}[c]{0.25\textwidth}  \centering \includegraphics[scale=0.08]{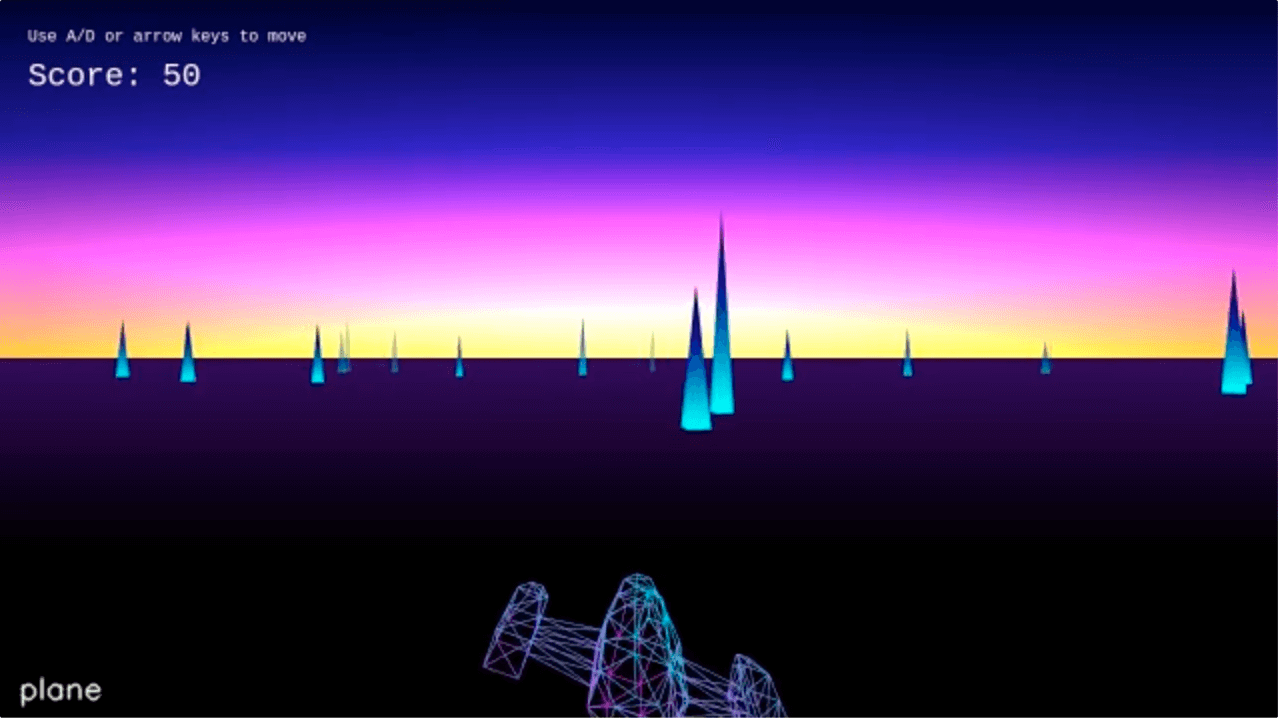}\end{minipage} & \begin{tabular}[c]{@{}l@{}}Thought: The position of the obstacle ahead has changed, need to adjust direction to the right, maintain distance from the \\obstacle, and continue flying forward. Action: keyPress(d)\end{tabular} \\

\multicolumn{3}{l}{...} \\ 

21 & \begin{minipage}[c]{0.25\textwidth}  \centering \includegraphics[scale=0.08]{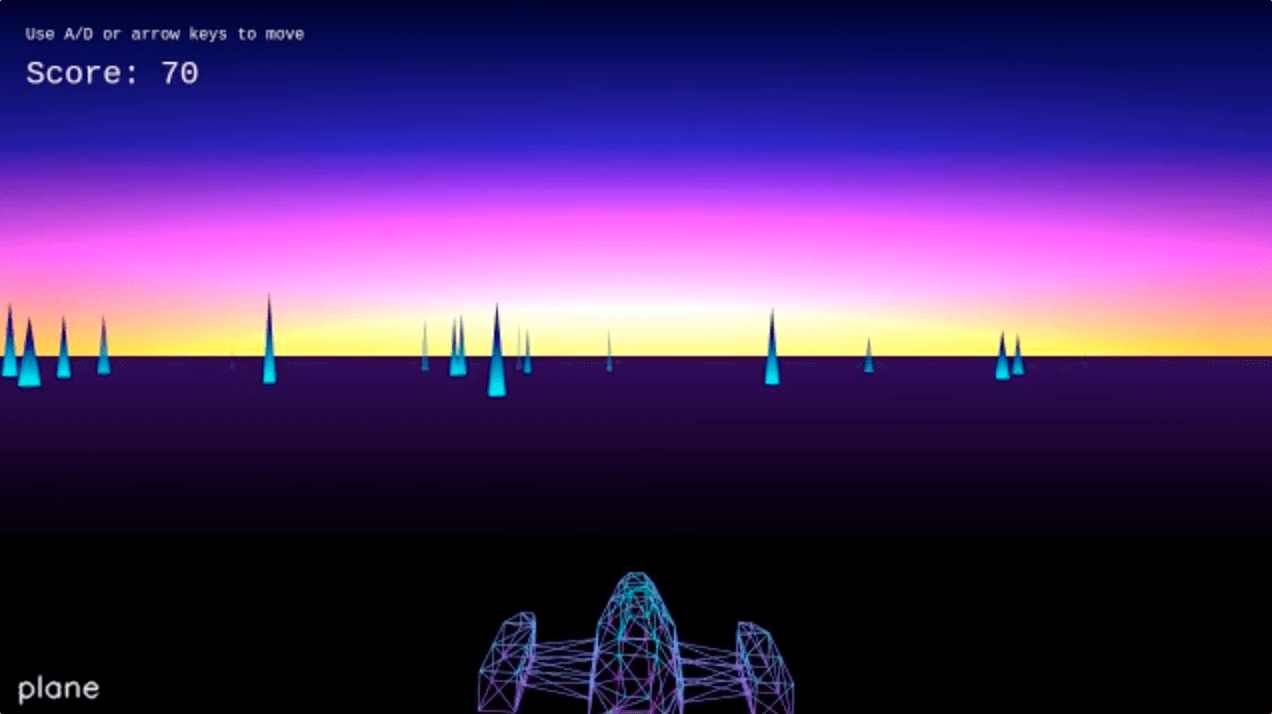}\end{minipage} & \begin{tabular}[c]{@{}l@{}}Thought: There are new obstacles appearing again, need to adjust the direction to the left to avoid this, cannot let the plane \\hit any obstacles. Action: keyPress(a)\end{tabular} \\

\multicolumn{3}{l}{...} \\ 

35 & \begin{minipage}[c]{0.25\textwidth}  \centering \includegraphics[scale=0.08]{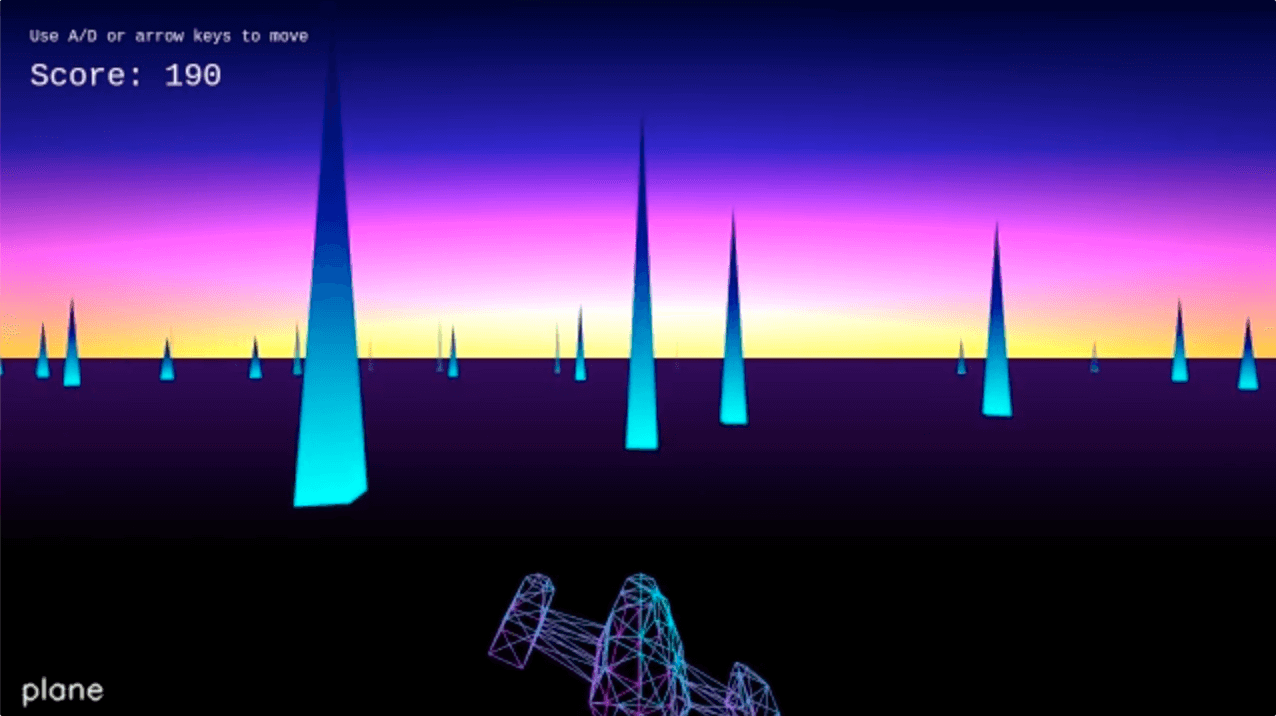}\end{minipage} & \begin{tabular}[c]{@{}l@{}}Thought: The obstacles in front are arranged quite densely, so I need to adjust the direction to the right to avoid them; I can't \\let the plane hit them, otherwise the game will be over. Action: keyPress (d)\end{tabular} \\

\multicolumn{3}{l}{...} \\ \bottomrule
\end{tabular}%
}
\end{table}

\begin{table}[h]
\centering
\caption{Rollout trajectory of \gametars on Temple Run game. We skip some timestamps. }
\label{tab:demo_poki_templerun}
\resizebox{\textwidth}{!}{%
\begin{tabular}{lll}
\toprule
$t$ & Observation & Model Response \\ \midrule
0 & \begin{minipage}[c]{0.25\textwidth}  \centering \includegraphics[scale=0.08]{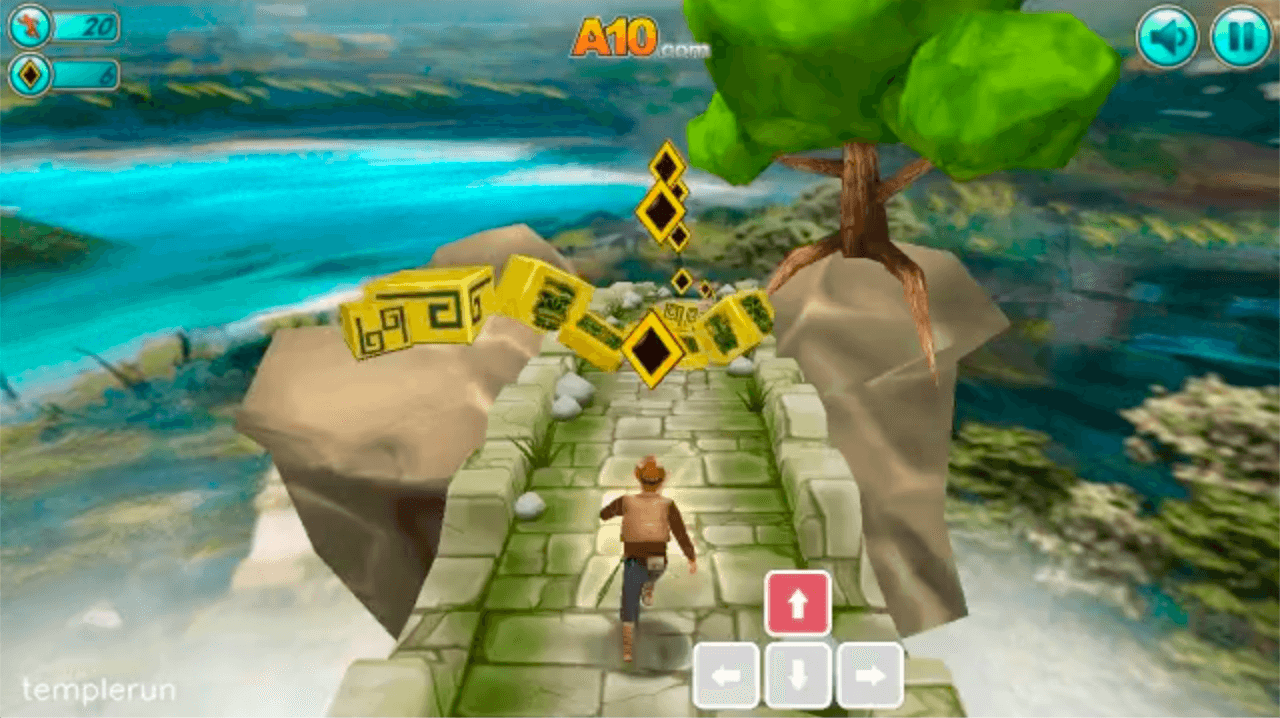}\end{minipage} & \begin{tabular}[c]{@{}l@{}}Thought: A diamond-shaped obstacle appeared ahead, I need to jump up quickly to avoid it, otherwise I will crash and lose health.\\ Action: keyPress(arrowup)\end{tabular} \\
1 & \begin{minipage}[c]{0.25\textwidth}  \centering \includegraphics[scale=0.08]{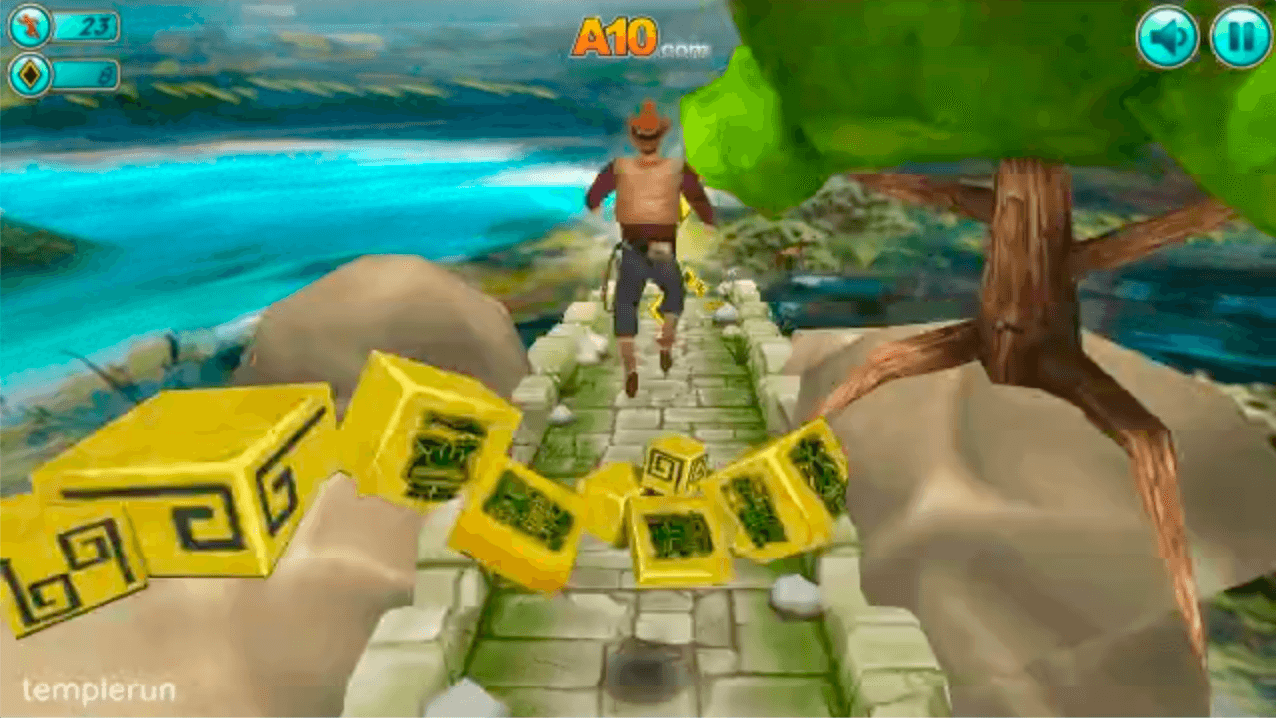}\end{minipage} & \begin{tabular}[c]{@{}l@{}}Action: no\_op\end{tabular} \\

\multicolumn{3}{l}{...} \\ 

12 & \begin{minipage}[c]{0.25\textwidth}  \centering \includegraphics[scale=0.08]{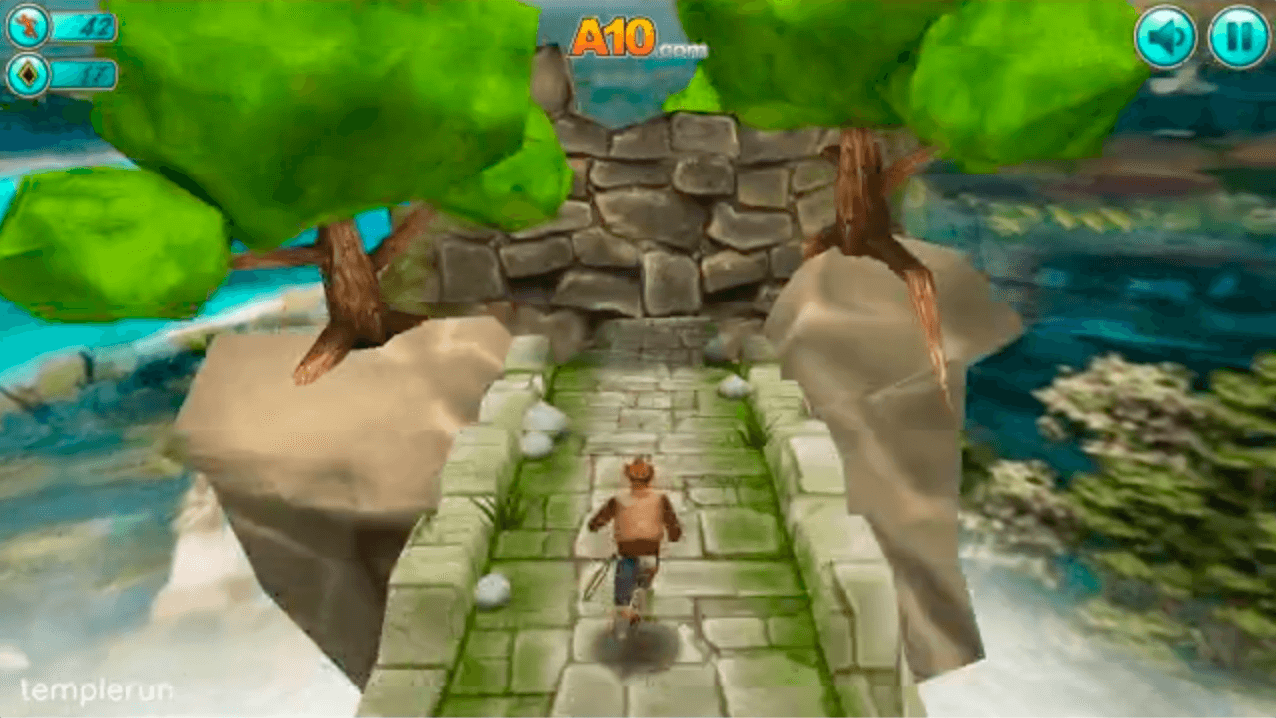}\end{minipage} & \begin{tabular}[c]{@{}l@{}}Thought: There was a stone wall ahead, you had to crouch down to pass through, otherwise you would be blocked, so \\quickly press the crouch key. Action: keyPress (arrowdown)\end{tabular} \\

\multicolumn{3}{l}{...} \\ 

24 & \begin{minipage}[c]{0.25\textwidth}  \centering \includegraphics[scale=0.08]{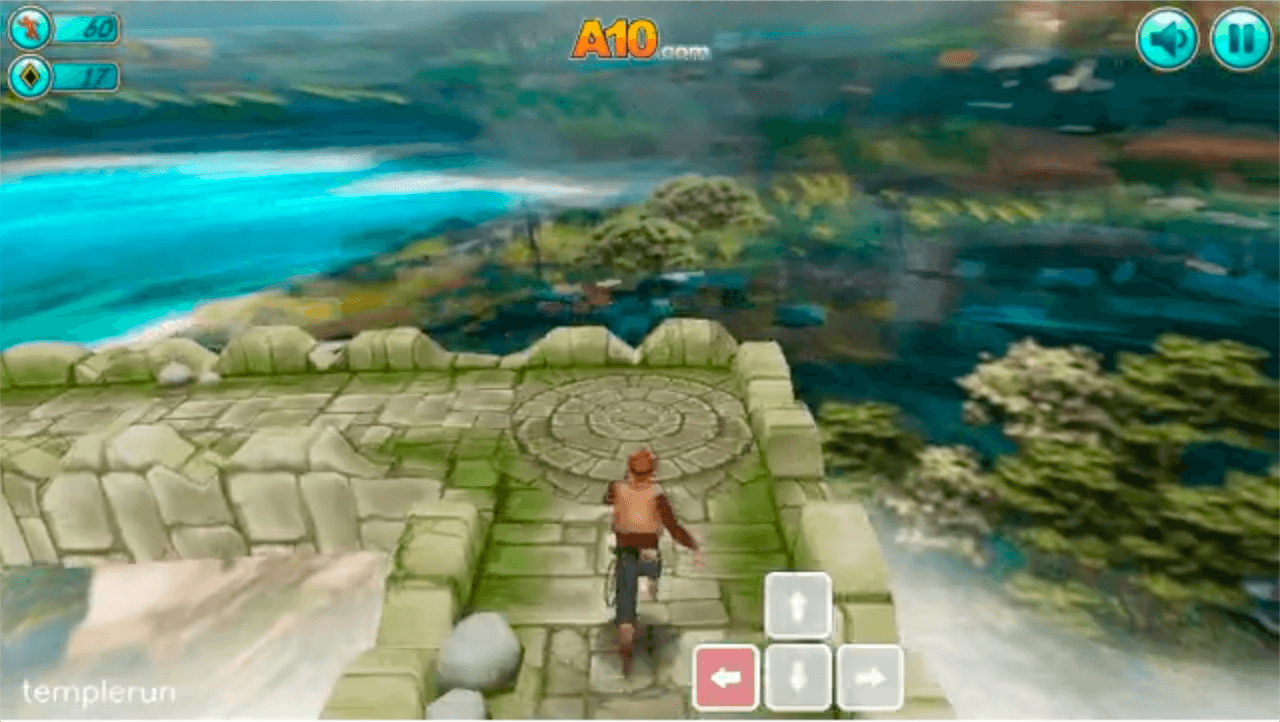}\end{minipage} & \begin{tabular}[c]{@{}l@{}}Action: keyPress(arrowleft)\end{tabular} \\

\multicolumn{3}{l}{...} \\ 

37 & \begin{minipage}[c]{0.25\textwidth}  \centering \includegraphics[scale=0.08]{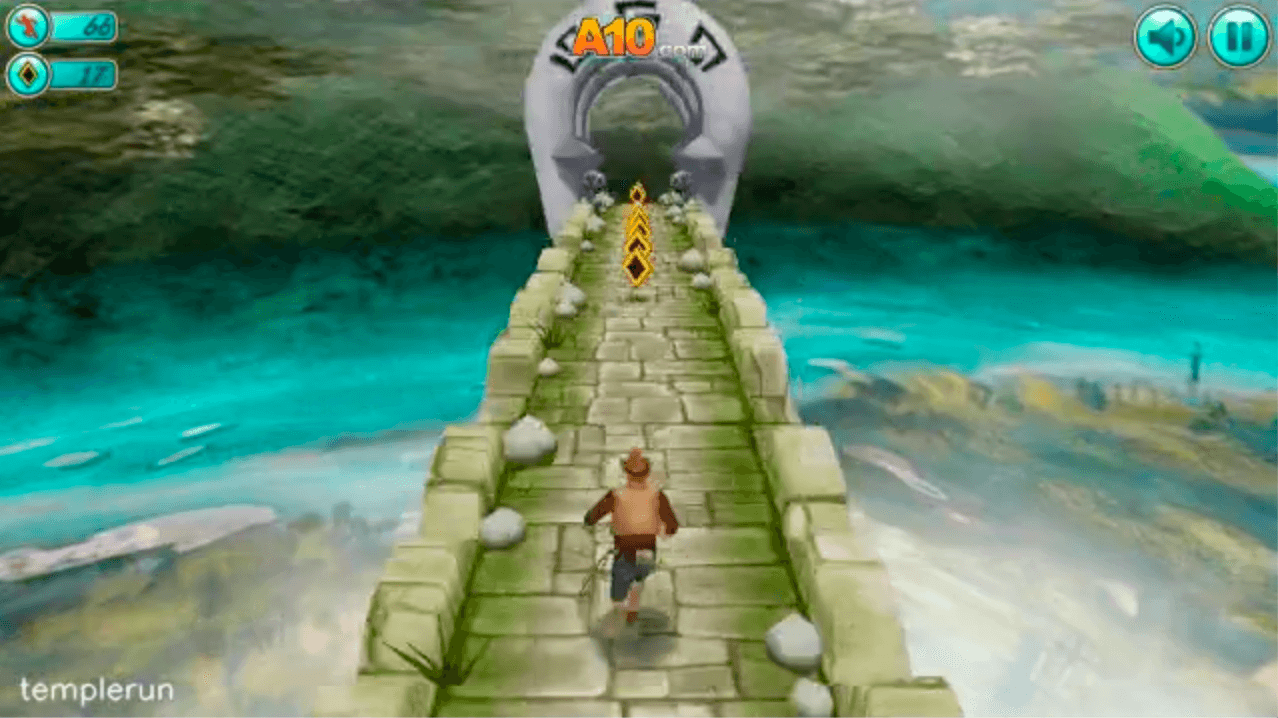}\end{minipage} & \begin{tabular}[c]{@{}l@{}}Thought: There are diamonds and stone pillars ahead; be careful to avoid the stone pillars while jumping over the diamonds, \\and stay focused. Action: no\_op\end{tabular} \\

\multicolumn{3}{l}{...} \\ 

53 & \begin{minipage}[c]{0.25\textwidth}  \centering \includegraphics[scale=0.08]{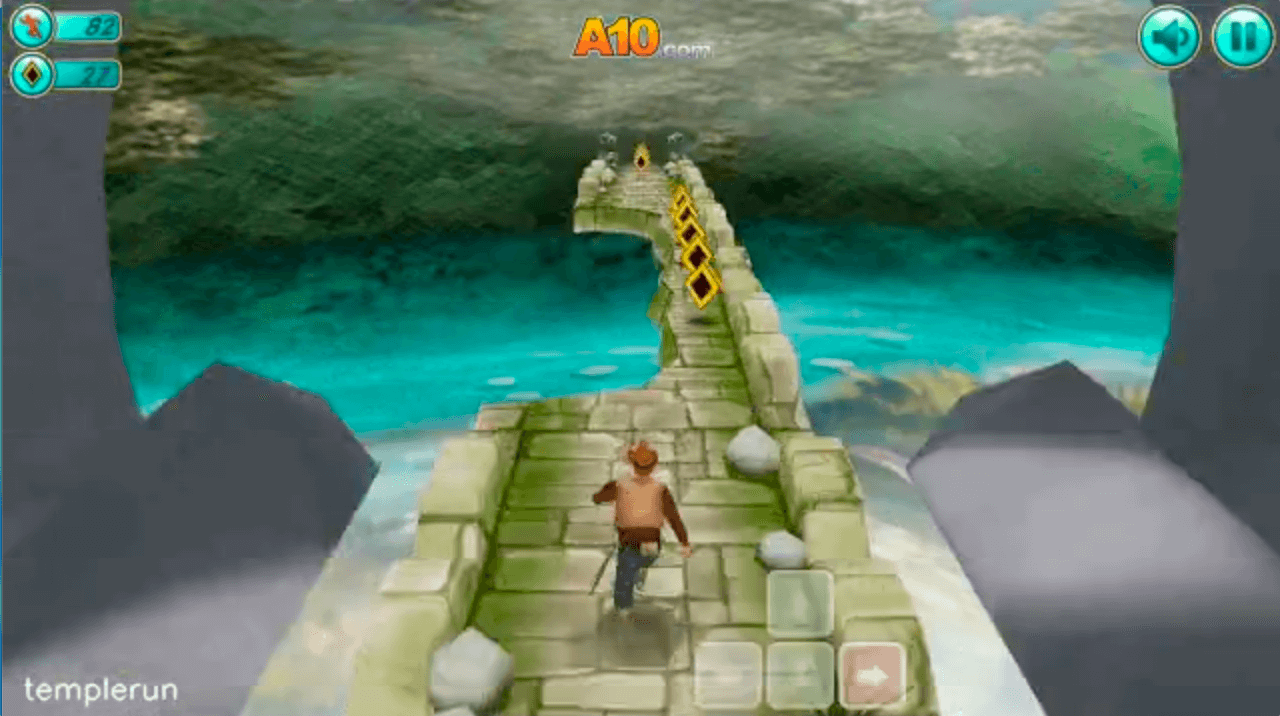}\end{minipage} & \begin{tabular}[c]{@{}l@{}}Thought: There is a gap ahead, you need to avoid it, otherwise you will fall. Press the right button to ensure a safe passage. \\Action: keyPress(arrowright)\end{tabular} \\

\multicolumn{3}{l}{...} \\ 

67 & \begin{minipage}[c]{0.25\textwidth}  \centering \includegraphics[scale=0.08]{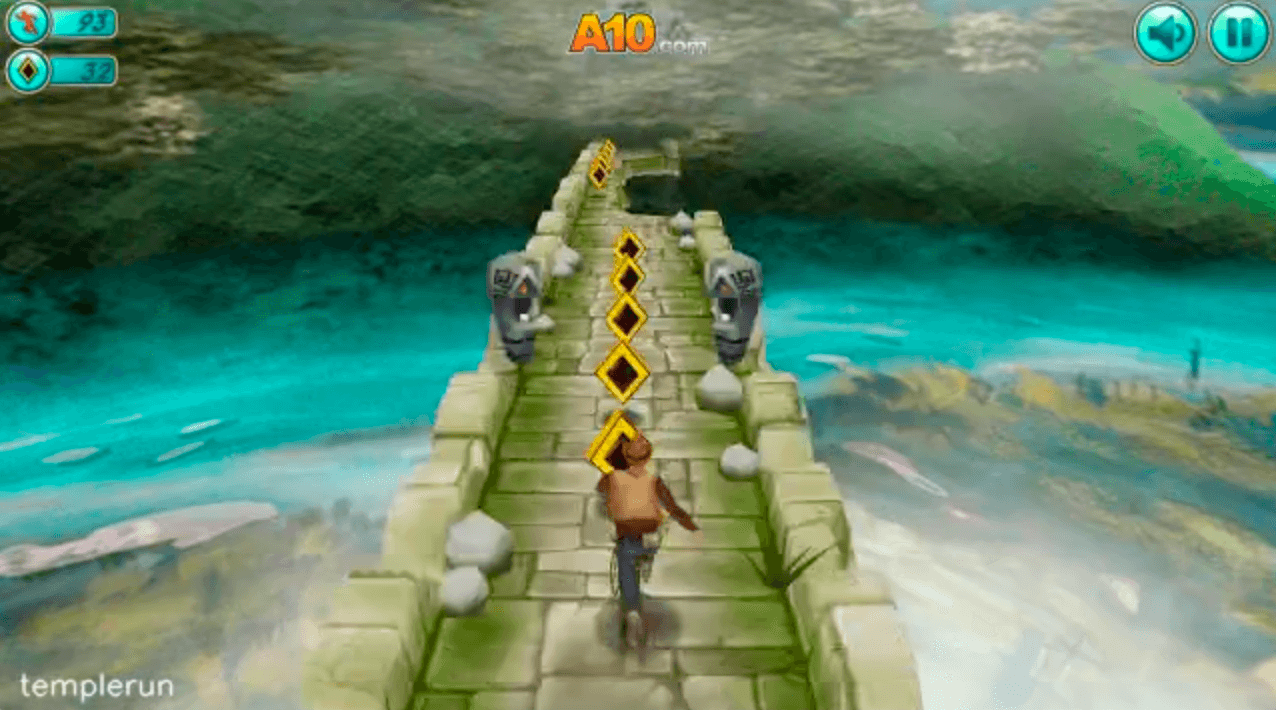}\end{minipage} & \begin{tabular}[c]{@{}l@{}}Action: keyPress(arrowleft)\end{tabular} \\

\multicolumn{3}{l}{...} \\ 

82 & \begin{minipage}[c]{0.25\textwidth}  \centering \includegraphics[scale=0.08]{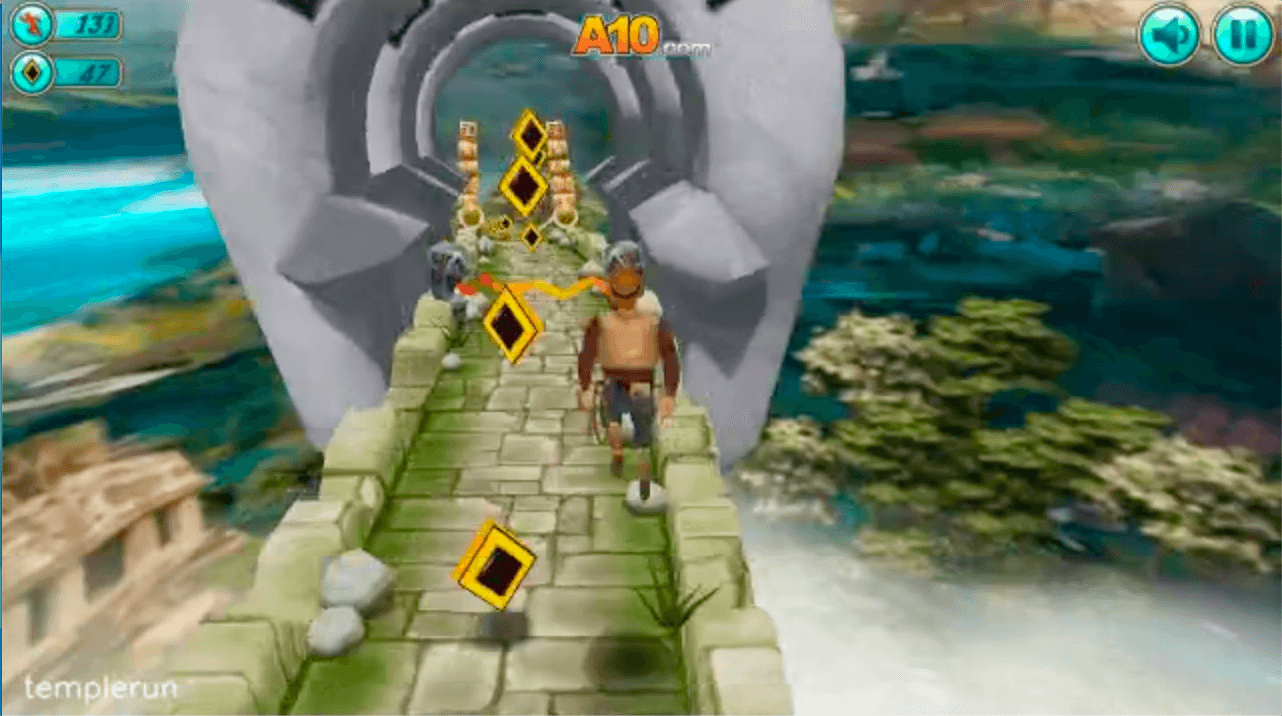}\end{minipage} & \begin{tabular}[c]{@{}l@{}}Action: keyPress(arrowup)\end{tabular} \\

\multicolumn{3}{l}{...} \\ 

109 & \begin{minipage}[c]{0.25\textwidth}  \centering \includegraphics[scale=0.08]{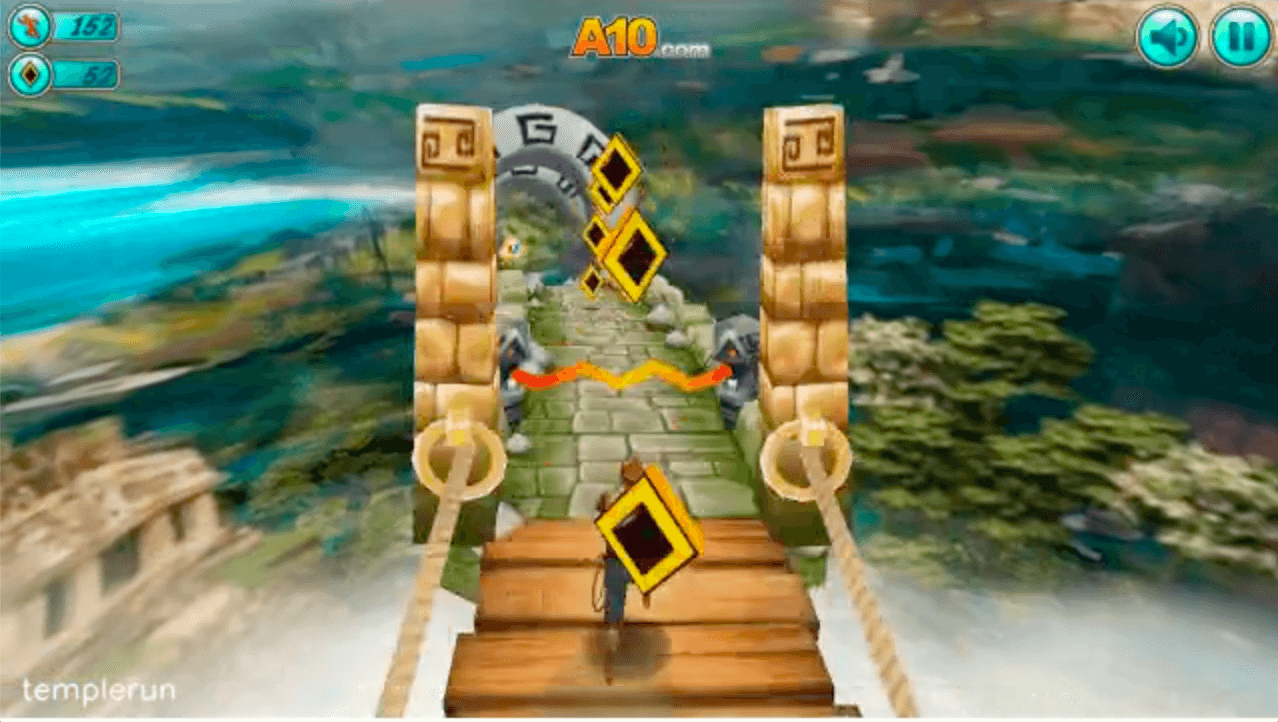}\end{minipage} & \begin{tabular}[c]{@{}l@{}}Action: keyPress(arrowup)\end{tabular} \\

\multicolumn{3}{l}{...} \\ \bottomrule
\end{tabular}%
}
\end{table}

\begin{table}[]
\centering
\caption{Rollout trajectory of \gametars on Poki Race game. We skip some timestamps. }
\label{tab:demo_poki}
\resizebox{\textwidth}{!}{%
\begin{tabular}{lll}
\toprule
$t$ & Observation & Model Response \\ \midrule
0 & \begin{minipage}[c]{0.25\textwidth}  \centering \includegraphics[scale=0.28]{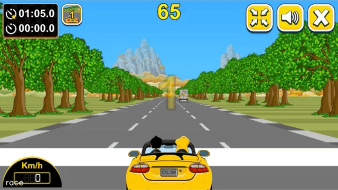}\end{minipage} & \begin{tabular}[c]{@{}l@{}}Thought: The game interface has now loaded, and I need to start driving the vehicle. First, I have to hold down the \\acceleration key to get the car moving; only then can I start the timer. I must ensure that I maintain acceleration \\from the very beginning and not let my speed drop so that I can reach the finish line as quickly as possible. Action: \\keyPress(arrowup)\end{tabular} \\

\multicolumn{3}{l}{...} \\ 

6 & \begin{minipage}[c]{0.25\textwidth}  \centering \includegraphics[scale=0.07]{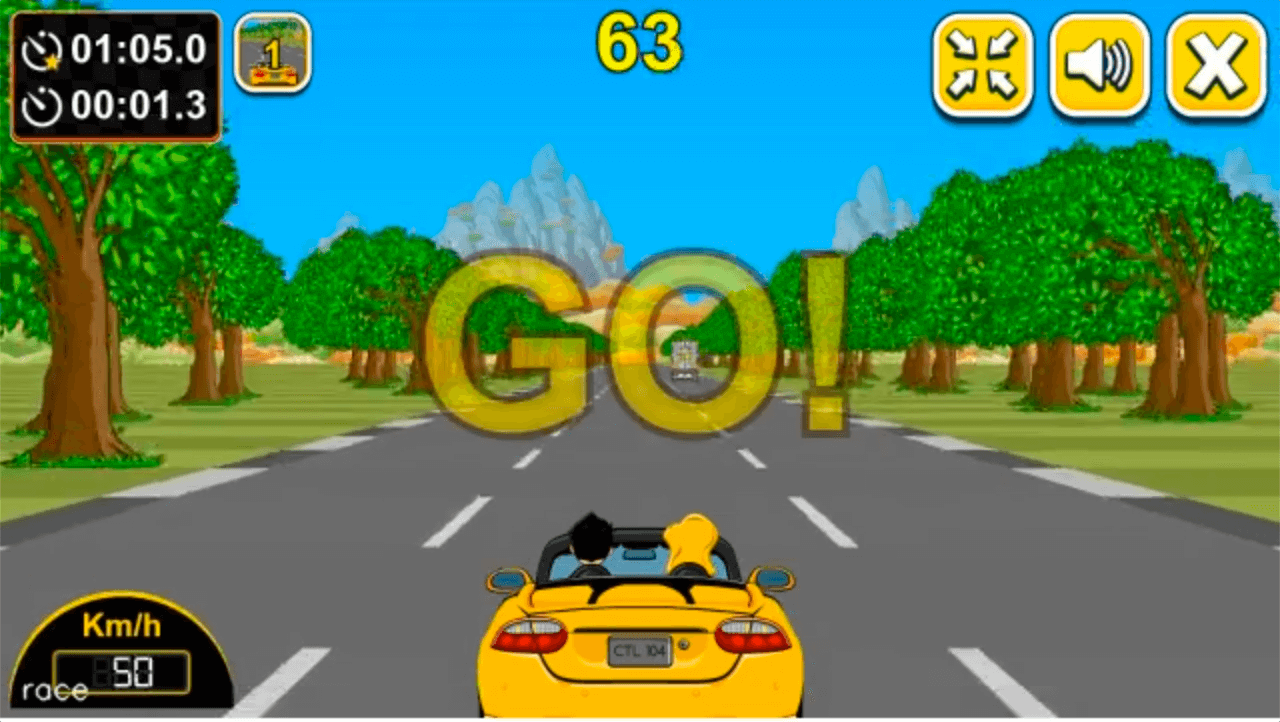}\end{minipage} & \begin{tabular}[c]{@{}l@{}}Thought: I see the "GO!" signal now, it's time to accelerate with all my might and pick up speed. The road ahead \\looks straight, so I'll stay in the middle lane to avoid hitting the trees or other cars on either side. I need to pay \\attention to the vehicles ahead, but for now, I'll focus on accelerating and aim to get my speed up right from the \\start. Action: keyPress(arrowup)\end{tabular} \\
7 & \begin{minipage}[c]{0.25\textwidth}  \centering \includegraphics[scale=0.07]{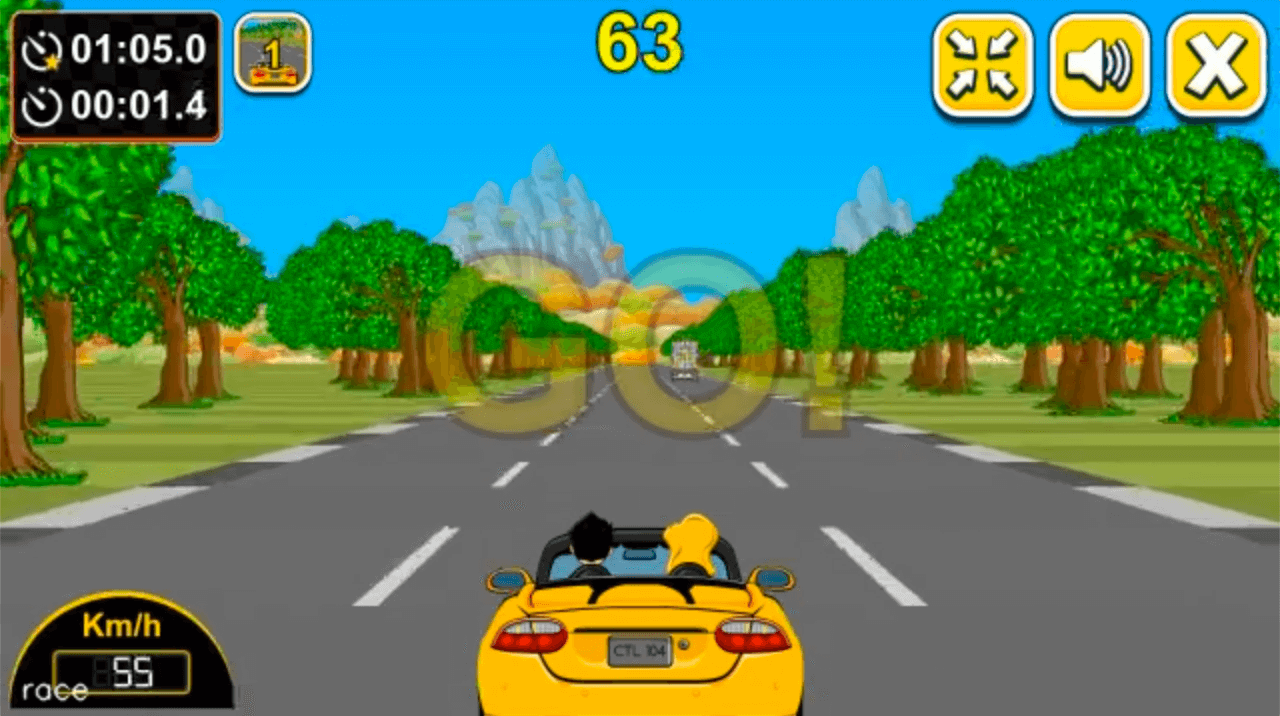}\end{minipage} & \begin{tabular}[c]{@{}l@{}}Action: keyPress(arrowup)\end{tabular} \\

\multicolumn{3}{l}{...} \\ 

17 & \begin{minipage}[c]{0.25\textwidth}  \centering \includegraphics[scale=0.07]{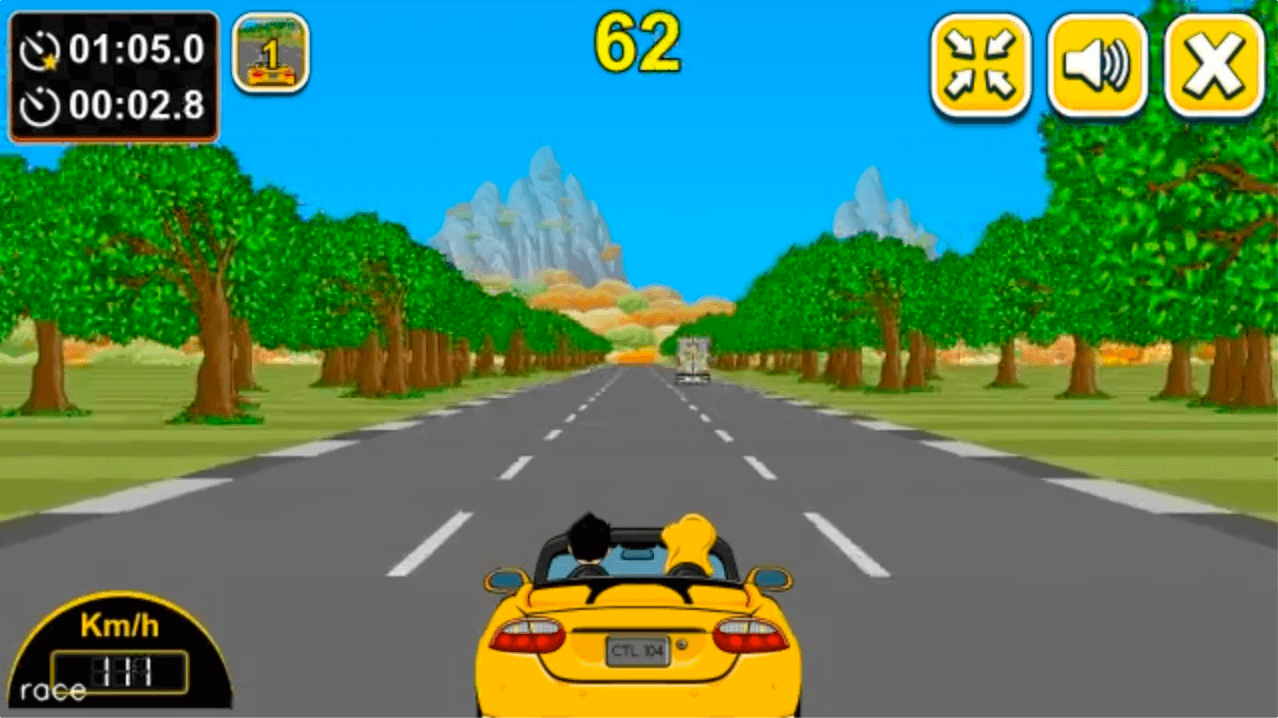}\end{minipage} & \begin{tabular}[c]{@{}l@{}}Thought: The speed has already reached over 100, and the road ahead is still very straight, so continue to \\accelerate. Now I need to pay attention to the vehicles in front, but currently there are no other cars, so I can keep \\accelerating. At the same time, I need to ensure that I'm in the center of the lane; this way it's both safe and helps \\maintain speed, avoiding veering off the road which could lead to a slowdown. Action: keyPress(arrowup)\end{tabular} \\

\multicolumn{3}{l}{...} \\ 

56 & \begin{minipage}[c]{0.25\textwidth}  \centering \includegraphics[scale=0.07]{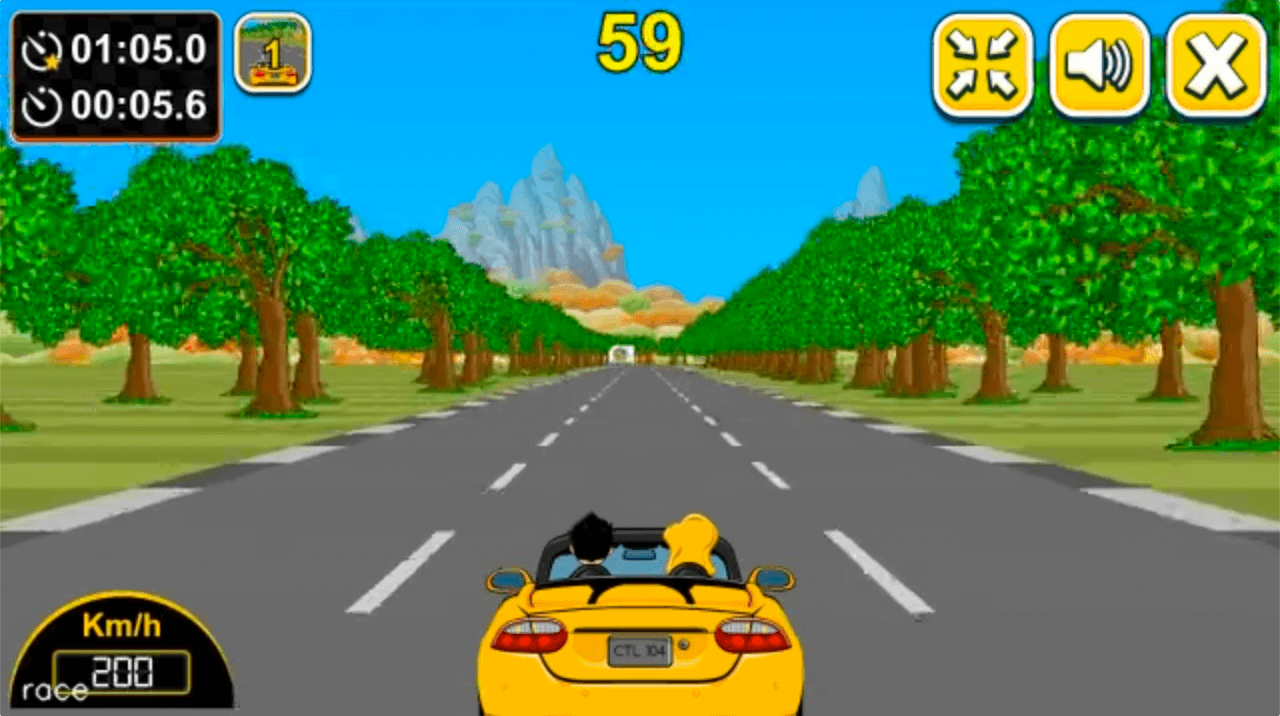}\end{minipage} & \begin{tabular}[c]{@{}l@{}}Thought: The car on the right has passed, now return to the middle lane and continue to accelerate. The speed is \\already 200; maintain this speed. The road ahead is still straight, as long as I don't deviate from the road, I can keep \\going at high speed. I need to pay attention and look further ahead to anticipate any potential situations. Action: \\keyPress(arrowup)\end{tabular} \\

\multicolumn{3}{l}{...} \\ 

82 & \begin{minipage}[c]{0.25\textwidth}  \centering \includegraphics[scale=0.07]{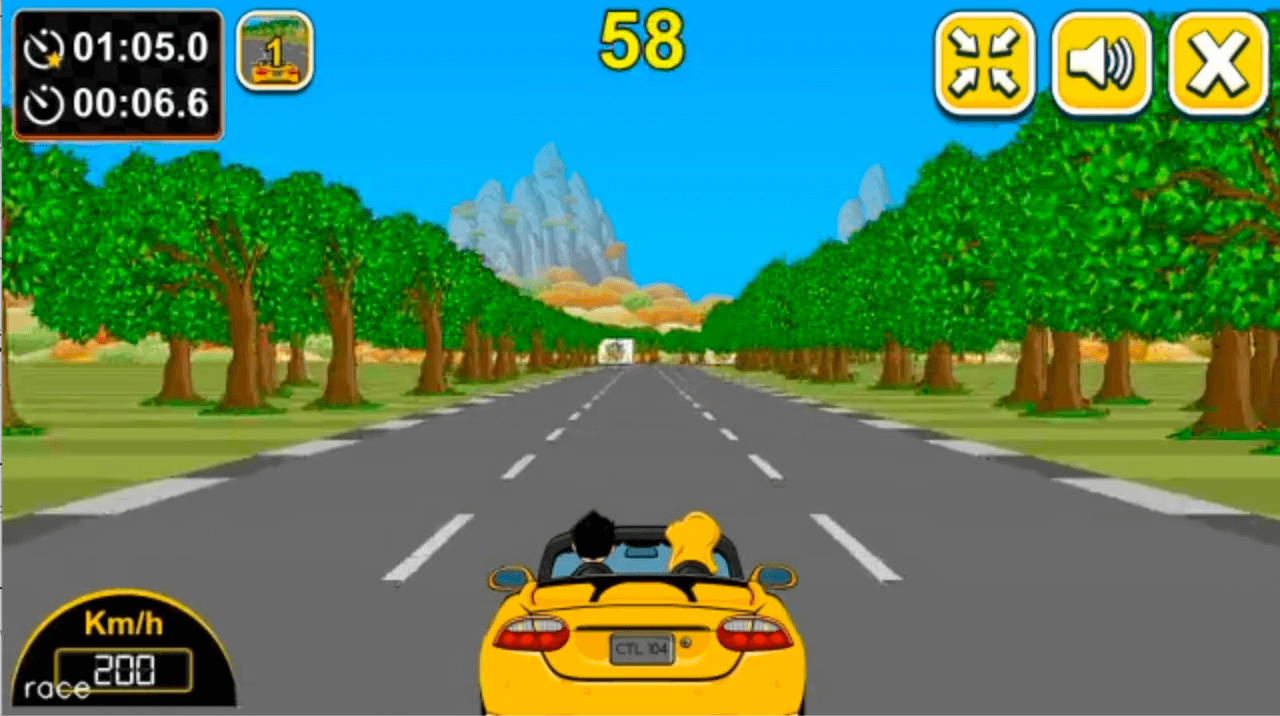}\end{minipage} & \begin{tabular}[c]{@{}l@{}}Action: keyPress(arrowup)\end{tabular} \\
83 & \begin{minipage}[c]{0.25\textwidth}  \centering \includegraphics[scale=0.07]{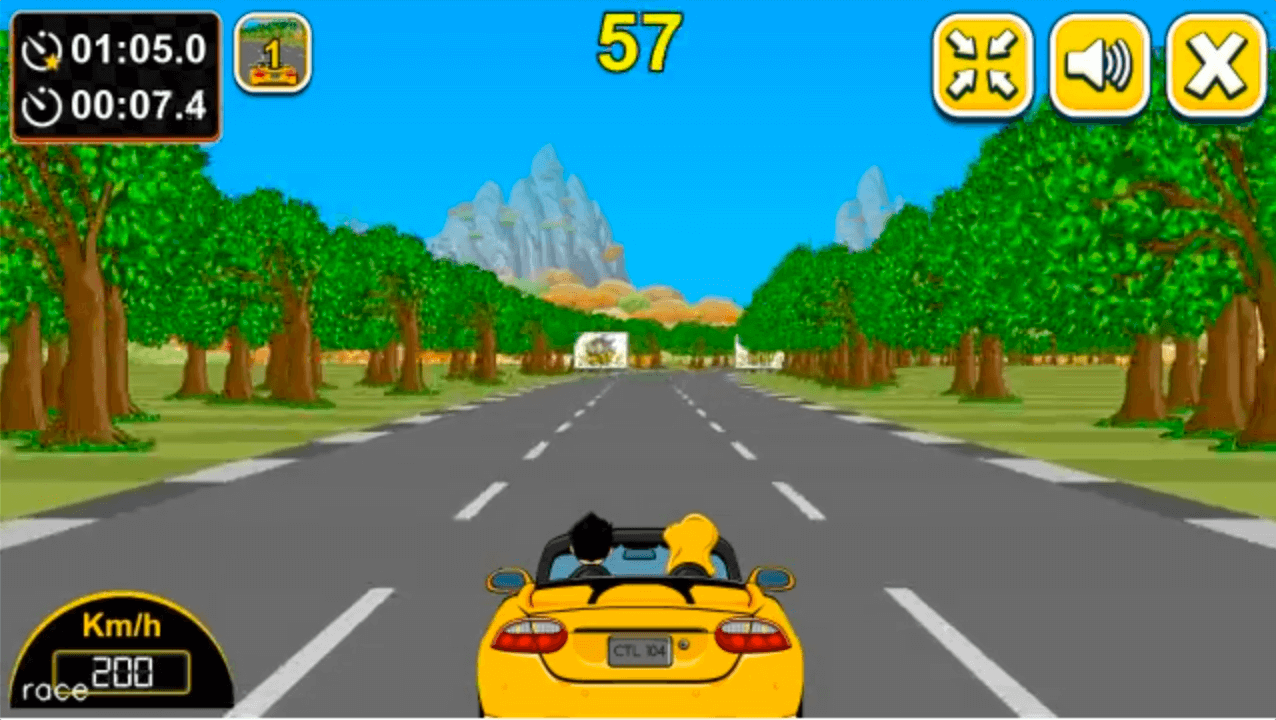}\end{minipage} & \begin{tabular}[c]{@{}l@{}}Thought: The car in front has already passed, now return to the middle lane and continue to accelerate. The speed \\is still 200, the road remains straight, and there are no other vehicles around, so I can maintain this speed. I need to \\pay attention to the mini-map to ensure I'm in the center of the lane; this way it's both safe and allows me to keep \\my speed, avoiding deceleration due to drifting off the road.  Action: keyPress(arrowup)\end{tabular} \\
84 & \begin{minipage}[c]{0.25\textwidth}  \centering \includegraphics[scale=0.07]{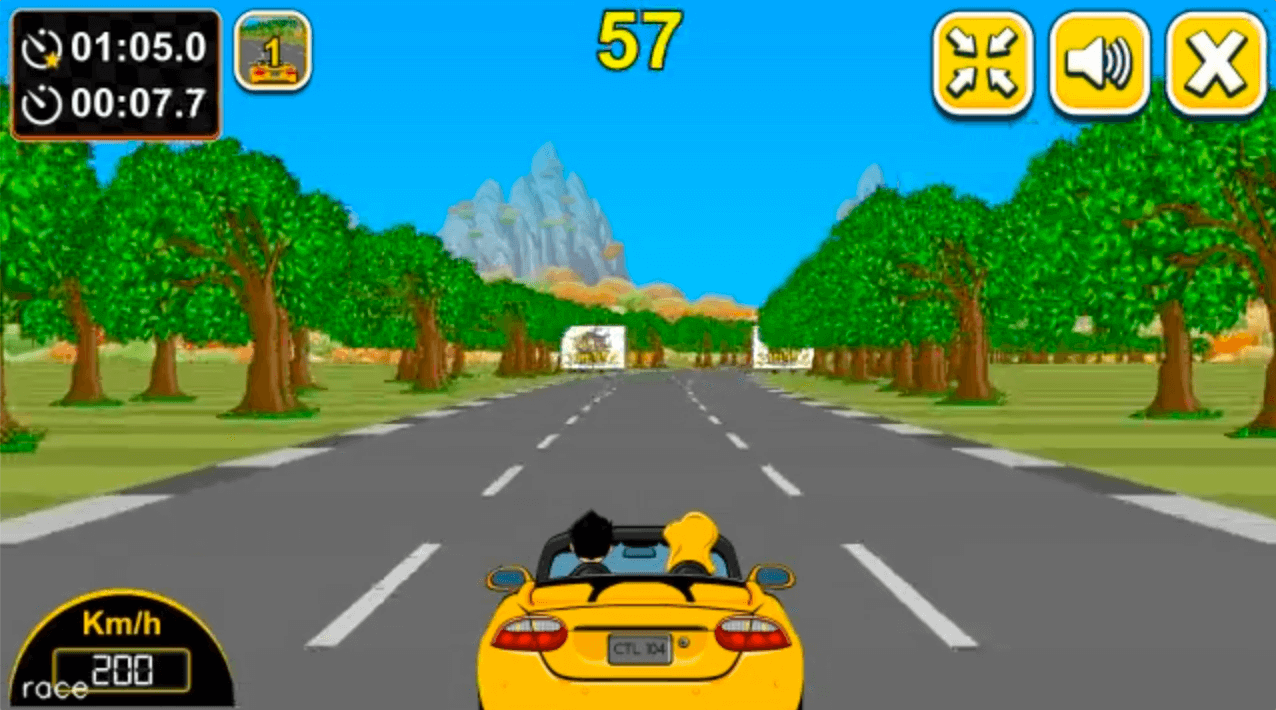}\end{minipage} & \begin{tabular}[c]{@{}l@{}}Action: keyPress(arrowup)\end{tabular} \\

\multicolumn{3}{l}{...} \\ 

100 & \begin{minipage}[c]{0.25\textwidth}  \centering \includegraphics[scale=0.07]{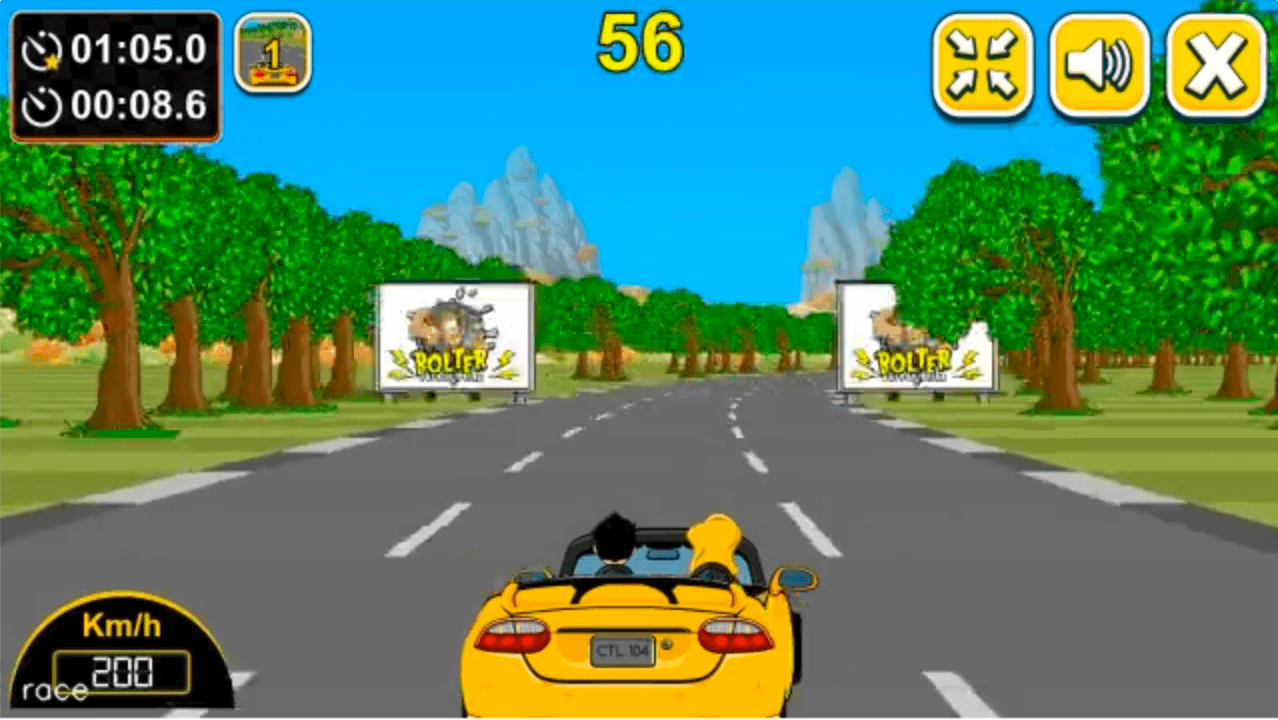}\end{minipage} & \begin{tabular}[c]{@{}l@{}}Thought: The road ahead is starting to turn, so I need to adjust my direction in advance and stay in the center of the \\road. When turning, I can't reduce speed too much, so I need to steer early, turn slowly while continuing to \\accelerate, and try to maintain my speed. This way is both safe and allows me to keep the fastest speed.   Action: \\keyPress(arrowup, arrowright)\end{tabular} \\
101 & \begin{minipage}[c]{0.25\textwidth}  \centering \includegraphics[scale=0.07]{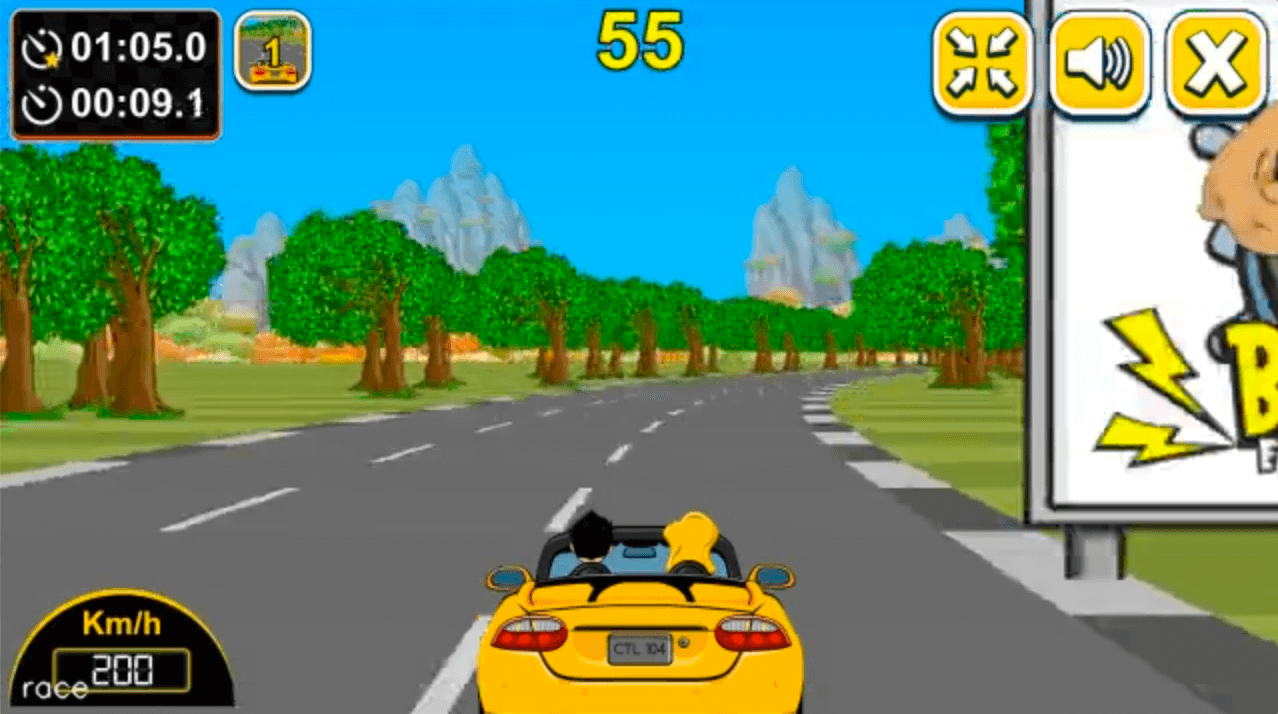}\end{minipage} & \begin{tabular}[c]{@{}l@{}}Action: keyPress(arrowup)\end{tabular} \\

\multicolumn{3}{l}{...} \\ 

130 & \begin{minipage}[c]{0.25\textwidth}  \centering \includegraphics[scale=0.07]{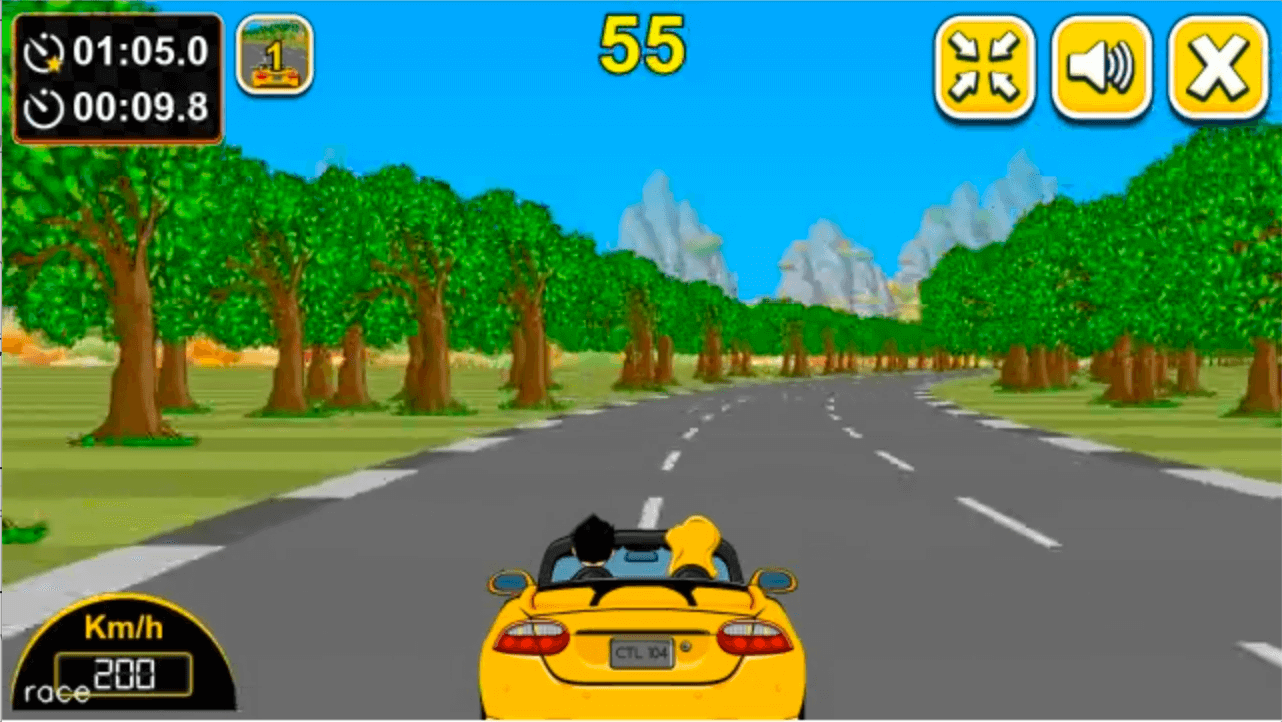}\end{minipage} & \begin{tabular}[c]{@{}l@{}}Action: keyPress(arrowup, arrowright)\end{tabular} \\

\multicolumn{3}{l}{...} \\ 

150 & \begin{minipage}[c]{0.25\textwidth}  \centering \includegraphics[scale=0.07]{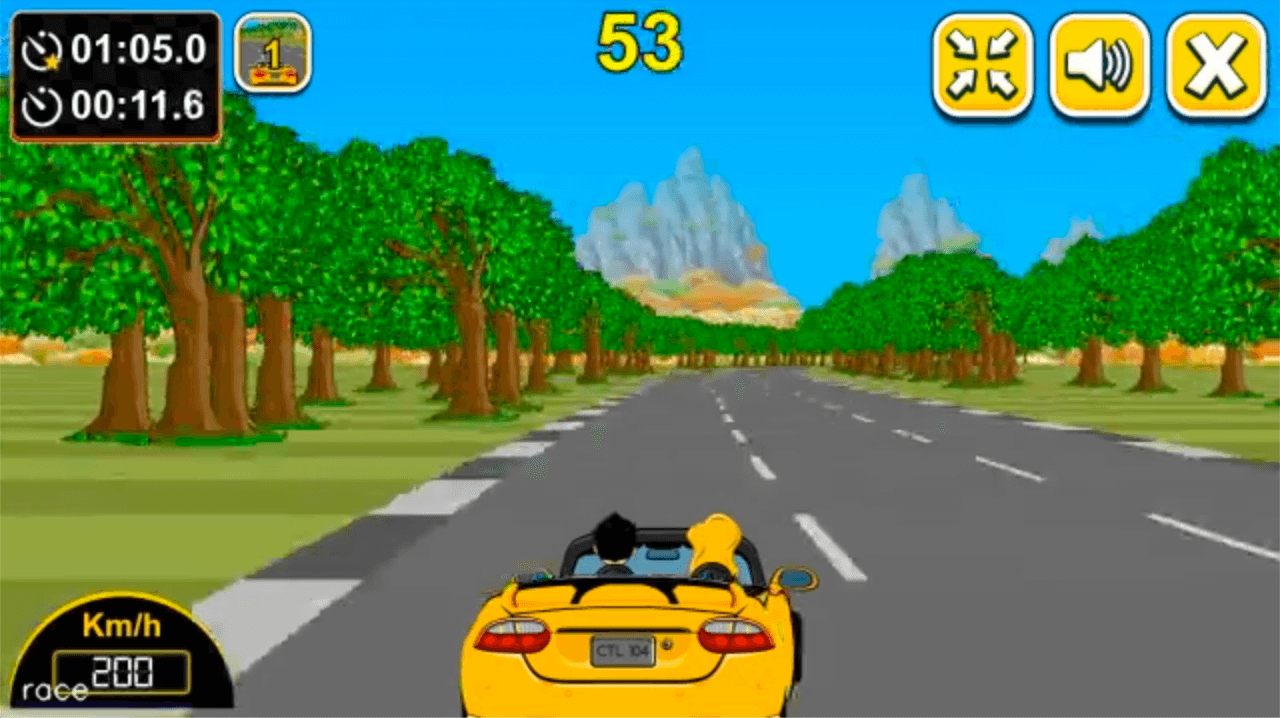}\end{minipage} & \begin{tabular}[c]{@{}l@{}}Action: keyPress(arrowup, arrowright)\end{tabular} \\

\multicolumn{3}{l}{...} \\ 

180 & \begin{minipage}[c]{0.25\textwidth}  \centering \includegraphics[scale=0.07]{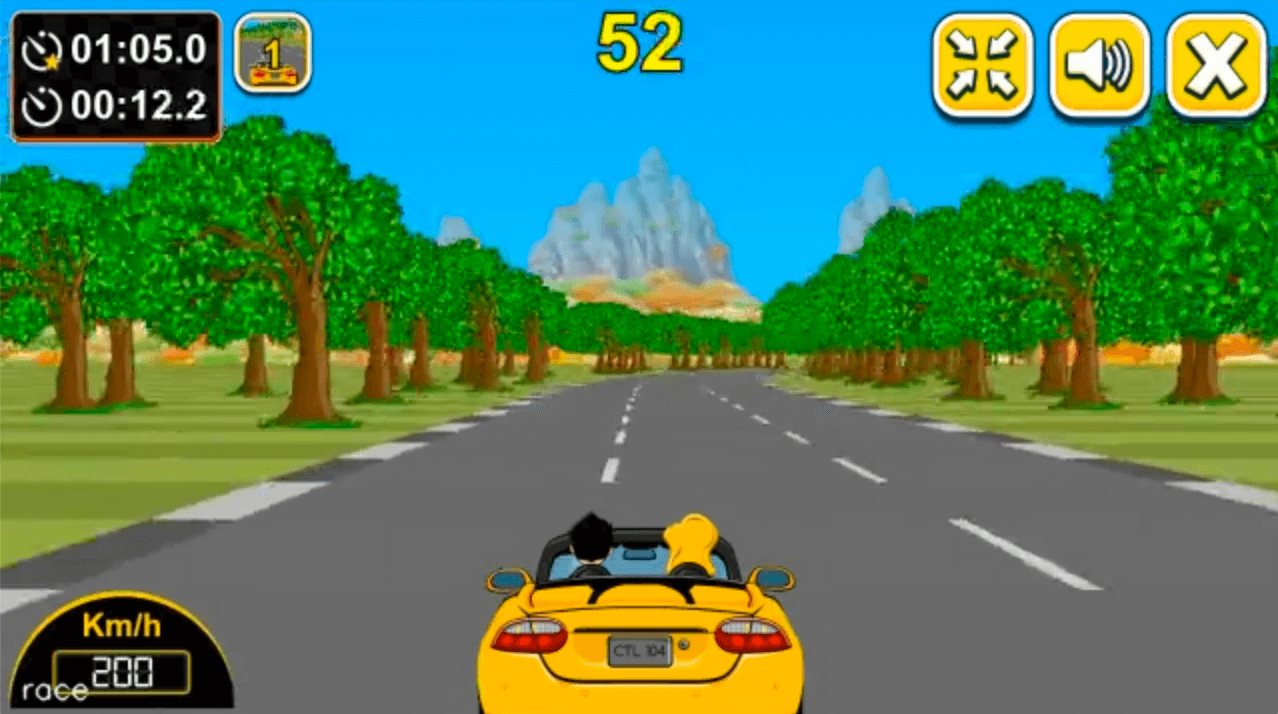}\end{minipage} & \begin{tabular}[c]{@{}l@{}}Thought: The runway is still very straight, and the speed is maintained at 200. There are still no other vehicles \\around, so I can continue to accelerate. Pay attention to the mini-map to ensure I'm in the center of the road, while \\also looking further ahead to anticipate any potential situations. This way, I can maintain the fastest speed. Action: \\keyPress(arrowup)\end{tabular} \\

\multicolumn{3}{l}{...} \\ 

220 & \begin{minipage}[c]{0.25\textwidth}  \centering \includegraphics[scale=0.07]{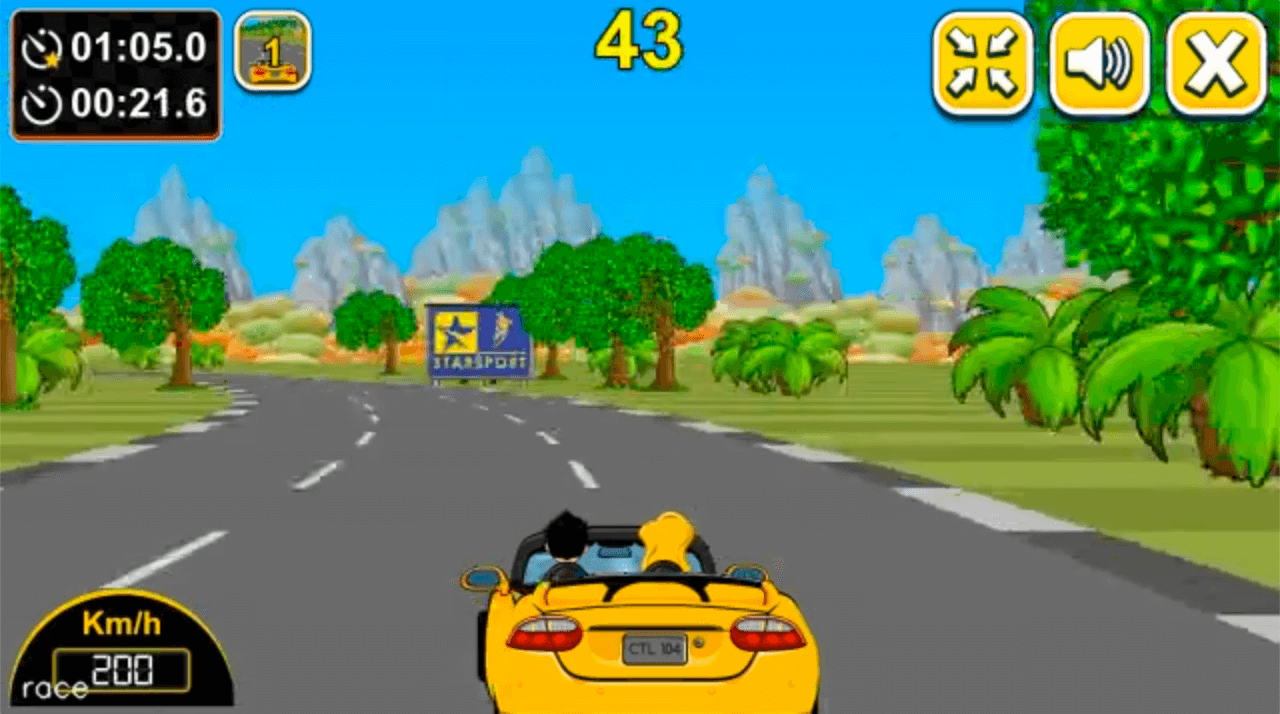}\end{minipage} & \begin{tabular}[c]{@{}l@{}}Action: keyPress(arrowup, arrowleft)\end{tabular} \\

\multicolumn{3}{l}{...} \\ \bottomrule
\end{tabular}%
}
\end{table}

\clearpage

\section{Prompts}

\subsection{Prompts of \gametars on Minecraft}

\begin{prompt}[title={System Prompt of \gametars on Unseen Poki Plane Game}, label=] 
\begin{center}
\begin{minipage}{\dimexpr0.95\linewidth}
\vspace{5pt}

You are an AI agent performing tasks in Minecraft based on given instructions, action history, and visual observations (screenshots). Your goal is to take the next optimal action to complete the task.

\#\# Output Format\\
Thought: ...\\
Action: ...\\

\#\# Action Space\\
* mouseMove(dx, dy) \# Move the mouse position; dx and dy represent horizontal and vertical movement, respectively.\\
* mouseClick('left' or 'right') \# left click or right click the mouse\\
    - left \# Attack; In GUI, pick up the stack of items or place the stack of items in a GUI cell; when used as a double click (attack - no attack - attack sequence), collect all items of the same kind present in inventory as a single stack.\\
    - right \# Place the item currently held or use the block the player is looking at. In GUI, pick up the stack of items or place a single item from a stack held by mouse.\\
* keyPress(keys) \# press the keyboard buttons\\
    - w \# Move forward.\\
    - s \# Move backward.\\
    - a \# Strafe left.\\
    - d \# Strafe right.\\
    - e \# Open or close inventory and the 2x2 crafting grid.\\
    - space \# Jump.\\
    - q \# Drop a single item from the stack of items the player is currently holding. If the player presses ctrl-Q then it drops the entire stack. In the GUI, the same thing happens except to the item the mouse is hovering over.\\
    - 1-9 \# Switch active item to the one in a given hotbar cell.\\
    - left.ctrl \# Move fast in the current direction of motion.\\
    - left.shift \# Move carefully in current direction of motion. In the GUI it acts as a modifier key: when used with attack it moves item from/to the inventory to/from the hotbar, and when used with craft it crafts the maximum number of items possible instead of just 1.\\
* no\_op \# wait and do not interact with the world\\

If multiple actions are activated, use and connect. \\
Use Chinese in Thought part. Provide a brief plan in the Thought section, specifying your next move and objective.\\
If this step does not require thought, you can only output Action.\\
Your history thoughts will accumulate continuously in history conversations.\\

\#\# User Instruction\\
\{instruction\}\\

\vspace{3pt}
\end{minipage}
\end{center}
\end{prompt}

\subsection{Prompts of \gametars on Unseen Games}

\begin{prompt}[title={System Prompt of \gametars on Unseen Poki Plane Game}, label=] 
\begin{center}
\begin{minipage}{\dimexpr0.95\linewidth}
\vspace{5pt}

You are an AI agent playing "Plane Avoid Barrier" game, based on given instructions, action history, and visual observations (screenshots). Your goal is to take the next optimal action to achieve higher scores.\\

\#\# Output Format\\
Thought: ...\\
Action: ... (if thought is necessary in this frame, you need to judge)\\
or\\
Action: ... (low level action is enough)\\

\#\# Action Space\\
* keyPress(keys) \# press the keyboard buttons \\
$~~~~$- a \# Strafe left.\\
$~~~~$- d \# Strafe right.\\
* no\_op \# wait and do not interact with the world. \\

\#\# User Instruction\\
Avoid the obstacle ahead.\\

\vspace{3pt}
\end{minipage}
\end{center}
\end{prompt}

\begin{prompt}[title={System Prompt of \gametars on Unseen Poki Race Game}, label=] 
\begin{center}
\begin{minipage}{\dimexpr0.95\linewidth}
\vspace{5pt}

You are an AI agent playing "Drive Freedom" game, based on given instructions, action history, and visual observations (screenshots). Your goal is to take the next optimal action to achieve higher scores.\\

\#\# Output Format\\
Thought: ...\\
Action: ... (if thought is necessary in this frame, you need to judge)\\
or\\
Action: ... (low level action is enough)\\

\#\# Action Space\\
* keyPress(keys) \# press the keyboard buttons\\
    - arrowup \# Accelerate and start the vehicle.\\
    - arrowdown \# Operate vehicle brakes\\
    - arrowleft \# Turn left \\
    - arrowright \# Turn right \\
* no\_op \# wait and do not interact with the world\\

\#\# Game Tutorial\\
Press the arrowup key to start the vehicle.\\
Hold down arrowup to maintain high speed.\\
At the appropriate time, press arrowleft and arrowright to control steering.\\
**Do not hit the vehicles on the road**, as this will affect your speed.\\
**Avoid hitting the trees and billboards on the roadside**.\\
**Do not leave the road**, as this will slow down your speed.\\

\#\# User Instruction\\
Drive at the fastest speed to reach the destination.\\
\vspace{3pt}
\end{minipage}
\end{center}
\end{prompt}

\begin{prompt}[title={System Prompt of \gametars on Unseen Poki TempleRun Game}, label=] 
\begin{center}
\begin{minipage}{\dimexpr0.95\linewidth}
\vspace{5pt}

You are an AI agent playing "Temple Run" game, based on given instructions, action history, and visual observations (screenshots). Your goal is to take the next optimal action to achieve higher scores.\\

\#\# Output Format\\
Thought: ...\\
Action: ... (if thought is necessary in this frame, you need to judge)\\
or\\
Action: ... (low level action is enough)\\

\#\# Action Space\\
* keyPress(keys) \# press the keyboard buttons\\
    - arrowup \# Make the character jump upward (to avoid ground obstacles such as traps and tree roots)\\
    - arrowdown \# Make the character slide downward (to avoid low-hanging obstacles such as short stone gates and hanging vines)\\
    - arrowleft \# Make the character move to the left (switch to the left track to avoid obstacles on the current track)\\
    - arrowright \# Make the character move to the right (switch to the right track to avoid obstacles on the current track)\\
* no\_op \# wait and do not interact with the world\\

\#\# Game Tutorials\\
1. Survival First: Continuously avoid various obstacles in the game (such as cliffs, traps, flames, moving stone gates, and wild animals) to prevent the character from dying and extend the running duration.\\
2. Resource Collection: Actively collect coins (used to unlock new characters and upgrade power-ups) and power-ups (magnets, shields, speed boosts, revives, etc.) in the scene to increase the amount of resources obtained in a single run.\\
3. Score Improvement: Maximize the score of a single game through "continuous survival time," "number of coins collected," and "score-boosting effects triggered by power-ups" to compete for a higher ranking.\\
4. Scene Exploration: On the premise of ensuring survival, try to explore hidden resources on different tracks (such as hidden coin piles and rare power-up spawn points) and familiarize yourself with the obstacle distribution rules of different scenes (jungles, mines, ancient tombs, etc.).\\

Hint: At the beginning of the game, a red arrow will be used on the screen to tell you which key to press, and when a prompt appears on the screen, press the key while holding the screen prompt.\\
\#\# User Instruction
Play the temple run game.\\
\vspace{3pt}
\end{minipage}
\end{center}
\end{prompt}

\begin{prompt}[title={System Prompt of \gametars on Unseen Poki Plane Game}, label=] 
\begin{center}
\begin{minipage}{\dimexpr0.95\linewidth}
\vspace{5pt}

You are an AI agent playing "Jump Only" game, based on given instructions, action history, and visual observations (screenshots). Your goal is to take the next optimal action to achieve higher scores.\\

\#\# Output Format\\
Thought: ...\\
Action: ... (if thought is necessary in this frame, you need to judge)\\
or\\
Action: ... (low level action is enough)\\

\#\# Action Space\\
* keyPress(keys) \# press the keyboard buttons \\
    - arrowleft  \# Jump to the upper left\\
    - arrowright \# Jump to the upper right\\
* no\_op \# wait and do not interact with the world. You will fall freely\\

Each step will activate only one key.\\

\#\# Game Tutorial\\
Jump! This is your only way to escape. "Jump Only" is a high-skill platform game where you can showcase your amazing reaction time and jumping skills. \\
On your way to the finish line, you need to jump over dangerous obstacles, avoid spikes, and saw blades. \\
There are 49 different levels for you to conquer. The theme changes every few levels, bringing you different challenges.\\

Hint: You need to master the most suitable timing for jumping. Holding down the left or right key will make you jump upwards. You cannot keep pressing left or right continuously, and you need to use the no\_op key to wait until you land in a suitable position before jumping again.\\

\#\# User Instruction\\
Avoid obstacles and jump to the finish line.\\

\vspace{3pt}
\end{minipage}
\end{center}
\end{prompt}

\end{document}